\documentclass[twocolumn]{article}
\usepackage[utf8]{inputenc}
\usepackage{blindtext}
\usepackage{graphicx}
\usepackage{multicol,caption,float,amsmath,array,wrapfig,authblk,booktabs}
\usepackage{pdflscape}
\usepackage{afterpage}
\usepackage{tablefootnote,footmisc}
\usepackage{makecell}
\usepackage{placeins}
\usepackage{xr}
\usepackage[T1]{fontenc}

\usepackage[a4paper,top=2cm,bottom=2cm,left=2cm,right=2cm,marginparwidth=1.75cm]{geometry}
\usepackage[
  backend=biber,
  style=numeric,
  maxbibnames=99,
  giveninits=true,
  nohashothers=true,
  nosortothers=false,
  eprint = false,
  ]{biblatex}
\title{Benchmarking of Clustering Validity Measures Revisited}

\begin{document}

\author[1]{Connor Simpson}
\author[2]{Ricardo J. G. B. Campello}
\author[3]{Elizabeth Stojanovski}
\affil[1]{School of Information and Physical Sciences, The University of Newcastle \\
email: connor.simpson@uon.edu.au }
\affil[2]{Department of Mathematics and Computer Science, University of Southern Denmark \\
email: campello@imada.sdu.dk}
\affil[3]{School of Information and Physical Sciences, The University of Newcastle \\
email: elizabeth.stojanovski@newcastle.edu.au}
\renewcommand\Affilfont{\itshape\small}
\date{}

\twocolumn[
  \begin{@twocolumnfalse}
    \maketitle
    \begin{abstract}
    Validation plays a crucial role in the clustering process. Many different internal validity indexes exist for the purpose of determining the best clustering solution(s) from a given collection of candidates, e.g., as produced by different algorithms or different algorithm hyper-parameters. In this study, we present a comprehensive benchmark study of 26 internal validity indexes, which includes highly popular classic indexes as well as more recently developed ones. We adopted an enhanced revision of the methodology presented in \cite{Vendramin2010}, developed here to address several shortcomings of this previous work. This overall new approach consists of three complementary custom-tailored evaluation sub-methodologies, each of which has been designed to assess specific aspects of an index's behaviour while preventing potential biases of the other sub-methodologies. Each sub-methodology features two complementary measures of performance, alongside mechanisms that allow for an in-depth investigation of more complex behaviours of the internal validity indexes under study. Additionally, a new collection of 16177 datasets has been produced, paired with eight widely-used clustering algorithms, for a wider applicability scope and representation of more diverse clustering scenarios. \\
    {\bf Keywords:} clustering, cluster validation, benchmark 

    \end{abstract}
    \vspace*{3mm}
  \end{@twocolumnfalse}
]

\tableofcontents
\section{Introduction} \label{Introduction}

Clustering is an unsupervised learning technique that aims to identify patterns that consist of similar or inter-related observations within data \cite{Jain1988,Xu2005}. 
Many existing clustering algorithms are often categorised into three primary groups \cite{Jain1988,Vendramin2010}: partitioning algorithms such as K-Means \cite{Jain1988} and Spectral Clustering \cite{Yu2003}, hierarchical algorithms such as Single Linkage \cite{Jain1988} and HDBSCAN* \cite{Campello2013,Campello2015}, and soft (fuzzy or probabilistic) algorithms such as Fuzzy c-Means (FCM) \cite{Bezdek1984} and Expectation Maximisation with Gaussian Mixture Models (EM-GMM)  \cite{doi:10.1198/016214502760047131}. Partitioning clustering algorithms partition data into a given number of $k$ clusters, while hierarchical clustering algorithms produce a sequence of nested partitions with incrementally varying numbers of clusters. Soft clustering algorithms are similar to partitioning techniques except that each data observation is assigned a degree of membership or probability to each cluster, rather than a full assignment to a single cluster. It is worth mentioning that within the aforementioned categories there are clustering algorithms that may not necessarily assign all observations to clusters, due to outlier trimming or noise detection. Two examples of such algorithms are trimmed K-means \cite{10.1214/aos/1031833664} and the previously mentioned HDBSCAN*, each of which may produce solutions where not all observations are assigned to clusters.

\emph{Clustering validation} or \emph{validity} is an important step of the clustering process irrespective of the algorithm used \cite{Jain1988,Halkidi2001}, as it is crucial to determine the best produced partition(s) and number of clusters within the data \cite{Gurrutxaga2011}. Two main categories of clustering validation methods exist, namely, \emph{internal validation} and \emph{external validation} \cite{Lei2017}. Internal validation techniques use only the internal information of the clustering problem, while external validation techniques function by comparing the produced partitions to a known ground-truth partition. However, internal validation techniques are more commonly utilised in real-world clustering due to the lack of a known ground-truth partition in practical clustering problems.

Many different \emph{internal validity indexes} have been proposed to determine the best partitioning of a dataset \cite{Milligan1985a,Halkidi2001}, which can be split into two basic categories: difference-like criteria and relative indexes \cite{Vendramin2010}. Difference-like criteria assess a sequence of partitions with a varying number of clusters, generally produced by a hierarchical algorithm, and try to determine the optimal cut-off point by comparing consecutive candidates. Difference like-criteria therefore operate strictly as \emph{stopping rules} to determine the number of clusters in very specific (e.g. hierarchical) clustering scenarios. Relative or optimisation-like criteria, on the other hand, are much more general and widely applicable, as they can relatively compare and rank any set of candidate partitions produced from a given dataset. Given their wider applicability, this study focuses exclusively on relative criteria.

There is an extensive history of studies comparing the performance of internal validity indexes, such as the seminal papers of Milligan and Cooper \cite{Milligan1981,Milligan1985a}. Several different methodologies for studying internal validity indexes have been produced, however, many studies are still based on the work of \cite{Milligan1985a}, or a variant of it, as they rely on assessing each index's ability to determine a single best partition from a collection of candidate partitions.

There are two primary methods of determining the best partition, which is assumed to be either the partition containing the same number of clusters as a ground-truth partition \cite{Milligan1985a} or the partition that produces the highest value of an external validity index when compared against a ground-truth partition \cite{Vendramin2010}. These will be referred to as the \emph{optimal number of clusters method}, where $k^*$ is the number of clusters in the ground truth, and the \emph{optimal partition method}, where $k_o$ is the number of clusters in the optimal partition according to an external index, respectively. 

\begin{figure}[h]
 \centering
 \includegraphics[width=\linewidth]{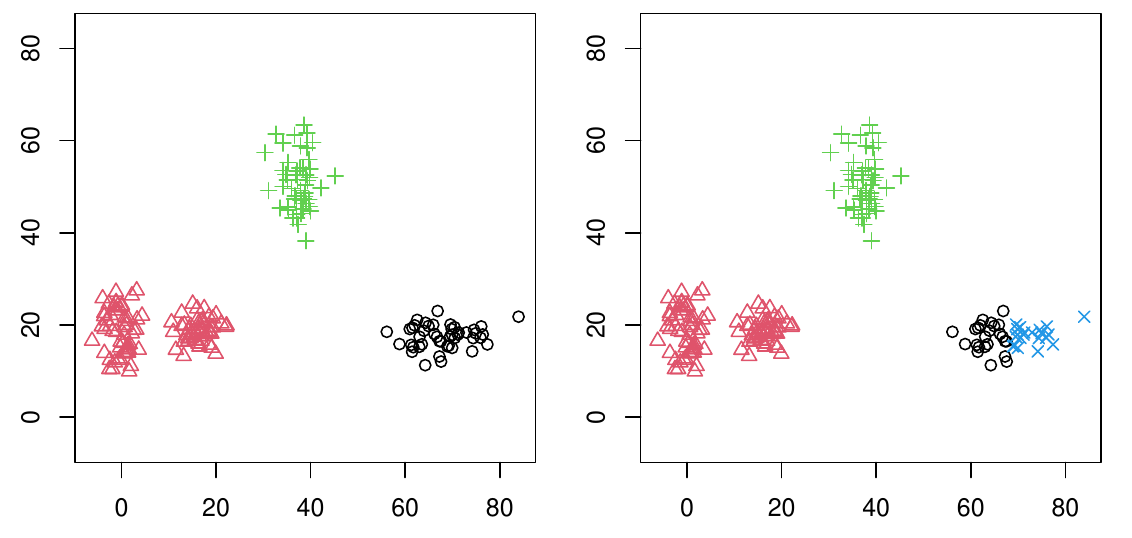}
 \caption{\label{fig:NcIssue} Example partitions produced by K-Means for a dataset with four ground-truth clusters, where the partition with $k=3$ clusters (left) produces a better solution compared to the solution with $k=4$ clusters (right). The groupings of the partitions are represented by the shape and color of the observations.}
\end{figure}

The optimal number of clusters method aims to determine whether an internal index can correctly identify the number of clusters in a dataset; however, this method can be flawed as it assumes the best candidate partition will contain the same number of clusters as the ground truth. This is not necessarily the case, as clustering algorithms may produce poor solutions for the optimal number of clusters while producing better solutions with fewer or greater granularity. An example can be seen in Figure \ref{fig:NcIssue}, where the solution produced with the correct number of clusters is notably worse compared to the solution with the incorrect number. This will lead to internal indexes that select a worse partition with the ``correct" number of clusters being deemed better than an index that selects a good partition with the ``incorrect" number of clusters. 

The optimal partition method circumvents the aforementioned issue by using an external validity index that takes the entire ground-truth partition as a reference, rather than just its number of clusters. However, the use of an external validity index presents its own challenges as such indexes have been noted to exhibit different behaviours in determining partition quality. Additionally, some indexes show potential biases in the form of systematic tendencies to favour certain types of clustering regardless of the true underlying structure of the data, such as being monotonically increasing or decreasing with the number of clusters under specific conditions \cite{Pfitzner2009,Horta2015,Lei2017,Arinik2021}. Without an objective measure of performance available, these potential biases were previously studied using controlled test datasets, for example datasets where no clusters are present, which may not reflect real world use cases, or may instead represent subjective differences in behaviour of the indexes. Either way, differences in behaviours of external indexes pose the additional challenge of a careful, proper selection of such index(es) for the purposes of internal index evaluation based on ground-truth reference partitions.

A second issue common to both methods of assessing internal validity indexes is that they equally penalise an index for selecting as best a slightly worse or a much worse partition than the reference best partition. The study in \cite{Vendramin2010} noted that as a result of only assessing an index's ability to identify a single best partition, the index's ability to assess all other partitions, and thus its ability to distinguish between better and worse partitions more broadly (and possibly rank them), is not accounted for. This aspect of clustering validation is important as it reflects the robustness of an index to perform reasonably well even in more difficult model selection scenarios when an ideal candidate partition may not be available or cannot be trivially found by an index. This is particularly important when considering real-world data where, depending on the application, there may not be a single ``correct'' partition but rather a selection of reasonable partitions that should be discriminated from poor solutions, or where the best solution may be specific to the use of the data \cite{VonLuxburg2012}.

Many studies of internal validity measures also present more general issues. For instance, there are only a few studies that compare a large number of indexes, such as those in \cite{Milligan1985a,Vendramin2010,Arbelaitz2013}, which each evaluate at least 30 indexes. In contrast, most studies focus on a smaller number of indexes, typically less than 10, as seen in the work of \cite{Liu2010a,Guerra2012,Meroufel2017,Hamalainen2017}. The current literature also contains a heavy focus on older validity indexes and often fail to include newer types of indexes such as validation criteria for density-based clustering. Furthermore, the number of datasets used within the current literature ranges from 5 \cite{Liu2010a} to 1080 \cite{Vendramin2010} unique datasets with only a limited range of properties considered. This limits the interpretability of results from these studies as both the data and clustering algorithms used have been shown to affect the performance of internal validity indexes \cite{Guerra2012}. Although it is not possible to capture all possible clustering problems, we extend the range of test data to include a new collection of 16177 unique datasets featuring 7 properties, in addition to 972 of the datasets examined in \cite{Vendramin2010}.

The study \cite{Vendramin2010} attempted to address many of the aforementioned issues by introducing an improved benchmarking methodology that aimed to measure an internal index's performance for determining the quality of partitions across a diverse range of candidate solutions. This was achieved by measuring the correlation between each internal validity index and an external validity index in controlled experiments. Such a strategy is lenient towards indexes selecting good solutions rather than simply either the optimal solution or the solution with the ``correct'' number of clusters. By reducing the penalty for selecting a high-quality alternative solution, it provides robustness in the case where valid solutions outside the ground truth exist. Notice that this methodology relies on the assumption that external validity indexes are an accurate measure of partition quality, and therefore, a well performing internal index should be highly correlated with the results of an external index despite the fact that the former only uses the data and the candidate clustering solutions under assessment, whereas the latter also has access to a ground-truth solution (which would not be available in practical clustering applications), used as a reference. As there is an increased reliance on an external validity index to not just highlight a single solution, but also to accurately rank solutions relative to the ground truth, this methodology increases the possibility for the behaviours of different external indexes to impact the results of a study.

\begin{figure}[h]
 \centering
 \includegraphics[width=\linewidth]{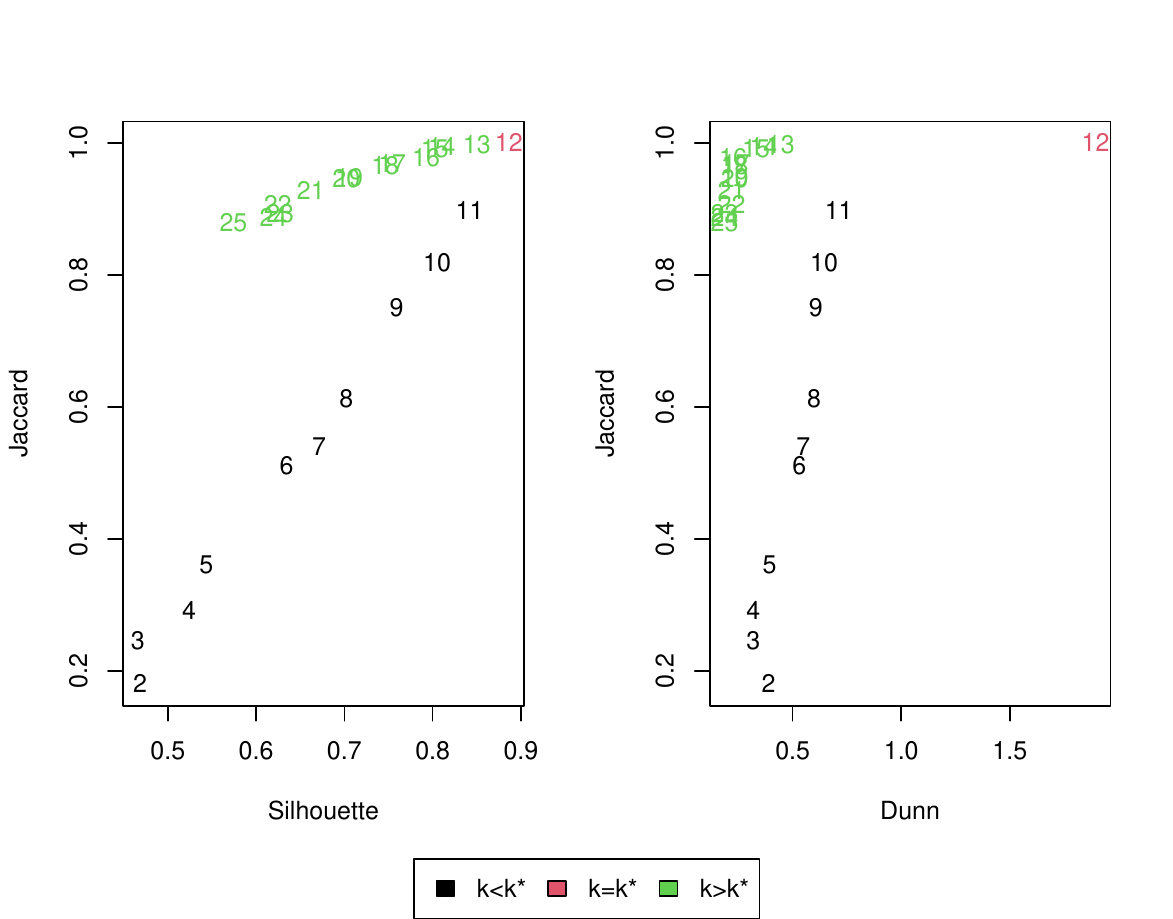}
 \caption{\label{fig:CorrelationExample} Scatter plots of the external Jaccard index against two internal indexes, Silhouette (left) and Dunn (right). Each point represents a partition, where the number indicates the number of clusters within that partition, and the color indicates if the number of clusters is less than (black), equal to (red) or greater than (green) the ground-truth number of clusters. Both indexes select the ground-truth solution (top-right corner) as the best partition, however, when considering all the other candidate solutions the Pearson correlation with the Jaccard index is 0.77 for the Silhouette index and only 0.04 for the Dunn index.}
\end{figure}

In Figure \ref{fig:CorrelationExample} it can be seen how this method can be used to distinguish between two indexes that both select the ground-truth partition as the best, however, one index performs significantly better at ranking all other partitions, which is reflected by its higher correlation with the external index (Jaccard). It can be seen that if the best partition highlighted by both indexes (right-/upper-most point at the top-right corner) was not produced by the clustering algorithm, then the Silhouette index would select a significantly better partition (with 15 clusters and a Jaccard value close to 0.9) as compared to the Dunn index (with 11 clusters and a Jaccard value less than 0.6).

\begin{figure}[h]
\centering
\includegraphics[width=\linewidth]{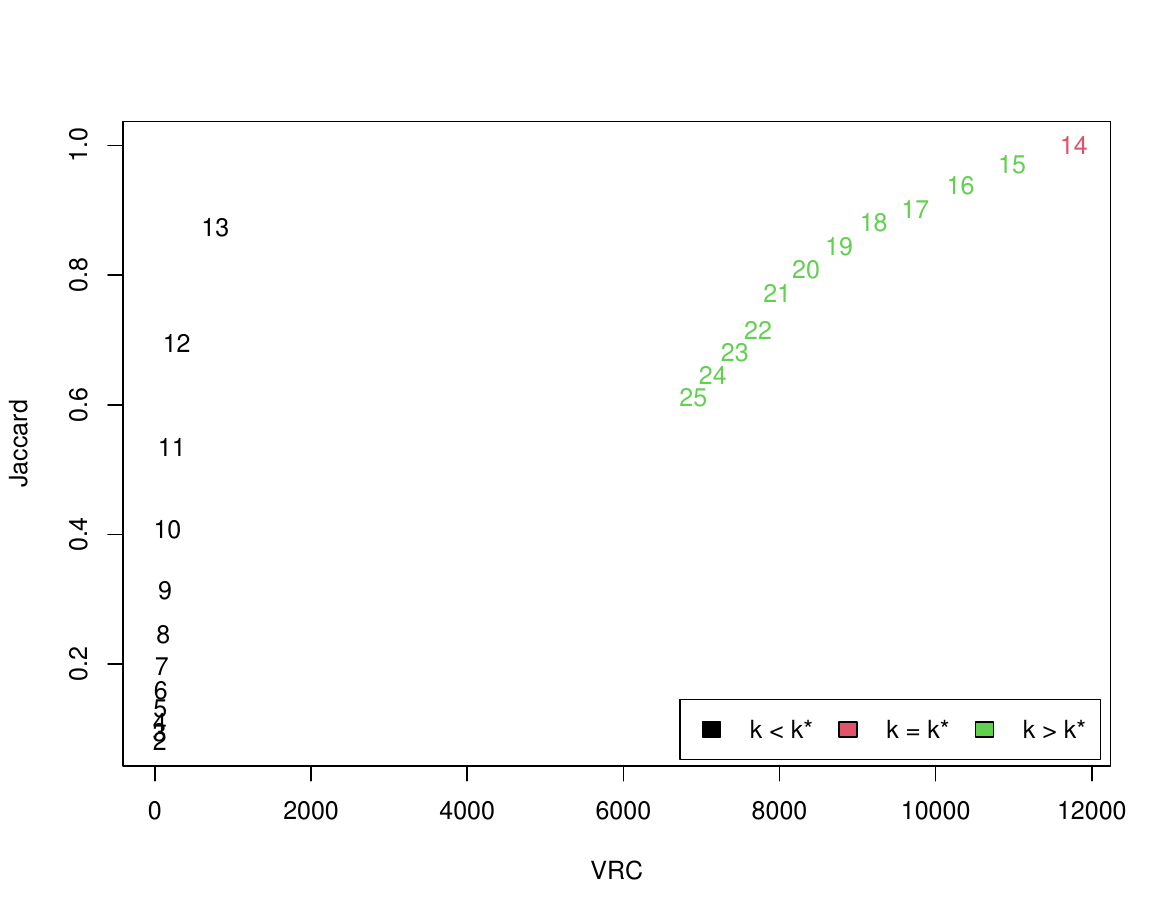}
\caption{\label{fig:example1} External index (Jaccard) plotted against an internal index (VRC) for a test dataset. A non-linear relationship can be seen to form two regions in the plot based on the partition containing a larger ($k > k^*$) or smaller ($k < k^*$) number of clusters compared to the ground-truth partition.}
\end{figure}

\begin{figure}[h]
\centering
\includegraphics[width=\linewidth]{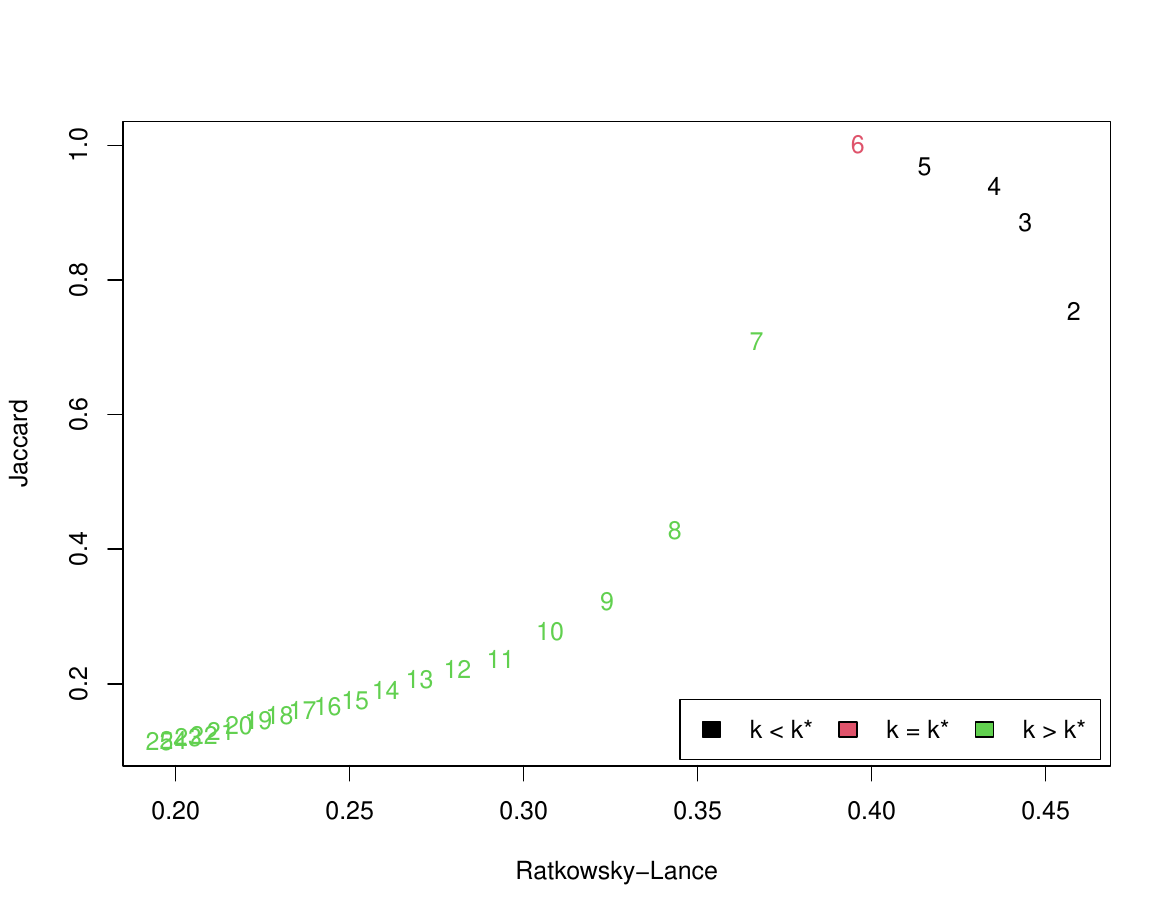}
\caption{\label{fig:example2} Ratkowsky-Lance plotted against an external (Jaccard) index for a test dataset. The internal index exhibits a monotonically decreasing behaviour as a function of the number of clusters, which still results in a high Pearson correlation of 0.93.}
\end{figure}

Despite the apparent improvements of this methodology when compared to the previous work, it was not noticed in \cite{Vendramin2010} that many internal indexes exhibit non-linear relationships that cannot be properly captured by the correlation measures used in that study, such as the Pearson correlation. There may be non-linear relationships potentially missed or underestimated by this measure where the partitions are still correctly ranked by an internal index. Contrarily, there may be non-linear relationships that will be partially captured despite the fact that they indicate an undesirable behaviour of the internal index in question. An example of the former can be seen in Figure \ref{fig:example1}, while Figure \ref{fig:example2} illustrates the latter case.

It is noted in Figure \ref{fig:example1} that both the internal (Variance Ratio Criterion --- VRC) and external (Jaccard) indexes decrease in value as clusters are either successively merged or split starting from the ground-truth solution; however, when all candidate solutions are considered together, the overall relationship is highly non-linear as it is divided into two distinct regions that cannot be properly captured by a linear correlation measure. In contrast, Figure \ref{fig:example2} shows that the internal index (Ratkowsky-Lance) appears to be monotonically increasing with the number of clusters in a partition, even beyond the correct number. However, the relationship between this undesirable behaviour and the proper behaviour of the external index is still captured quite prominently by a linear correlation (Pearson = 0.93) despite being non-linear and undesirable.  

Some papers following the methodology introduced in \cite{Vendramin2010} --- e.g., \cite{jaskowiak_area_2022,Nguyen2020} --- have utilised Spearman correlation instead, which does not assume a linear relationship but instead measures general monotonic relationships. This makes Spearman correlation less sensitive to differences in scale or distribution compared to Pearson correlation, however, more complex non-linear relationships may also exist.
Understanding when and how these relationships occur is an important step towards properly assessing the performance and reliability of internal validity indexes, which may also require visual inspection. Visual inspection of these relationships may provide insight into the reasons behind disagreements in the results obtained in previous studies when adopting different evaluation methodologies. For instance, in \cite{Vendramin2010} the Ratkowsky-Lance ($C\over{\sqrt{K}}$) index performed poorly in terms of its ability to select the best candidate partition, yet it exhibited one of the highest correlations with the external index across the collection of candidate partitions. Figure \ref{fig:example2} disambiguates this apparent contradiction, showing that the latter result is actually a misleading artifact of the use of correlation, which in this case is high despite the fact that the index mostly fails to properly distinguish between good and bad partitions and instead appears to decrease monotonically with the number of clusters. These cases demonstrate a need for an overhaul of the comparison methodology with a focus on the specific relationships between external and internal indexes, since their summarised representation as single correlation values cannot alone sufficiently describe these complex relationships.

At a fundamental level, it should be noted that studies using external validity indexes and external information unavoidably rely on a preconceived, particular definition of a ``true'' or ``ground-truth'' clustering. Some studies have recognised that recovery of a ground-truth partition is not the only valid measure of clustering performance, as multiple valid solutions may exist within a problem, the best of which may depend on the specific use case \cite{HENNIG201553,https://doi.org/10.1002/widm.1511}. Consequently, the method of evaluation used in benchmarking should be carefully chosen to align with the goals of the study.

Although the performance in recovering a ground truth is not the only criterion that could be considered, it may be used as a proxy for specifying --- in the form of cluster labels --- what type of clustering is cared about within a particular context.  While in specific application domains such labels could be obtained, e.g., through manual annotation by domain experts, in this study the ``ground-truth’’ labels are reflective of the underlying mechanisms behind controlled synthetic data generation. We advocate that these mechanisms, such as known probability density functions, are in alignment with the domain-agnostic, statistical motivation behind most if not all the internal validity indexes considered in this study. Indeed, internal validity indexes implicitly or explicitly subsume in their own definitions what type of clustering they are based upon, and the use and adoption of such indexes presumes that this is the clustering view one is interested in.

In summary, this paper aims to make the following major contributions to the field of internal validity index evaluation and benchmarking:
\begin{itemize}
  \item An improved evaluation methodology building upon \cite{Vendramin2010}, based on both visual as well as statistical analyses of the relationships between external and internal validity indexes to produce a deeper understanding of the behaviour of the latter that does not exclusively rely on linear correlation.
  \item A complementary evaluation methodology where the external index is replaced by a reference ranking of partitions that allows for internal indexes to be assessed --- in certain evaluation setups --- independently of any particular external indexes and their different behaviours or potential biases.
  \item An extensive benchmarking study that makes use of the above methodologies and goes beyond previous studies by adopting:
  \begin{itemize}
  \item An updated list of internal validity criteria (as compared to previous studies) including both classic and more recent indexes.
  \item An extended list of clustering algorithms, including: K-Means, Hierarchical clustering, EM-GMM, HDBSCAN* and Spectral Clustering. This aims to produce partitions that are better representative of a more diverse suite of potential application scenarios.
  \item Multiple external validity criteria including a distance sensitive external index which takes into account the geometry of clustering solutions in addition to the clustering labels.
  \item An extended collection of clustering problems containing 16177 unique datasets which also includes the datasets from \cite{Vendramin2010} as a subset for a comparative analysis of the results. The extended collection of datasets will represent clustering problems with varying dimensionality, cluster shapes, density, size, balance, level of overlap, and noise. 
  \end{itemize}
\end{itemize}

The paper is structured as follows:
Section \ref{related} covers the history of previous studies on internal validity and the common methodologies adopted therein. In Section \ref{methods}, we describe the methodologies used within this study, including the selection of validity indexes, the data utilised, and a description of the three evaluation methodologies employed to assess each index. Section \ref{results} presents the results of each of the three methodologies. The primary results for Evaluation Scenario 1, where the number of clusters are varied, are found in Section~\ref{sec:eval1}, featured in Tables \ref{Table:Experiment1}, \ref{Table:Exp1Alg} and \ref{Table:CorProp}, which display the overall rankings of each of the indexes, the rankings of indexes separated by clustering algorithm, and the rankings of indexes separated by the properties of the data, respectively. The primary results for Evaluation Scenario 2, where all solutions contain the same number of clusters, are found in Section \ref{sec:eval2}, featured in Tables \ref{Table:Exp2Correct}, \ref{Table:Exp2Cor}, and \ref{Table:RepProp}, which provide the rankings of each index based on ``top pick'' percentage, correlation, and separated by the properties of the data, respectively. Finally, the results for Evaluation Scenario 3, where a benchmark is conducted independent of the clustering algorithms and external validity indexes, is found in Section \ref{sec:eval3}, along with a discussion comparing the results to Evaluation Scenarios 1 and 2. Section \ref{discussion} offers a detailed discussion of the overall results. In Section \ref{comp}, we compare the results of our study to the previous study \cite{Vendramin2010}, highlighting key differences and similarities. The conclusion is presented in Section \ref{conclusion}, alongside a brief summary of main takeaways, including Table \ref{Table:Recommended}, which displays general recommendations for the best performing indexes and which clustering algorithms and potential data properties they have functioned the best with according to our experiments.

\section{Related Work} \label{related}

The related work will be subdivided into four parts. The first part discusses common methodologies used in the literature to assess the behaviour and performance of internal validity indexes as well as their advantages and disadvantages. The second part discusses the selection of validity indexes used across benchmarking studies. The third part discusses the clustering algorithms used for assessing the performance of such indexes in these studies, whereas the fourth part discusses the range of data and properties adopted.

{\bf Evaluation Methodologies:} There are two primary reasons to carry out a comparison study of validity indexes, the first being to survey the current literature and determine which available indexes tend to be more suitable or perform better in certain classes of clustering problems, as in \cite{Milligan1985a,Vendramin2010,Arbelaitz2013}, and the second is to compare a new index to previous indexes, as in \cite{Moulavi2014a,jaskowiak_area_2022,Liang2020}. Studies with the former motivation are of primary interest here due to 
comparing larger collections of indexes in varied scenarios. In contrast, studies with the latter motivation are of limited scope and relevance as they naturally tend to focus on the areas of strength and superiority of the proposed index \cite{https://doi.org/10.1002/widm.1511}.

Studies of clustering validation techniques have a long history dating back to an early paper by Milligan and Cooper \cite{Milligan1985a}, where the performance of 30 internal validity indexes were assessed through observing the frequency of each index selecting the partition with the optimal number of clusters (out of a set of candidate partitions produced through hierarchical clustering). Many studies of clustering validity indexes have been inspired from this seminal work and multiple variants of its methodology have been proposed. For instance, the most basic method used for evaluating the performance of internal validation indexes involves generating a range of candidate partitions for a dataset using a clustering algorithm such as K-Means, and then comparing the indexes based on their ability to identify the partition with the optimal number of clusters as determined by a ground-truth partition. This is often performed on few (most commonly toy) datasets as in \cite{Halkidi2001,Liu2010a,Starczewski2017,Cebeci2019,Zhou2021}, thus limiting its usefulness to understanding how specific properties in such datasets may impact the performance of validation indexes. Although this procedure does not provide a comprehensive evaluation of an index due to the limited scope of data as well as its focus purely on the number of clusters, it has been the primary method adopted in papers introducing new indexes, e.g., \cite{Hu2019,Xiong2019,Liang2020}.

A more systematic generalisation of this method, similar to that originally proposed in \cite{Milligan1985a}, adopts a more diverse collection of datasets and ranks the indexes based on the percentage of datasets for which they correctly identify the optimal number of clusters. This methodology provides a more comprehensive assessment of an index's performance compared to the previous method, however, it does have critical shortcomings as highlighted in \cite{Vendramin2010}. The main limitation is that this method does not account for the possibility that the best partition produced by a clustering algorithm may not have the same number of clusters as the ground truth. This can result in the penalisation of indexes that correctly identify good partitions with a different number of clusters, while indexes that incorrectly identify poor partitions that happen to have the same number of clusters as the ground truth will be favoured. Despite these limitations, this methodology is still commonly used --- e.g. \cite{Guerra2012,Li2016a,Chouikhi2015,Hamalainen2017}.

Papers such as \cite{Gurrutxaga2011,Arbelaitz2013,Moulavi2014a,Liang2020} have employed methodologies similar to those described above, with the main difference being that they use as referential target the optimal partition of a dataset as determined by an external validity index, rather than just the number of clusters in the ground truth. The indexes are ranked by their success rate at selecting the optimal partition for each dataset (as determined by the external index) instead of the optimal number of clusters from the ground truth. This approach provides a significant improvement in the reliability of the results, as indexes cannot be incorrectly penalised for selecting the best partition, or favoured for selecting a bad partition that happens to contain the optimal number of clusters. However, criticisms have been raised about these basic methodologies as they penalise selecting a sub-optimal (slightly worse or nearly identical) partition the same as selecting a significantly worse partition \cite{Vendramin2010}. This issue can be seen as a consequence of the assessment strategy fully relying on the quality of a single partition elected as the best by each index, regardless of how well or poorly the other candidate partitions have been evaluated. 

Studies such as \cite{Baarsch2012,Luna-Romera2018,Karanikola2021} circumvent this drawback by measuring instead the absolute difference between the number of clusters in the best partition as indicated either by the number of clusters in the ground truth or an external index, and the number of clusters in the best partition as indicated by the internal index. The internal indexes are then ranked based on their mean difference, where a lower difference is considered better. However, this methodology again relies on the incorrect assumption that the number of clusters is an accurate measure of partition quality, which can produce misleading results as previously discussed.

The study in \cite{Vendramin2010} proposed that if an external validity index is a good measure of clustering quality when a ground-truth solution is available as reference, then a good internal validity index should be highly correlated with the results of the external index despite not making use of any ground truth. Based on this assumption, the authors adopted the Pearson correlation between an external validity index and each internal index under assessment as a measure of the quality of the latter. This has several advantages compared to previous evaluation methods as it allows measuring an index's ability to distinguish between good and bad partitions, rather than simply an index's ability to detect a single best partition or the correct number of clusters. In other words, the reliability of an index is more clearly captured by this methodology. However, as we will see in subsequent sections, the use of Pearson correlation is questionable as it only accurately measures linear correlations. Another issue is that this methodology relies on the assumption that the adopted external index is a gold standard measure of partition quality, however, several authors have noted different types of biases in various external indexes \cite{Pfitzner2009,Horta2015,Lei2017,Arinik2021}.

Subsequent studies have also adopted the methodology from \cite{Vendramin2010} for assessing internal clustering validation measures \cite{Rabbany2012,Moulavi2014a,Horta2015,jaskowiak_area_2022}. In \cite{Nguyen2020}, the same basic methodology was followed, however, the authors made two subtle changes, namely, they adopted Spearman correlation, which is a rank-based correlation, and they used multiple external validation indexes in part of their analyses. The Spearman correlation coefficient measures general monotonic relationships removing the impact of differing scales between indexes, while the combined use of multiple external indexes we expect to mitigate their differences and any potential individual biases to some extent. 

The studies in \cite{Jose-Garcia2021} and \cite{Gagolewski2021} use a methodology where the value of an external index for the best partition selected by each internal validity index is recorded as their evaluation metric. The analysis then involves comparing the distribution of these values for each candidate internal index to assess their performance. This process has the benefit of not pre-defining a best partition for the indexes to identify, avoiding many of the issues seen in previous studies as indexes will not be penalised for selecting good partitions that are not predefined and targeted as best. However, this method is still only assessing the quality of the best solutions presented by each index and not an index's ability to distinguish between good and bad partitions.

A common addition seen in many of these methodologies is the use of statistical tests to determine if there is a significant difference in performance between the internal validity indexes. The study in \cite{Vendramin2010} uses two statistical tests, the Willcoxon/Mann-Whitney and the Friedman test, with a 5\% significance level to determine if there is a statistically significant difference in performance. In \cite{Arbelaitz2013} a Shaffer test with a 10\% significance level is used, and according to these results, indexes are then categorised into three groups based on their performance. The paper \cite{Karanikola2021} utilised the Friedman test alongside a Bonferroni-Dunn post-hoc test to determine statistical significance. Finally, \cite{Nguyen2020} carried out ANOVA with each internal index as well as the properties of the datasets as factors, where the F-test was used to determine significance of the indexes and dataset properties. 

{\bf Internal Validity Indexes:} The number and range of internal validity indexes assessed can vary significantly between benchmarking studies. Several large-scale studies exist such as the one performed by Milligan and Cooper \cite{Milligan1985a}, which assessed 30 indexes including both difference-like and relative criteria. This list was extended to 40 internal indexes in \cite{Vendramin2010}. Other papers assessing a wide range of internal indexes include \cite{Arbelaitz2013} and \cite{Chouikhi2015}, both of which evaluated 30 indexes. In contrast, many studies only assess a small collection, such as \cite{Liu2010a} with 11 indexes and \cite{Guerra2012} with 5 indexes.

Despite new internal validity indexes being proposed in the clustering literature from time to time, many studies focus on similar selections consisting of primarily well-established classic indexes. This may result in the potential strengths of newer indexes not being recognised due to a lack of comparative assessment. Studies such as \cite{Liang2020} stand out as much less usual for also including more recent types of indexes, most noticeably, internal validation measures for evaluation of density-based clustering results.

{\bf Clustering Algorithms:} The clustering algorithms used in studies assessing internal validity measures are a relevant consideration as the types of partitions produced can vary between different algorithms. Some indexes may be better suited to determining the best solution for specific types of algorithms, typically for explicitly or implicitly making similar assumptions about the clustering problem and the underlying nature of the expected solutions.

Many previous studies, however, feature very limited types of algorithms in the experiments, with the primary algorithms used being K-Means \cite{Vendramin2010,Arbelaitz2013,Li2016a,Nguyen2020,Karanikola2021,Jose-Garcia2021} and linkage-based hierarchical algorithms \cite{Vendramin2010,Arbelaitz2013,Nguyen2020,Karanikola2021,Jose-Garcia2021}. Other clustering algorithms found in comparisons studies are EM-GMM \cite{Guerra2012}, Spectral Clustering \cite{Nguyen2020}, and an Evolutionary Clustering Algorithm \cite{Jose-Garcia2021}. Noticeably, density-based clustering algorithms are seldom seen in any studies despite their wide use in practice and growing popularity.

{\bf Datasets}: There are two primary types of data used in studies of internal validation measures. The first is synthetic data produced for the purpose of assessing clustering tools where an accurate ground-truth partition is precisely known, e.g., from the true data distribution. The second is real-world data where a reference partition can be obtained and used as ground truth, such as from domain knowledge.

As synthetic or simulated datasets are produced for the purpose of testing, there is a much greater level of control over the properties within the data compared to real-world datasets. Previous studies have shown significant variation in the number of datasets used, ranging from 5 in \cite{Liu2010a} to 1080 in \cite{Vendramin2010}. As results may only be applicable to problems featuring similar properties to the datasets used in a study, the use of a diverse collection of datasets is very important. The most common properties featured in previous studies have been: number of clusters \cite{Vendramin2010,Liu2010a,Guerra2012,Arbelaitz2013,Li2016a,Nguyen2020,Karanikola2021,Jose-Garcia2021}, number of dimensions \cite{Vendramin2010,Guerra2012,Arbelaitz2013,Li2016a,Nguyen2020,Karanikola2021,Jose-Garcia2021}, overlap \cite{Arbelaitz2013,Jose-Garcia2021,Nguyen2020}, cluster density \cite{Arbelaitz2013,Liu2010a,Nguyen2020}, cluster balance \cite{Vendramin2010,Liu2010a,Karanikola2021}, noise \cite{Liu2010a,Arbelaitz2013}, and cluster shape/distribution \cite{Nguyen2020,Jose-Garcia2021}. The range of values considered for each attribute also varies between studies; for example, the number of clusters used in these studies ranges from between 2 and 8 clusters per dataset in \cite{Arbelaitz2013} to between 2 and 17 clusters per dataset in \cite{Karanikola2021}.

Additionally, many studies include real-world datasets in their comparisons, however, as clustering problems do not have a known ground truth, real-world data typically consists of classification data with the class labels used as a ground truth. This can be a flawed methodology as it has been noted that clusters within a dataset do not necessarily align with classification labels \cite{Frber2010OnUC, VonLuxburg2012}. There is also a significant amount of overlap in the real-world data used in studies of clustering validity indexes. The UCI repository \cite{UCI} has been used as a source of datasets in most studies involving real-world data \cite{Arbelaitz2013,Nguyen2020,Karanikola2021,Jose-Garcia2021}, with only a few studies \cite{Li2016a,Nguyen2020} using real-world datasets from alternative sources. This makes results more comparable across different studies, but it is a concern as only a limited pool of data has been used across the majority of studies, limiting the generalisation of the results.

\section{Experimental Methodology} \label{methods}



\subsection{Internal Relative Indexes} \label{IntInd}

The selection of internal validity indexes within this paper is an extension of the list used in \cite{Vendramin2010}, including all optimisation-like criteria assessed in that paper, with three exceptions, namely: (a) The Gamma index has been replaced with AUCC, which is a computationally much faster algorithmic implementation that is fully equivalent to the Gamma index \cite{jaskowiak_area_2022}; (b) Only one variant of the Dunn index has been included for clarity and compactness, since Dunn's variants have shown poor performance both in \cite{Vendramin2010} as well as in our experiments; (c) Only one variant of the Silhouette Width Criterion has been used due to similar performance observed between the variants. Alongside the previously studied internal validity indexes, this paper also assesses an additional 14 indexes, totalling 26 internal validity indexes studied within this paper. 

This includes a variety of density-based internal validity criteria, a type of index that has not been included in other studies such as \cite{Vendramin2010}. Although previous studies such as \cite{Vendramin2010} and \cite{Arbelaitz2013} appear to include more indexes than ours, they include multiple similar versions of fewer indexes, such as several variants of the Dunn index, rather than many unique indexes as we do here.

It should be noted that other internal validity indexes exist that are tied to specific clustering algorithms, such as e.g. the Gap Statistic \cite{50274103-6d7b-3be0-9c14-af318bc52b3f} and the index of Sugar and James \cite{doi:10.1198/016214503000000666} for K-Means variants, or Bayesian Information Criterion (BIC) \cite{c4048c8f-6ca9-3965-96a3-653ab8996955,10.1093/comjnl/41.8.578} and Integrated Complete-data Likelihood (ICL) \cite{Bouveyron_Celeux_Murphy_Raftery_2019} for EM-GMM and alike. While these indexes may be effective within their respective algorithms, they do not fit the requirements of this paper as we intend to investigate general-purpose indexes rather than those limited in scope. Many such indexes, including the Gap statistic, also fall into the category of a difference-like criterion, which are unable to evaluate in relative terms a given arbitrary set of candidate clustering solutions, potentially from different algorithms. As previously mentioned, these are not within the scope of our paper. 

Some internal validity indexes can be noted to have implicit relationships with specific clustering algorithms. For example, the Calinski–Harabasz Index and the K-means algorithm both rely on squared Euclidean distances to measure within-cluster variance. While it may be theoretically preferable to pair indices and clustering methods that share assumptions and conceptual underpinnings, such as applying the Silhouette index with the same distance measure used in clustering (e.g., Mahalanobis distance for EM-GMMs), this is not a strict constraint commonly adopted in practice. This study features many indexes that cannot be adjusted to particular clustering setups, such as the similarity measure of choice, so for the sake of consistency we have not attempted to pair certain indexes with preferred algorithms and/or similarity measures. In real-world applications though, this is an aspect practitioners should bear in mind.


There are two primary approaches used by internal validation measures for quantifying clustering quality. The first is traditional indexes, which measure and compare the compactness and separation of clusters, where compactness refers to how close or tightly packed the observations within a cluster are, while separation refers to the distance between separate clusters within the space. These properties, however, do not have a single formulation, with both being measured and compared differently by each validation measure, giving rise to a multitude of different indexes. Traditional indexes are best suited to identifying globular clusters, with many of these formulations relying on the assumption that the clusters of a problem are globular, forming roughly spherical or ``blob'' like shapes distributed around a central region or centroid, rather than elongated or irregular shapes. 
The second approach to clustering validation is a density-based approach based on the principle of clusters corresponding to denser regions of observations in the space, surrounded by sparser regions. Such indexes are, in theory, expected to be better suited to detecting clusters that are not globular while also being more robust to noise.

\begin{table}[t!]
\small
\begin{tabular}{ |m{3.8cm}|m{1.65cm}|m{1.65cm}| }
 \hline
 Index & Maximum/ Minimum & Approach \\ [0.5ex] 
 \toprule\hline
 Calinski–Harabasz/Variance Ratio Criterion (VRC)\tablefootnote{\label{imp1}Code provided by \cite{JoseGarciaGF23}} \cite{doi:10.1080/03610927408827101} & Max & Sep/Comp \\  
 \hline
 Davies–Bouldin (DB)\footref{imp1} \cite{4766909} & Min & Sep/Comp \\  
 \hline
  Dunn\footref{imp1} \cite{doi:10.1080/01969727408546059} & Max & Sep/Comp \\  
 \hline
  Silhouette Width Criterion \tablefootnote{Code provided by \cite{cluster2022}} \cite{Rousseeuw1987,LeonardKaufman1990} & Max & Sep/Comp \\  
 \hline
  Pakhira, Bandyopadhyay and Maulik Index (PBM)\footref{imp1} \cite{PAKHIRA2004487} & Max & Sep/Comp \\  
 \hline
  C-Index\footref{imp1} \cite{Hubert1976} & Min & Sep/Comp \\  
 \hline
  Point-Biserial\tablefootnote{\label{imp3}Code from https://doi.org/10.6084/m9.figshare.25670751} \cite{Milligan1981,Milligan1985a} & Max & Sep/Comp \\  
 \hline
 Density Based Clustering Validation (DBCV)\footref{imp1} \cite{Moulavi2014a} & Max & Density Based\\  
 \hline
  SSDD\footref{imp1} \cite{Liang2020} & Min & Density Based \\  
 \hline
   Area Under the Curve for Clustering (AUCC)\tablefootnote{Code from https://github.com/pajaskowiak/clusterConfusion} (gamma) \cite{jaskowiak_area_2022} & Max & Sep/Comp\\  
 \hline
   Local Cores-Based Cluster Validity (LCCV)\footref{imp1} \cite{Cheng2019} & Max & Density Based \\  
 \hline
   CS Index\footref{imp1} \cite{Chou2004} & Min & Sep/Comp \\  
 \hline
   SV Index\footref{imp1} \cite{ZALIK2011221} & Max & Sep/Comp \\  
 \hline
   Wemmert-Gancarski Index\tablefootnote{\label{imp2}Code provided by \cite{Crit2018} } \cite{WEMMERT2000}  & Max & Sep/Comp  \\  
 \hline
   WB Index\footref{imp1} \cite{10.1007/978-3-642-04921-7_32} & Min & Sep/Comp \\  
 \hline
   Cluster Validity Based on Density-Involved Distance (CVDD)\footref{imp1} \cite{Hu2019} & Max & Density Based \\  
 \hline
   Composed Density Between and Within Clusters (CDbw)\tablefootnote{Code provided by \cite{fpc2023} } \cite{Halkidi2008} & Max & Mixed (Sep/Comp and Density Based) \\  
 \hline
   S\_Dbw\footref{imp1} \cite{989517} & Min & Mixed (Sep/Comp and Density Based) \\  
 \hline
   Clustering Validation Based on Nearest Neighbours (CVNN)\footref{imp1} \cite{Xiong2019} & Min & Mixed (Sep/Comp and Density Based)\\  
 \hline
   SD Index\footref{imp2} \cite{Halkidi2000} & Min & Sep/Comp \\  
 \hline
  Global Overlap Index Relaxed (Grex)\footref{imp3} \cite{Iglesias2020} & Max & Density Based \\
  \hline
  Global Overlap Index Strict (Gstr)\footref{imp3} \cite{Iglesias2020} & Max & Density Based \\
 \hline
 G(+) Index\footref{imp2} \cite{doi:10.1146/annurev.es.05.110174.000533,Milligan1981} & Min & Sep/Comp \\  
 \hline
 Tau\footref{imp2} \cite{doi:10.1146/annurev.es.05.110174.000533,Milligan1981} & Max & Sep/Comp \\  
 \hline
  Ratkowsky-Lance\footref{imp2} ($C \over{\sqrt{K}}$) \cite{10.2307/2474421,Krzanowski1988ACF} & Max & Sep/Comp \\  
 \hline
 Xie-Beni Index\footref{imp1} \cite{85677} & Min & Sep/Comp \\  
 \hline
\end{tabular}
\caption{\label{IndexTable} Internal validity indexes used within this study and their characterisation into major types, namely: Minimisation (Min) or Maximisation (Max) types; Separation/Compactness (Sep/Comp), Density-Based, or Mixed. Details of each index can be found within the corresponding references. A compiled summary of the formulation and computational complexity of many of these indexes can be found in \cite{Vendramin2010}.}
\end{table}

Certain internal validity indexes share properties from both of these categories, such as S\_Dbw, which uses a density-based measure of compactness combined with prototype-based quantities that don't follow the density-based principles and assumptions. Table \ref{IndexTable} includes all internal validity indexes considered in this study, with note of which indexes are minimisation or maximisation measures, as well as their approach.

Most internal validity indexes are not capable of adjusting for observations being identified as noise or not assigned to any cluster by a clustering algorithm, with the exception of DBCV \cite{Moulavi2014a}. Since clustering algorithms such as HDBSCAN* may produce partitions containing observations labelled as noise, it is necessary to adjust most indexes in order to accurately evaluate the produced partitions. It is important to stress that noise observations are \emph{not} considered to be clustered together into a ``noise cluster'' by the corresponding algorithms, and should not be interpreted as such by an index. Although probabilistic methods like EM-GMMs can be configured to model noise as an additional uniform component treated as a separate statistical mechanism, none of the selected validity indexes can handle a background noise distribution as a cluster without causing significant detriment to the indexes ability to assess the quality of a partition.
This is because such ``noise clusters'' violate the assumptions underlying these measures: they are neither compact and well-separated, as assumed by traditional measures, nor dense regions, as assumed by density-based indexes.
On the other hand, interpreting each noise observation as a cluster on its own (a singleton) can also be shown to produce unexpected and undesired evaluation results, e.g., by critically distorting measures of cluster separation. 

Instead, the alternative adopted here is the same adjustment used within DBCV, which will be applied to all internal indexes in this paper. The basic principle is to first assess the partition excluding the noise, and then rescale the resulting value of the index by the percentage of data successfully partitioned into clusters using Equation \ref{eq:Noise_Adjustment}, where $N$ is the number of points in the dataset and $N_{noise}$ is the number of points identified as noise by a clustering algorithm. In other words, an index will penalise a partition under assessment proportionally to the fraction of data left unclustered in that partition, which will discourage trivial candidate solutions such as, e.g., clustering only very dense data regions while leaving most of the data unclustered as noise. Indexes ranging within a negative scale will be converted to positive before the noise adjustment in Equation \ref{eq:Noise_Adjustment} is applied. For maximisation (respectively minimisation) indexes which may produce both positive and negative values, the adjustment will not be applied to negative (resp. positive) values as it would improve rather than penalize such values. 

\begin{equation} \label{eq:Noise_Adjustment}
    Index_{adjusted} = Index \cdot \frac{N-N_{noise}}{N}
\end{equation}

\subsection{External Indexes} \label{ExtInd}

External validity indexes are utilised in part of the experiments in this paper where external measures of clustering validity are used to evaluate/rank cluster quality with respect to a ground truth, and the results are then compared to those produced by internal validity indexes by quantifying the agreement between the two. The choice of external index can thus significantly impact the conclusions as it serves as a reference for the assessment of the internal indexes. Our choices are supported by extensive literature, as detailed next.

There exists a wide range of external validity indexes that have been studied in the literature --- see e.g. \cite{Pfitzner2009,Arinik2021}. These papers note various properties, behaviours and potential biases of external indexes, for example, it has been noted in \cite{Arinik2021} that there is no consensus about whether external indexes should be (in)sensitive to cluster size or (im)balance, i.e., treating all clusters as equally important regardless of the number of observations within. In this sense, it is important to understand that external validity indexes do not produce an ultimate, absolute measure of partition quality, but rather, a specific criterion based on each index's assumptions about the clustering problem. As such, they should not be seen as an indisputable, perfect unique measure of partition quality.

The study in \cite{Pfitzner2009} features an extensive assessment of 56 external validity indexes, which are classified into two main categories: pair counting measures, such as the Adjusted Rand index (ARI), and information theoretic measures, such as Normalised Mutual Information. This paper outlined 6 desirable properties of external validity indexes in order to categorise indexes based on their behaviours. 
It is however noted in \cite{Pfitzner2009} that the interpretation of these properties may differ depending on the use case. An example is the property that the fall-off of the similarity measure should match the intuition of decreasing similarity: in some cases, this may mean a linear relationship between the fall-off of an index and the differences between partitions, or alternatively a non-linear relationship where the fall-off increases significantly as the differences increase. All indexes were categorised into 6 groups based on their evaluation behaviours and which of the studied properties they met. It was shown that the type of index was not the primary identifier of performance or quality. This assessment highlighted several indexes as fulfilling more desirable properties compared to others, namely: several variants of Normalised Mutual Information, the Powers measure, and the Lopez Rajski's measure. 

The study in \cite{Arinik2021} adopted a new framework for assessing the effect of various properties of external indexes previously highlighted in the literature, and compared a small set of external indexes using this framework. The results were shown to be consistent with previous studies such as \cite{Pfitzner2009}.

In \cite{Lei2017}, biases in external indexes as trends determined by the ground truth of a partition were investigated for 26 indexes. The indexes highlighted as exhibiting such trends were: the Rand index, Mirkin index, Hubert index, Gower and Legendre index, and the Rogers and Tanimoto index. However, only pair counting indexes were investigated in this study. The study \cite{Vinh2010}, in turn, assessed information theoretic external indexes. Trends associated with increasing number of clusters were noted for all versions of Normalised Mutual Information (NMI); Normalised Information Distance (NID) was proposed as alternative.

Based on these papers, six external indexes were selected for our study, as outlined below. These indexes aimed at fulfilling various properties in an attempt to limit any potential biases and balance differences in behaviour associated with each single index:

\vspace*{2mm}
{\bf Pair Matching}

\begin{itemize}
\item Jaccard \cite{Jaccard1912}
\item Sokal and Sneath 3 \cite{10.2307/1217562}
\item Adjusted Rand Index (ARI) \cite{Hubert1985}
\end{itemize}

\vspace*{2mm}
{\bf Information Theory}

\begin{itemize}
\item Normalised Mutual Information (NMI) \cite{Strehl2002}
\item Powers \cite{powers2007expected}
\item Normalised Information Distance (NID) \cite{1362909}.
\label{ExternalIndexTable}
\end{itemize}

\begin{figure}[h]
 \centering
 \includegraphics[width=\linewidth]{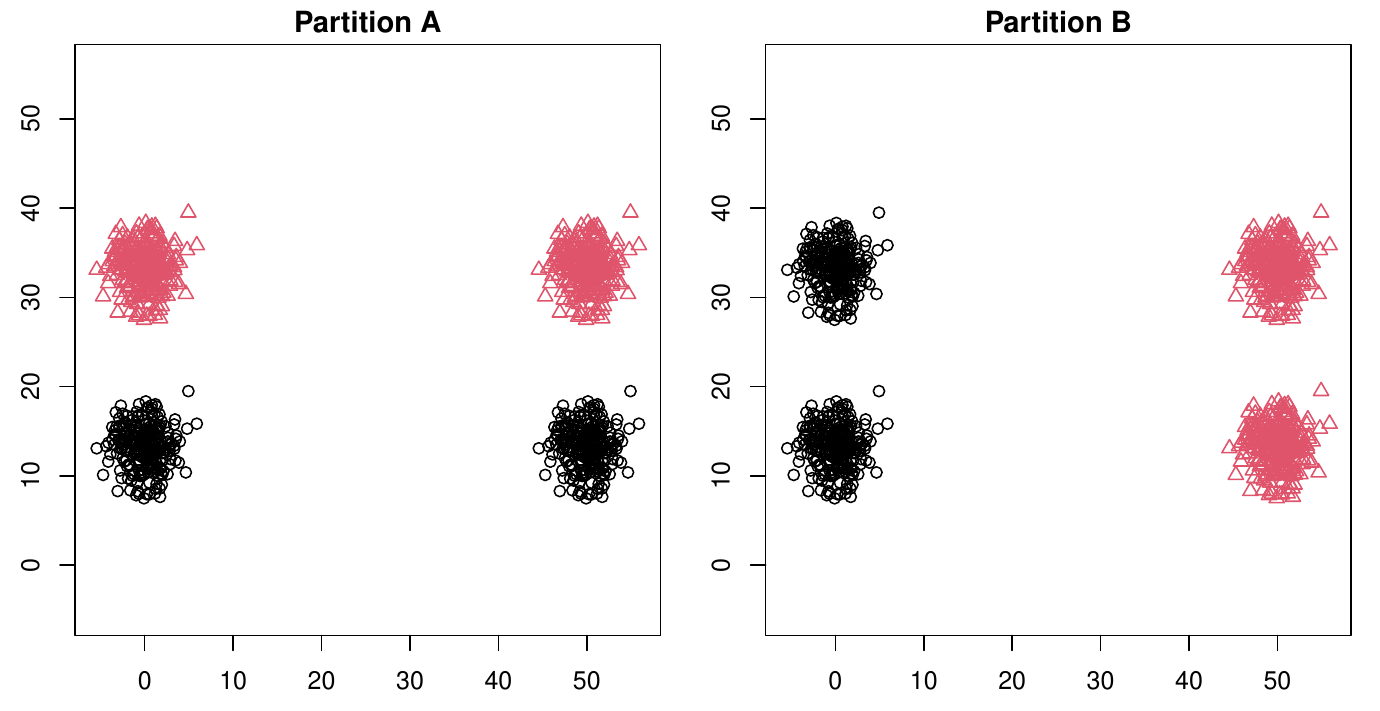}
 \caption{\label{fig:ExtIndex} Two partitions of a dataset with 4 ground-truth clusters. Both partitions (A and B) have identical values of the Adjusted Rand Index: ARI = 0.51. }
\end{figure}

All of the aforementioned external indexes assess the quality of a given partition by comparing its cluster labels against the cluster labels of a ground truth. It is important to note here that the geometry or distances between clusters of the clustering problem is not considered in these indexes. Although this is generally not an issue and potentially desirable as to not exhibit similar flaws to internal validity indexes, there are several scenarios where external validity indexes either fail to distinguish between better and worse partitions or select the worse partition due to geometric differences. This is of particular importance when considering partitions with the same number of clusters as they may have similar numbers of points ``correctly'' clustered, however, with significantly different geometric properties. An example can be seen in Figure \ref{fig:ExtIndex}, where external indexes cannot distinguish between partitions of significantly different quality. 

Few external indexes exist that are capable of assessing the quality of a partition with respect to both the geometry and ground-truth labels. CDistance \cite{Coen2010} is one index capable of determining the quality of partitions taking geometric differences of the ground-truth clusters into account in addition to their labels, however, it suffers from two primary issues, namely: it has a high computational complexity of $O(n^3)$ and it tends to degrade in accuracy when comparing partitions of varied numbers of clusters, as observed in our preliminary analysis. As a result, this will only be employed for the scenarios described in Section \ref{Method2}, which will involve experiments with a fixed number of clusters.

\subsection{Correlation Measures} \label{subsec:correlation}

The methodology presented in \cite{Vendramin2010} consisted of using both external and internal validity indexes to evaluate the quality of a diverse collection of clustering partitions for a dataset and then calculating the Pearson correlation between the two, with a high correlation coefficient indicating good performance of the internal index. The same idea will be adopted for part of the experiments presented in this paper where external indexes are used, however, rather than a single external index, each partition will be ranked by a selection of external indexes and the summed rank will be adopted as an aggregated measure of partition quality with respect to a reference ground truth. The practice of combining multiple criteria through summation is a well-known and commonly used method in Multi-Criteria Decision Making \cite{Triantaphyllou2000}, and it is implemented to mitigate isolated differences in behaviour of an external index and limit the extent to which they can individually influence the results. Additionally, following other studies that also built upon the methodological procedure introduced in \cite{Vendramin2010} (e.g. \cite{Rabbany2012,Nguyen2020}), the Spearman correlation will be used in place of the Pearson correlation since the former is more appropriate for measuring monotonic relationships, particularly in the presence of potential differences in scale and non-linear relationships.

When all produced partitions do not seem to contain meaningful structure of the clustering problem, it is not reasonable to expect an internal validity index to accurately distinguish between poor quality solutions only. For this reason, any collection of candidate solutions where there isn't at least one solution that reaches an ARI of 0.6 or greater are removed. These cases are most likely due to a failure of all clustering algorithms in producing even a small subset of high-quality solutions, and their inclusion would negatively impact indexes by assessing them in scenarios where they cannot be expected to perform correctly, so ranks and their correlation tend to become highly susceptible to randomness. Assuming cases in which clustering solutions may actually contain reasonable structure, yet still lower ARI values have been produced, this would then mean the external validity index(es) may not be acting in an accurate or reliable manner. Either way, such results may be misleading due to improper performance of the clustering algorithms and/or external indexes. Overall, performance of all internal indexes are similarly impacted by the removal of these cases. Leaving these in could, however, skew specific results regarding trends in categories of datasets, such as the relationship to dimensionality and other properties, due to the proportion of such cases being higher for dataset collections with more difficult properties.

The adoption of Spearman correlation, which measures monotonic relationships, mitigates the impact of non-linear relations in magnitude and scale of external and internal indexes. Yet, as part of the contributions in this study, our experiments will reveal that the relationship between external and internal validity indexes may actually exhibit complex non-linear behaviours, often in the form of multiple discontinuous regions, which cannot be properly captured by a single measurement of correlation. These regions can generally be distinguished by whether candidate partitions contain greater or fewer clusters as compared to the best partition according to the external validity index. This translates into whether partitions contain greater or fewer clusters compared to the ground-truth partition when the collection of partitions is such that their similarity to the ground truth, according to the external index, aligns with their similarity in terms of the number of clusters. This typically occurs when the partitions produced by clustering algorithms are able to produce solutions that are similar to the ground truth. In this paper, we aim to properly capture this type of non-linear relationship between external and internal validity indexes by taking three different correlation measurements rather than a single one, namely: (a) across all partitions, (b) across partitions with fewer clusters than the optimal partition ($k < k_O$), and (c) across partitions with more clusters than the optimal partition ($k > k_O$). By measuring the difference between these correlations we can identify circumstances where a significant non-linear relationship has been produced due to aforementioned factors.

An additional issue is that correlation is subject to bias associated with the range of the number of clusters across the collection of partitions adopted. As the high end of the range is increased further beyond the number of clusters in the ground truth, partitions with too many clusters start dominating the results and the indexes are increasingly tested on their ability to discern between various poor solutions only, rather than also being tested on their ability to discriminate among good solutions as well as between good and poor solutions. This can also further create a bias towards indexes that are better at assessing solutions with more clusters than the ground truth, especially for datasets with few ground-truth clusters as there is a lower limit on the number of clusters that can be produced with fewer clusters. This will be further discussed in Section \ref{method1} and is aimed to be alleviated by properly limiting the upper bound on the number of clusters to ensure that a balanced mixture of good, moderate and poor clustering solutions are included in the collection of partitions used in the experiments. Additionally, a separate subcategory of experiments will be performed where the number of clusters is fixed across the whole collection of partitions, which are expected to vary in terms of quality as measured by their similarity to the ground truth, despite having the same number of clusters. Both these setups can mitigate potential correlation biases associated with the number of clusters.

Finally, despite the aforementioned strategies to address or mitigate issues associated with the use of correlation values between external and internal indexes as a measure of quality of the latter, there are still scenarios where their underlying non-linear relationships are too intricate to be captured by a correlation-based approach. As part of our contributions, we also propose to visually assess scatter plots and identify prominent or recurring relationships as potential patterns that may not be captured by correlation alone, determine their possible cause(s), and evaluate their effect on the performance of the internal indexes assessed.

\FloatBarrier

\subsection{Statistical Tests} \label{stat}

This paper will extend statistical testing as used in previous papers further, by featuring a more comprehensive examination of the correlations produced. This will involve two separate statistical tests, both performed at a 5\% significance level. The first test is similar to what was done in \cite{Vendramin2010,Arbelaitz2013}. Due to the data being paired as it is a repeated measurement on the same datasets and partitions, we use a pairwise Willcoxon test with Bonferroni correction to determine which internal clustering validity indexes (if any) outperform others by a statistically significant margin.

The second set of statistical tests are performed to compare how different properties of the datasets and partitions affect the performance of each index. For the properties of number of clusters, number of dimensions, overlap and imbalance, the Spearman correlation between these properties and the performance is measured and tested for statistical significance to determine the impact and direction. For the property of noise, a Kruskal-Wallis test is utilised to determine if the presence of noise has a significant impact on performance. Cluster compactness is also tested using a Kruskal-Wallis test comparing the two levels (0.1 and 0.8) defined in Section \ref{subsec:Data}. The aim of this testing is to determine common behaviours between indexes and develop further understanding of their performance.

Non-parametric tests are utilised for all statistical testing due to the distributions of correlations being significantly skewed and not containing equal variance between groups. Additionally, in this regard, median values will be reported alongside mean values for more accurate measures of index performance, an aspect not captured in \cite{Vendramin2010}, which only reported mean correlations. Although as medians are insensitive to the changing of up to half of the results, the mean performance is still preferable for overall comparison.

\subsection{Clustering algorithms} \label{Algs}

As the performance of internal validity indexes may be impacted by the different types of partitions produced by different clustering algorithms, it is important to utilise a representative selection of algorithms when producing clustering results for evaluation. A variety of popular clustering algorithms are described in Table \ref{Algorithm}, which have been selected to provide diversity of results in our comparative study. In addition to the classic K-Means and linkage-based hierarchical clustering algorithms, we have also included HDBSCAN*, Expectation Maximisation with Gaussian Mixture Models (EM-GMM), and Spectral Clustering.

HDBSCAN* \cite{Campello2013,Campello2015} is a popular density-based hierarchical clustering algorithm capable of identifying non-globular clusters possibly with highly varying densities as well as noise within data. The \mbox{HDBSCAN*} implementation within the \texttt{HDBSCAN} package in Python \cite{McInnes2017} is used (with \texttt{min\_samples = min\_cluster\_size = 4}, as originally adopted in \cite{Campello2013}) to construct the hierarchical density-based tree, from which global, horizontal cuts are taken at each density level to produce solutions with unique numbers of clusters. Due to our global cuts being performed in the condensed version of the tree produced by HDBSCAN*, the resulting collection of candidate clustering solutions may not comprise every number of clusters in the given range.

Expectation Maximisation is a traditional model-based clustering algorithm which produces clusters using one of several different models based on mixtures of Gaussians (so-called Gaussian Mixture Models --- GMMs). The \texttt{Mclust} package within R is used to automatically determine the best model for each dataset. As previously mentioned, the common practice for EM-GMM is to use a model-based validity method, such as the Bayesian Information Criterion (BIC), which are not tested within this study. As such, the performance of the measures studied here should not be taken as a recommendation over these methods. Also, EM-GMM can be used as a fuzzy clustering algorithm, as each observation is assigned a probability of membership to each cluster. This aspect has, however, not been investigated within this study, which focuses solely on its use as a hard clustering algorithm. The inclusion of EM-GMM is instead motivated by the desire to introduce greater diversity in the generated partitions and to incorporate widely used clustering algorithms.

Spectral Clustering performs clustering on the eigenvalues of a similarity-based matrix representation of the data in order to reduce the dimensionality of a clustering problem. We adopt the implementation of Spectral Clustering in the \texttt{sklearn} Python package \cite{scikit-learn}, which performs the final clustering with K-Means.

Each of these clustering algorithms is capable of producing different solutions to clustering problems, which may integrate better or worse with specific indexes and their underlying assumptions on different datasets.

\begin{table}
\small
\begin{tabular}{ |m{1.9cm}|m{1.2cm}|m{2cm}|m{1.4cm}| }
 \hline
 Algorithm & Software & Properties & Hyper-Parameters\\ 
  \hline\hline
  K-Means \cite{MacQueen1967SomeMF} & R \cite{R} & Partitioning & $k$\\  
 \hline
 Single Linkage \cite{LeonardKaufman1990} & R \cite{R} & Hierarchical & NA\\  
 \hline
 Average Linkage \cite{LeonardKaufman1990} & R \cite{R} & Hierarchical & NA\\  
 \hline
  Complete Linkage \cite{LeonardKaufman1990} & R \cite{R} & Hierarchical & NA\\  
 \hline
  Ward Linkage \cite{LeonardKaufman1990} & R \cite{R} & Hierarchical& NA \\  
 \hline
 HDBSCAN* \cite{Campello2013,Campello2015} & Python (HDBSCAN) \cite{McInnes2017} & Density-Based, Hierarchical, Noise Detection & Min Samples, Min Cluster Size (optional)\\
 \hline
  EM-GMM \cite{doi:10.1198/016214502760047131} & R (Mclust) \cite{mclust,MclustPackage} & Model-Based, Probabilistic, Noise Detection\tablefootnote{Requires additional optional initialisation not implemented within this study.} & $k$, Covariance Model\\
 \hline
 Spectral Clustering \cite{Yu2003} & Python (Scikit-learn) \cite{scikit-learn} & Partitioning, Dimensionality Reduction & $k$ \\
 \hline
\end{tabular}
\caption{\label{Algorithm} Clustering algorithms used with their implementations and Hyper-Parameters.}
\end{table}

\subsection{Data} \label{subsec:Data}

\begin{figure}[h]
 \centering
 \includegraphics[width=\linewidth]{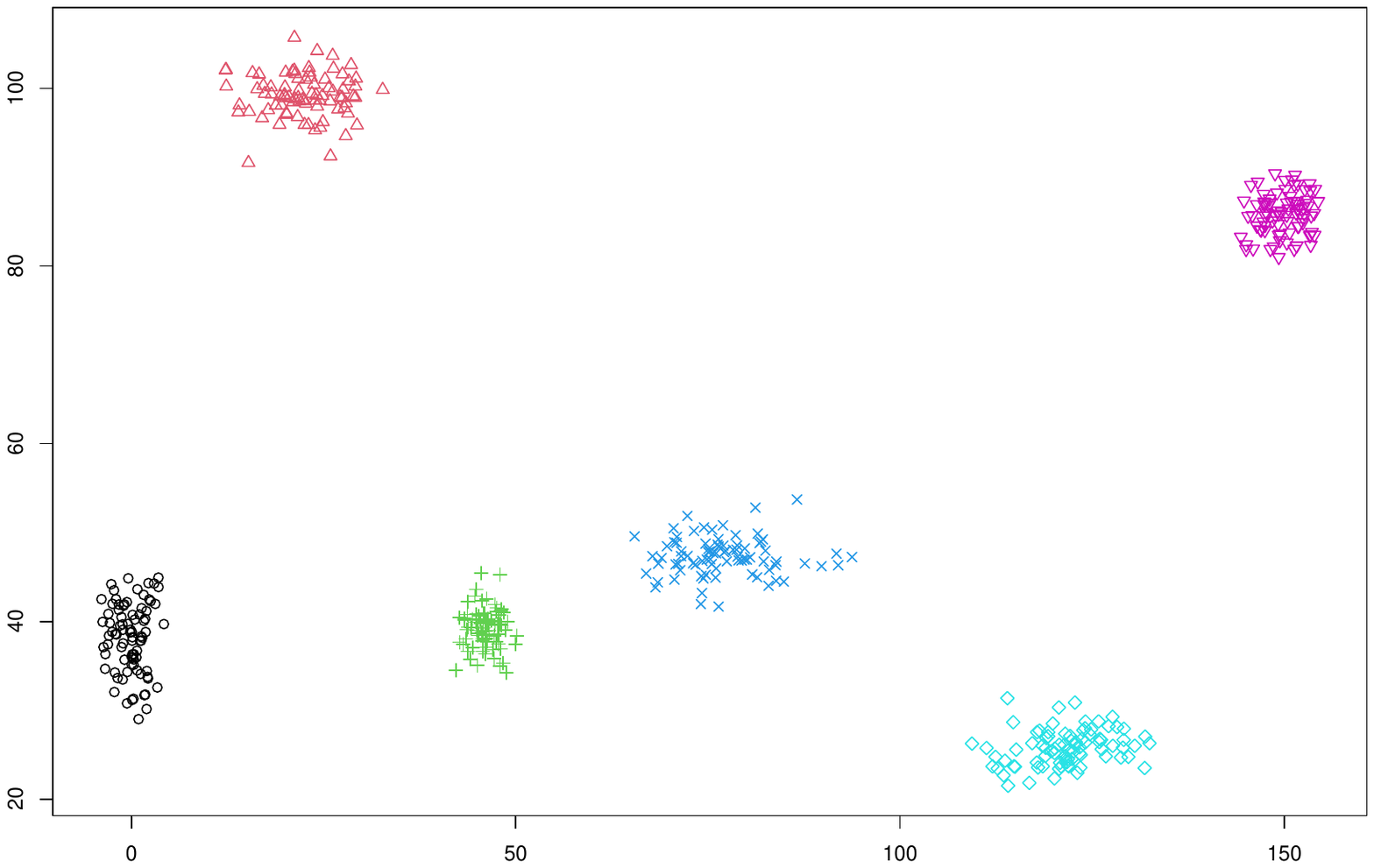}
 \caption{\label{fig:Type1} Two dimensional slice of a Type 1 dataset with 6 clusters and 4 dimensions.}
\end{figure}

\begin{figure}[h]
 \centering
 \includegraphics[width=\linewidth]{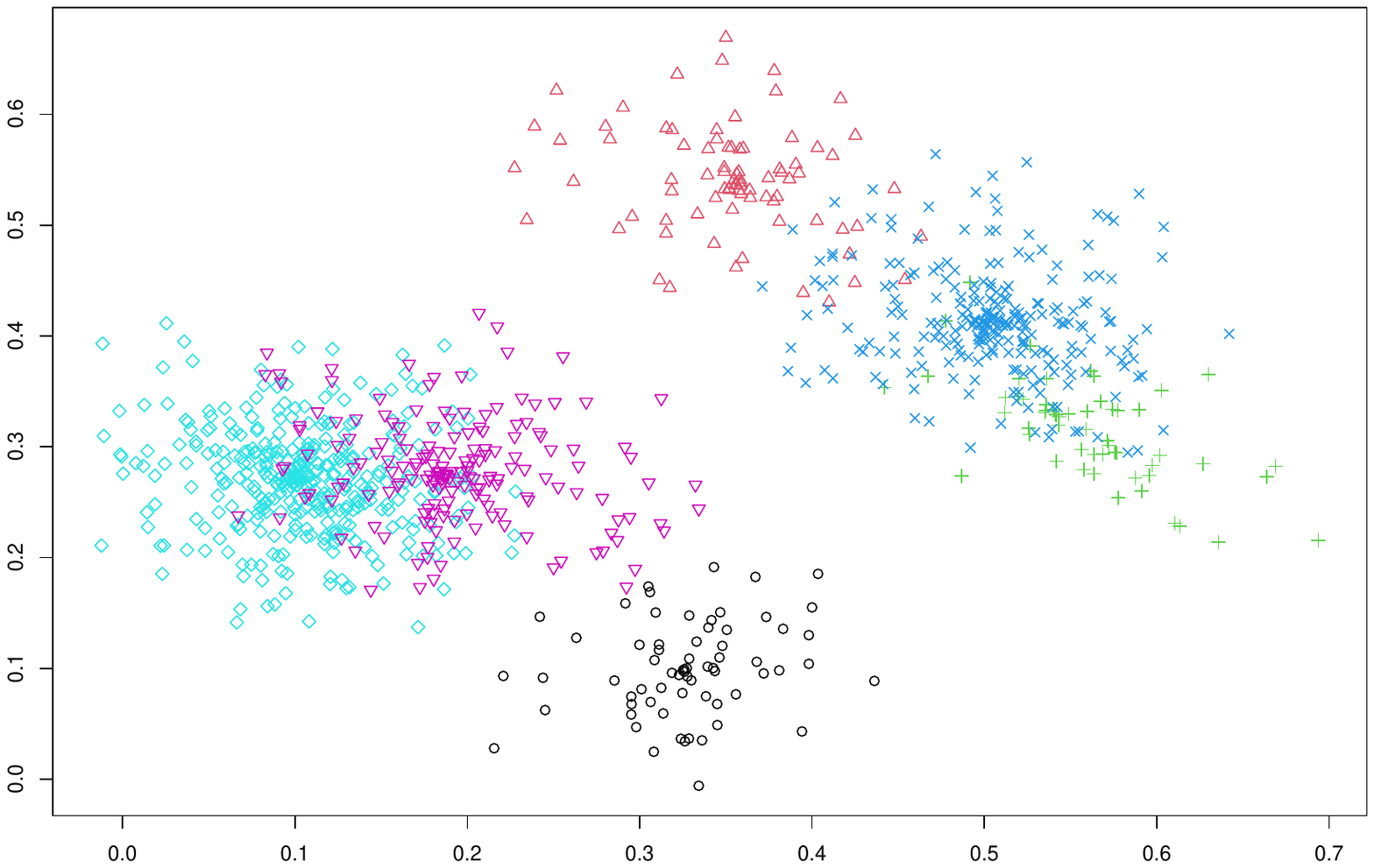}
 \caption{\label{fig:Type2} Two dimensional slice of a Type 2 data with 6 clusters and 4 dimensions.}
\end{figure}

This paper will utilise synthetic data due to the requirements of a known ground truth and flexibility in producing datasets with desired properties, which is not possible with real-world data.\footnote{All datasets, results, software used and scripts used to perform this study will be available upon publication at https://doi.org/10.6084/m9.figshare.25670751} Two different groups of synthetic data will be assessed, referred to here as Type 1 and Type 2 data: the first is a basic collection of datasets used in previous studies of internal validity indexes, whereas the second is an updated collection featuring more complex clustering problems. Example datasets of Type 1 and 2 are illustrated in Figures \ref{fig:Type1} and \ref{fig:Type2}, respectively. In the updated (Type 2) datasets we have aimed to produce more diverse clustering problems with challenging properties such as cluster overlap and outlier points, in contrast to more separated and compact clusters of the Type 1 data.

Although certain internal validity indexes (e.g., Silhouette) can operate on general dissimilarity matrices, others (e.g., VRC) are specifically designed for Euclidean data only. For this reason, both Type 1 and Type 2 dataset collections consist of Euclidean data. Their ground-truth partitions correspond to the known groupings based on the underlying statistical distributions used to generate the observations. This serves as the reference partition for use with the external validity indexes, with the aim of determining each internal index's ability to recover these distributions. 

{\bf Type 1 Data} are the same datasets for the second experiment in \cite{Vendramin2010} as well as in follow-up studies. This collection is included here in order to compare the results of our updated methodology against the original methodology from \cite{Vendramin2010} and observe to which extent the limitations of the original methodology affect the results and conclusions.
This collection consists of 972 datasets containing well-separated globular clusters. Within the datasets there are six different numbers of clusters $k^*\in \{2,4,6,12,14,16\}$, six different dimensions $D \in \{2,3,4,22,23,24\}$ and three levels of balance: 
(i) balanced clusters, (ii) one cluster contains 10\% of the observations whereas the rest is evenly distributed across the other clusters, and (iii) one cluster contains either 20\% or 60\% of the observations (depending on whether $k^*\ge12$ or not, respectively) whereas the rest is evenly distributed across the other clusters. For each combination of number of clusters, dimensionality and (im)balance, nine datasets were produced to generate a total of 972 unique datasets.

{\bf Type 2 Data} consists of a new collection featuring 16177 datasets with a larger and more diverse set of properties than the Type 1 collection. Major limitations of the Type 1 collection have been targeted, including the following: Type 1 data contain compact and well-separated globular clusters, their levels of balance, granularity, and dimensionality are very limited, and noise/outliers are not considered.
In order to produce results more representative of potential practical problems, in the Type 2 collection, new datasets have been generated with a wider range of properties and their values. Due to its versatility, the Multidimensional Dataset Generator for Clustering (MDCGen) \cite{Zseby2019} was adopted for this task.

The primary properties considered in the Type 2 collection are dataset size, number of dimensions, number of clusters, overlap, compactness, distribution/shape, (im)balance, and noise, discussed in more detail below.

The \textbf{number of clusters} produced ranges from 2 to 50 clusters as $k^* \in \{2,4,6,8,10,15,20,30,50\}$. Whenever a binary categorization is to be considered, datasets with 10 or fewer clusters will be said to have few clusters, whereas datasets with more than 10 clusters will be said to have many clusters.

The \textbf{dimensionality} of the datasets produced ranges from 2 to 200 as $D \in \{2,4,6,8,10,15,25,50,80,120,200\}$. It should be noted that additional dimensions are informative dimensions which may not be fully representative of real-world clustering problems where dimensions may also consist of noise or weakly discriminative information. Supplementary information provided by additional informative dimensions may make it ``easier'' for clustering algorithms and internal validity indexes alike to discriminate between clusters, in comparison to datasets where uninformative dimensions may obscure the clusters. However, the clustering problems could still be harder due to other aspects as compared to lower dimensional problems, such as the diminishing contrast between distances (and densities), i.e., their relative differences becoming less pronounced as dimensionality increases. Note that whenever a binary categorization is to be considered, datasets with 25 or fewer dimensions will be deemed low dimensional, whereas datasets containing more than 25 dimensions will be deemed high dimensional.

The \textbf{dataset size} is varied to generate sparser and denser clustering problems. To that end, the number of observations for each dataset was determined based on the number of clusters. Specifically, the cluster sizes range within $\{20,100\}$, with a total cap of 1000 observations per dataset due to computational constraints.

\textbf{Cluster distribution} or \textbf{shape} refers to the distribution and shape of each individual cluster. Within this study, we focus only on globular cluster distributions, using three of the different cluster distributions provided within MDCGEN: Uniform, Gaussian, and Logistic.\footnote{Although this is an improvement on the existing literature, future studies could extend this list further with more diverse distributions e.g. in terms of skewness or extreme tail behaviour.} Additionally, the distributions are set to be radial-based, meaning the points of a cluster are distributed around the center of each cluster following the specified distribution. 


\textbf{Imbalanced} clusters were included in \cite{Vendramin2010}, however, only three levels of cluster (im)balance were used, as described for the Type 1 data collection above. One primary issue with this approach is that, effectively, balance becomes dependent on the number of clusters. For instance, for few clusters one cluster containing 10\% of the data is a small cluster, while for more than 10 clusters it is a larger than average cluster. This makes an analysis of how cluster balance affects the performance of internal validity indexes difficult as the effective balance is not consistent across datasets with different numbers of clusters. In order to mitigate this issue, an alternative measure of cluster balance will be used to analyse the effective balance of datasets in both Type 1 and Type 2 collections, as follows \cite{Nie2010}: 

\begin{equation} \label{eq:imbalance}
    Balance = \frac{N_{max} - N_{min}}{N_{min}} 
\end{equation}

\noindent where $N_{Max}$ and $N_{Min}$ are the number of points in the largest cluster and in the smallest cluster, respectively.

Equation \ref{eq:imbalance} does not attempt to fully capture into a single value the overall distribution of cluster sizes and balance, but it improves over the previous notion of balance originally used in \cite{Vendramin2010} by providing a more useful statistic that is close to 0 for balanced clusters and increases as effective imbalance increases.\footnote{Alternative measures such as Entropy exist which may alternatively be adopted to capture the distribution of cluster sizes.} The Type 2 data collection was generated with three different levels for cluster (im)balance, namely: (i) balanced clusters where all clusters contain the same number of observations; (ii) imbalanced clusters where a cluster cannot contain fewer than half of the average number of observations per cluster; and (iii) imbalanced clusters where a cluster cannot contain less than 10\% of the average number of observations per cluster. For the latter two levels, (ii) and (iii), the number of observations within each cluster was randomised within these constraints. As such, they ensure a wide variety of effective cluster imbalance levels in the generated datasets. Whenever a binary categorization is to be considered, a value of 0.5 or greater by Equation \ref{eq:imbalance} will be considered imbalanced.

\textbf{Overlap} describes the degree to which clusters are not cleanly separable, such that observations from different clusters fall into the same region of the data space in a way that makes the cluster boundaries ambiguous. This occurs between clusters in the ground truth when there is insufficient spatial separation between clusters such that observations belonging to different clusters are generated within the same region of the data space. Although from an algorithm's perspective this can be tackled using a soft (probabilistic or fuzzy) partitioning approach, where observations may be associated with multiple clusters, the ground-truth partitions within this study have been produced considering that observations solely belong to a single cluster based on their underlying generating distributions.

Spatial overlap is a property that can potentially be present in real-world datasets, however, all the datasets in the Type 1 collection from \cite{Vendramin2010} consist of compact and well-separated clusters only. Overlap will be included to some extent in our newly produced datasets to make the resulting Type 2 collection more representative of potential practical problems; however, since all the clustering validity indexes considered in this study as well as most clustering algorithms used in the experiments have been specifically designed to operate with hard (rather than fuzzy or probabilistic) clustering solutions, the level of overlap will be limited so it can be consistent with these choices and the related assumptions underpinning our study. The measure of overlap used in our analysis is based on the notion of nearest neighbours, as formulated in \cite{Shand2019}:

\begin{equation} \label{eq:overlap}
\begin{aligned}
    Overlap = 1 - \frac{1}{N} \sum_{i=1}^{N} 1_{C^i}(i_{nn}) \\
    1_{C^i}(i_{nn}) = 
    \begin{cases}
    1, i_{nn} \in C^i\\
    0, i_{nn} \notin C^i
    \end{cases}
\end{aligned}
\end{equation}
where $C^i$ is the cluster observation $i$ belongs to, and $i_{nn}$ is the nearest neighbour of observation $i$. 

The datasets within this study were limited to having between 0-10\% overlap according to this measure, with datasets exceeding this being discarded. Overlap mentioned within this paper will refer to overlap in the ground-truth partition, not within partitions produced by clustering algorithms. Whenever a binary categorization is to be considered, all datasets with overlap not equal to 0 are deemed to be overlapping.

\textbf{Compactness} refers to how close or tightly packed the observations within a cluster are, with compact clusters featuring a dense grouping of similar observations. This is defined in MDCGen through the variance component for the distributions that form the clusters. It has been specified here using three levels, namely, a variance of 0.1, 0.8 and random within [0, 1] for each cluster. This was done to ensure that datasets with clusters at the low and high ends of compactness, as well as at intermediate levels, were produced. This was also used to vary the level of potential overlap between clusters, since lower values of compactness (i.e., higher levels of variance) are more likely to allow for overlapping clusters when generated by MDCGen. Notice that whenever a binary categorization is to be considered, only datasets specified at the first two levels of compactness will be adopted, namely: datasets produced at level 0.1 will be deemed to contain compact clusters, whereas datasets produced at level 0.8 will be deemed to contain sparse clusters.

\textbf{Noise}, which was not included in the Type 1 data collection originally from \cite{Vendramin2010}, has now been included in our Type 2 collection. Type 2 data feature datasets where an additional 10\% points are included as background noise. Noise was added to datasets post-generation by producing observations uniformly distributed in each dimension of the data, bounded by the minimum and maximum values in each dimension.

\subsection{Evaluation Methodology} \label{Evals}

Three different evaluation scenarios will be considered, each of which utilises a different variant of a core evaluation procedure. These three evaluation methodologies have been designed to complement each other and capture different aspects of the performance of internal validity indexes. 

\subsubsection{Evaluation Scenario 1: Varied Number of Clusters} \label{method1}

The first evaluation methodology closely follows the processes used in \cite{Vendramin2010}, where partitions of a dataset are systematically produced with the number of clusters varying incrementally from $k=2$ through to $k = k_{max}$ using a variety of clustering algorithms. Following this procedure across multiple datasets, each index is then evaluated in terms of (i) the percentage of cases for which both the external and internal validity indexes select the same partition as optimal (the ``Top Pick'' partition) out of the candidate partitions corresponding to each combination of clustering algorithm and dataset; as well as in terms of (ii) the Spearman correlations between the rankings of such candidate partitions by the external and each internal validity index.

The Scenario 1 evaluation is aimed at determining the performance of internal validity indexes in a typical use case scenario where a clustering algorithm is applied without knowledge of the ground-truth number of clusters by producing partitions over a range of number of clusters, and an internal validity index is then used to determine the best partition(s). Different from \cite{Vendramin2010}, where all partitions of a dataset produced by the multiple different clustering algorithms are dealt with conjointly, we evaluate the set of partitions produced by each individual algorithm and dataset separately. This is required to include the clustering algorithm as a predictor in the analysis aimed to determine whether or to which extent the clustering algorithm producing the partitions under evaluation affects the performance of an internal validity index. 

As previously discussed in Section \ref{subsec:correlation}, the range of number of clusters in the partitions produced will be limited in order to prevent a long upper tail of partitions near-equivalent in quality (or lack thereof) and far away from the ground truth from dominating the analysis. A common rule-of-thumb used in previous studies (e.g. \cite{Vendramin2010, Arbelaitz2013}) is $k_{max} = \sqrt{N}$, however, this may not produce partitions with enough clusters for small datasets containing many ground-truth clusters, or it may produce a long tail of partitions with too many clusters for large datasets containing few clusters. As the datasets in this study contain between 2-50 clusters and 40-1000 observations, we adopt an alternative upper limit as $k_{max} = max(25,1.75k^*)$ in order to circumvent both issues. 

As described in Section \ref{subsec:correlation}, the correlation between the external and internal validity indexes is also calculated in two separate regions --- relating to partitions with under- and overestimated number of clusters --- and compared to the overall correlation (across both regions) to help identify potential non-linear relationships and assess their impact on the performance of each index. Detected non-linear relationships can then be further visually inspected and analysed.

Finally, mean and median correlations are computed across the datasets within each of the data sub-collections sharing common properties outlined in Section \ref{subsec:Data}, in order to understand how the impact of each property affects the rankings and performance of the internal indexes. In addition, statistical testing is performed following the description in Section \ref{stat} in order to determine which properties have statistically significant effects on the performance of each validity index.

\subsubsection{Evaluation Scenario 2: Fixed Number of Clusters} \label{Method2}

The second evaluation methodology adopted in this study is intended to assess the performance of internal validity indexes in scenarios where all clustering partitions under evaluation have the same fixed number of clusters. This was motivated from 
preliminary experiments showing that while some indexes may successfully select the best partition produced by a given algorithm when the number of clusters is varied (which often translates into correctly detecting the ground-truth number of clusters), they may not be so effective in discriminating between good and bad partitions that all share a given number of clusters in common --- e.g., partitions produced by different types of clustering algorithms with the number of clusters specified by a domain expert. 

In order to produce a diverse variety of partitions under this setting, all clustering algorithms listed in Table \ref{Algorithm} are used to cluster each dataset into a specified number of clusters. Notice that while the hierarchical algorithms produce a single partition for any given number of clusters (as they are deterministic), K-Means and Spectral Clustering (equipped with K-Means) can be repeated multiple times from randomised initialisations, so they may produce different partitions with the number of clusters fixed. For this reason, each of these algorithms will be run 10 times on each dataset with the aim to increase the diversity of the results. Once all the partitions are produced, a final check is performed to ensure they are unique, removing any identical partitions that could skew the results. It is worth noting that for practical data analysis, it is good practice to initialise K-Means several times for a set number of clusters, and select the best solution based on the objective function rather than considering solutions from various initialisations equally valid. For our study, however, this has been performed for the sake of increasing the number and diversity of solutions.

We specify the number of clusters in three different ways: (i) as the correct number of clusters in each dataset according to its ground truth; (ii) with fewer clusters than the ground truth; and (iii) with more clusters than the ground truth. This is to determine the performance of internal validity indexes both in the ideal case where the ground-truth number of clusters is known as well as in cases where it is under- or overestimated. The number of clusters in candidate partitions used in settings (ii) and (iii) will be $\pm30\%$ of the number of clusters in the ground truth. Similar to Evaluation Scenario 1, each index in Evaluation Scenario 2 will be evaluated both in terms of (i) the percentage of cases for which both the external and internal indexes select the same partition as optimal (the ``Top Pick'' partition) out of the candidate partitions; as well as in terms of (ii) the Spearman correlations between the rankings of such candidate partitions by the external and each internal index. These measures will be computed for each of the three cases corresponding to the fixed number of clusters, for each dataset.

The selection of $\pm30\%$ of the number of clusters was made with the aim of assessing the behaviour of the indexes when the number of clusters was both over- and under-estimated. If the number of clusters were too close to the ground truth ($k^*$), then there was found to be little diversity in comparison to the partitions produced for the ground-truth number of clusters and changes in the behaviour may be too small to determine. On the other hand, if the number deviates too far from the ground truth, then the results may have limited relevance in practice. This choice was made as a middle ground to balance these two factors. 

In addition to the 6 traditional external validity indexes outlined in Section \ref{ExtInd}, the external index CDistance will also be employed here in Scenario 2. This enables us to correctly identify candidate partitions that are better or worse due to geometric reasons, despite having the same fixed number of clusters and potentially similar counts of observations ``correctly'' clustered according to ground-truth labels. As only one external validity index is capable of determining the geometric difference between partitions, it will be given a higher weight in determining the rank of each partition, specifically, three times the weight of each of the other 6 individual indexes. 

Similar to Evaluation Scenario 1, the mean and median correlations are computed across the datasets within each of the data sub-collections sharing common properties outlined in Section \ref{subsec:Data}, in order to understand how the impact of each property affects the rankings and performance of the internal indexes. This is supplemented by statistical testing, as described in Section \ref{stat}, to determine which of these properties have a statistically significant effect.

\subsubsection{Evaluation Scenario 3: Algorithm and External Index Independent} \label{AlgInd}

The two previously mentioned scenarios rely on the use of an external validity index in order to establish a reference ranking of partitions produced by clustering algorithms with respect to a ground truth. This assumes that the external validity index of choice is an ultimate supervised measure of clustering quality, however, there is no single, absolute way of measuring the similarity between two clustering solutions and the different external validity indexes are well-known to exhibit differing behaviours when ranking partitions, in addition to potential biases, as discussed in Section~\ref{ExtInd}. Combining multiple indexes into an ensemble mitigates this issue to a certain extent, but the analysis may still be influenced by the particular type of partition produced by each clustering algorithm in hand.

\begin{figure*}[h!]
 \centering
 \includegraphics[width=0.95\linewidth]{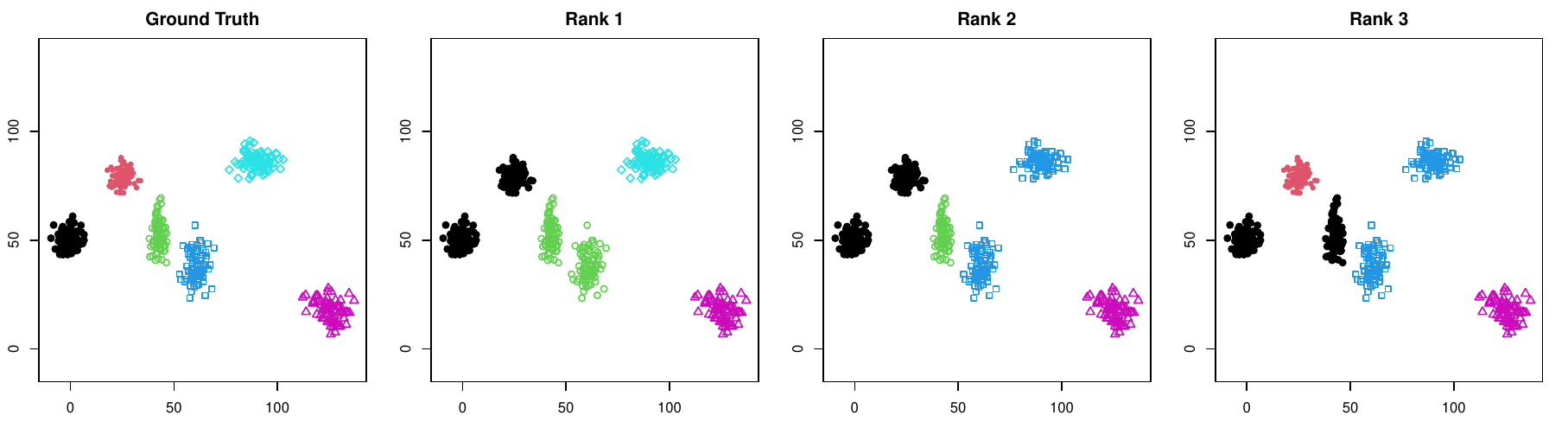}
 \caption{\label{fig:NoAlg} Example of multiple partitions with a known ranking produced with 4 clusters from a ground truth of 6 clusters using KL divergence. The value of the Jaccard index is 0.59 for all three of the produced partitions.}
\end{figure*}

In order to eliminate both the external indexes and clustering algorithms as potential confounding factors in our analysis, two methods (Procedure 1 and Procedure 2) for synthetically producing artificial clustering partitions from a ground-truth partition, in such a way that the resulting collection of partitions intrinsically possesses a statistically defined ranking of similarity with respect to the ground truth, have been devised. Both these (clustering algorithm free and external index free) methods rely on the assumption that the clusters within the data are produced by a multivariate Gaussian process, thus limiting this analysis to datasets satisfying this condition. For this reason, the datasets used in conjunction with this particular experimental methodology are a subset of the Type 2 datasets (see Section \ref{subsec:Data}) limited to Gaussian clusters with no background noise.

\begin{figure*}[h!]
 \centering
 \includegraphics[width=0.95\linewidth]{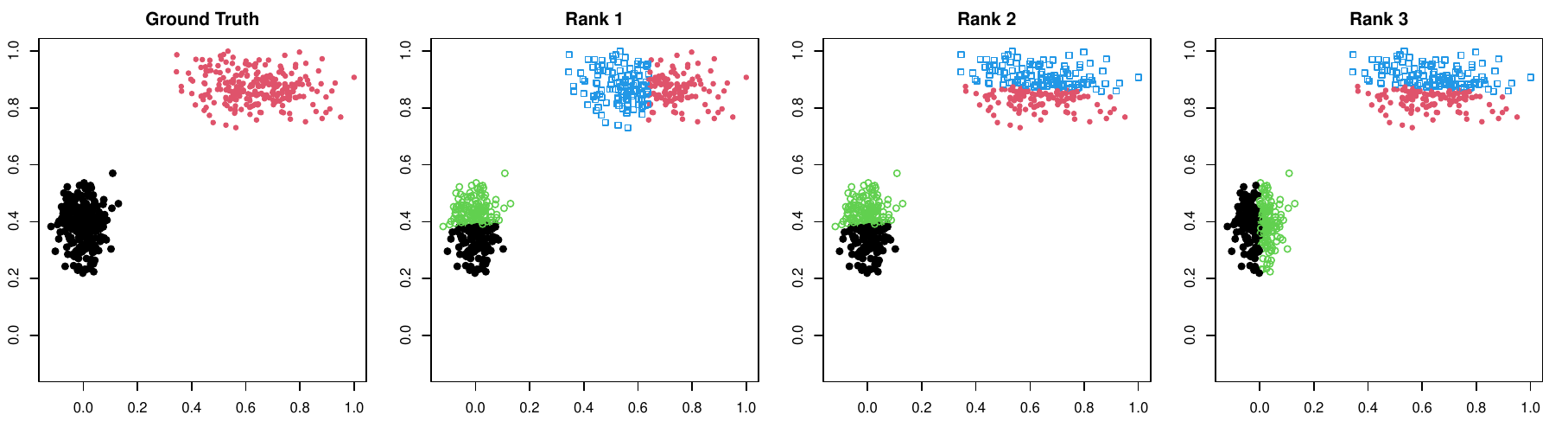}
 \caption{\label{fig:NoAlgSplit} Example of multiple partitions with a known ranking produced with 4 clusters from a ground truth of 2 clusters using variance. The value of the Jaccard index is 0.5 for all three of the produced partitions.}
\end{figure*}

One of the methods, Procedure 1, was designed to produce partitions with fewer clusters than the ground-truth partition ($k < k^*$), whereas the second method, Procedure 2, was designed to produce partitions with more clusters than the ground truth ($k > k^*$). These have been split into two different methods because rankings of partitions with underestimated and overestimated number of clusters are only comparable within (but not across) each of them.

Procedure 1, which produces partitions with fewer clusters than the ground truth ($k < k^*$), starts with the ground-truth partition and then iteratively joins at each step the most similar pair of clusters according to a \emph{supervised} measure of dissimilarity between the \emph{clusters' known distributions}. The well-known Kullback–Leibler (KL) divergence is adopted here as a statistically sound measure of dissimilarity. The KL divergence can be calculated between each pair of clusters in the ground truth to generate a dissimilarity matrix. In order to account for the KL divergence being asymmetric, we use the mean divergence between each pair of clusters calculated in both directions. Clusters are then merged iteratively, starting from the smallest to the largest divergence, to generate a partial referential hierarchical clustering based on the ground truth. Note that the KL dissimilarity matrix does not need to be updated after each merger because clusters resulting from a merger are not allowed to be further merged with other clusters.

This method can additionally be used to create rankings for a fixed number of clusters. This is achieved by following the previous steps to produce a hierarchy by merging clusters in a supervised way until the desired number of clusters is achieved. In order to produce different results with a desired number of clusters, this deterministic process is then repeated multiple times, each time excluding from being merged the most similar cluster pairs (according to the KL divergence) that had been allowed to be merged in the previous runs. This produces a set of partitions with a fixed number of clusters and known quality ordering as they involve mergers of progressively less similar clusters. One potential problem with this process may occur in cases where the best merger excludes several others that may be nearly as good, as in this case the exclusion of the best merger may allow for subsequent mergers to produce a net better solution.\footnote{E.g., consider four clusters A, B, C and D in a line. Let the distance from A to B be 2, B to C be 1, C to D be 2 and A to D be 5. By allowing the first merger of B and C, A and D is the only possible second merger, resulting in a solution with two mergers of total cost $1 + 5 = 6$. In the second run, with the merger of B and C excluded, the two mergers are A with B, and C with D, making for a better solution with total cost $2 + 2 = 4$.} In order to prevent these cases from happening, an additional check is put in place to ensure the dissimilarity between the merged clusters in the $n^{th}$ merger of each run is equal or greater than that between the merged clusters in the $n^{th}$ merger of the previous run. An example of the complete procedure can be seen in Figure \ref{fig:NoAlg}, where it is used to produce three partitions with the same number of clusters and a known ranking for one dataset.

Due to the nature of high dimensionality, the KL divergence between each cluster pair approaches equality as the number of dimensions increases, making any rankings by this procedure arbitrary. For this reason, Procedure 1 was limited to datasets with either 2, 4 or 6 dimensions in order to produce reliable results.

Procedure 2, which produces partitions with more clusters than the ground truth ($k > k^*$), starts by first calculating the covariance of each ground-truth cluster within the dataset. The clusters are then iteratively split in a decreasing order of cluster volume (under a multivariate Gaussian distribution) to produce a partial hierarchy. Each split is performed along the cluster's axis of largest variance (i.e., the eigenvector associated with the largest eigenvalue of a cluster's covariance matrix). Similar to Procedure 1, clusters are only split once, rather than being successively split multiple times. 

In order to produce multiple partitions for a fixed number of clusters greater than the ground truth, we follow the same basic strategy adopted for the case where the desired number of clusters is smaller than the ground truth (Figure \ref{fig:NoAlg}). The difference is that now we progressively exclude the axes (of largest variances) along which clusters had previously been allowed to split. An example of this procedure can be seen in Figure \ref{fig:NoAlgSplit}, where it is used to produce three partitions with the same number of clusters and a known ranking.

From the above two methods, designed to produce partitions with fewer or with more clusters than the ground-truth without relying on any particular clustering algorithm, hierarchies of clustering partitions are produced where the order of partitions with respect to their quality is known by construction, and it is independent of any particular external index. The order within each hierarchy is then taken as the rankings of partitions to be compared against the rankings produced by internal validity indexes. 
For the methods where the number of clusters is fixed, such a fixed number of clusters is chosen in the same way as in Scenario 2 (Section \ref{Method2}), namely,  $\pm30\%$ of the number of clusters in the ground truth. In order to ensure that there are sufficient points for correlation computations, only datasets where five or more partitions are produced by either supervised partitioning method are used.

Again, similar to Evaluation Scenario 1 and 2, each index in Evaluation Scenario 3 will be evaluated both in terms of (i) the percentage of cases for which the internal validity indexes selected the optimal candidate partition (the ``Top Pick'' partition) from the known ordering; as well as in terms of (ii) the Spearman correlations between the known rankings of the candidate partitions and each internal validity index. A key difference here is for each dataset, four sets of partitions are produced corresponding to the two aforementioned supervised partitioning methods for the two cases of varied and fixed number of clusters, for each of which the evaluation measures will be calculated using the resulting reference rankings instead of one produced by an external validity index.\footnote{In addition to assessing the performance of the internal validity indexes, the performance of the external validity indexes adopted in Evaluation Scenarios 1 and 2 (but not here, in Evaluation Scenario 3) will also be assessed by measuring the correlation between its ranking of partitions against the known reference rankings, in order to determine the impact from the use of such external indexes in Evaluation Scenarios 1 and 2.}

This allows the assessment of each index's ability to determine partition quality without the use of clustering algorithms or external indexes, e.g., in scenarios where external indexes are unable to determine partition quality accurately. It should be noted that these partitions are simpler than what may be produced by traditional clustering algorithms, and as such, the results in this Evaluation Scenario 3 alone cannot be relied upon to determine an internal index's performance, they are rather \emph{complementary} to Scenarios 1 and 2. 

\section{Results} \label{results}

\subsection{Evaluation Scenario 1} \label{sec:eval1}

\begin{table*}[h]
\centering
\resizebox{\textwidth}{!}{%
\begin{tabular}{rlllllll}
  \hline
  & Top Pick & Top Pick  & Mean Correlation  &  Mean Correlation  &  Median Correlation  &  Median Correlation \\ 
  Index & Type 1 & Type 2 & Type 1 & Type 2 & Type 1 & Type 2 \\
   \hline
  VRC & 83.2\% (1) & 57.1 \% ( 2 ) & 0.8 ( 1 ) & 0.66 ( 3 ) & 0.92 ( 1 ) & 0.86 ( 3 ) \\ 
  WB & 77.2\% (8) & 55.1 \% ( 4 ) & 0.75 ( 2 ) & 0.66 ( 2 ) & 0.91 ( 2 ) & 0.88 ( 1 ) \\ 
  Silhouette & 79.6\% (5) & 57.2 \% ( 1 ) & 0.69 ( 4 ) & 0.69 ( 1 ) & 0.74 ( 14 ) & 0.81 ( 5 ) \\ 
  Point-Biserial & 64\% (17) & 52.3 \% ( 6 ) & 0.66 ( 5 ) & 0.66 ( 4 ) & 0.88 ( 3 ) & 0.87 ( 2 ) \\ 
  Wemmert-Gancarski & 81.6\% (3) & 55.7 \% ( 3 ) & 0.65 ( 6 ) & 0.6 ( 7 ) & 0.76 ( 13 ) & 0.77 ( 7 ) \\ 
  DBCV & 79.6\% (4) & 54.5 \% ( 5 ) & 0.6 ( 10 ) & 0.62 ( 6 ) & 0.76 ( 12 ) & 0.78 ( 6 ) \\ 
  AUCC & 81.7\% (2) & 44.6 \% ( 11 ) & 0.6 ( 9 ) & 0.49 ( 9 ) & 0.85 ( 5 ) & 0.71 ( 9 ) \\ 
  PBM & 75.8\% (10) & 43.7 \% ( 12 ) & 0.71 ( 3 ) & 0.44 ( 10 ) & 0.88 ( 4 ) & 0.69 ( 10 ) \\ 
  Grex & 79.1\% (6) & 47.1 \% ( 8 ) & 0.62 ( 8 ) & 0.44 ( 11 ) & 0.8 ( 8 ) & 0.63 ( 11 ) \\ 
  C-index & 75.9\% (9) & 45.8 \% ( 10 ) & 0.58 ( 12 ) & 0.52 ( 8 ) & 0.81 ( 7 ) & 0.73 ( 8 ) \\ 
  G(+) & 78.3\% (7) & 41.9 \% ( 14 ) & 0.45 ( 16 ) & 0.36 ( 12 ) & 0.84 ( 6 ) & 0.61 ( 12 ) \\ 
  CDbw & 26.5\% (22) & 43 \% ( 13 ) & 0.26 ( 20 ) & 0.64 ( 5 ) & 0.65 ( 17 ) & 0.83 ( 4 ) \\ 
  Gstr & 74.6\% (11) & 36.6 \% ( 16 ) & 0.59 ( 11 ) & 0.3 ( 17 ) & 0.77 ( 11 ) & 0.45 ( 16 ) \\ 
  CS & 70.5\% (12) & 36.3 \% ( 17 ) & 0.51 ( 14 ) & 0.36 ( 13 ) & 0.73 ( 15 ) & 0.53 ( 14 ) \\ 
  CVDD & 69.4\% (13) & 22.5 \% ( 22 ) & 0.65 ( 7 ) & 0.11 ( 22 ) & 0.79 ( 9 ) & 0.14 ( 22 ) \\ 
  SV & 65.6\% (16) & 33.8 \% ( 19 ) & 0.48 ( 15 ) & 0.33 ( 15 ) & 0.68 ( 16 ) & 0.5 ( 15 ) \\ 
  DB & 54\% (19) & 47.6 \% ( 7 ) & 0.32 ( 18 ) & 0.32 ( 16 ) & 0.4 ( 20 ) & 0.43 ( 17 ) \\ 
  XieBeni & 67.4\% (15) & 46.4 \% ( 9 ) & 0.07 ( 24 ) & 0.27 ( 18 ) & 0.0035 ( 23 ) & 0.28 ( 20 ) \\ 
  Tau & 23.9\% (24) & 19 \% ( 24 ) & 0.31 ( 19 ) & 0.36 ( 14 ) & 0.53 ( 18 ) & 0.55 ( 13 ) \\ 
  CVNN & 69\% (14) & 20.8 \% ( 23 ) & 0.57 ( 13 ) & -0.09 ( 26 ) & 0.78 ( 10 ) & -0.15 ( 26 ) \\ 
  LCCV & 47.9\% (21) & 33.6 \% ( 20 ) & 0.35 ( 17 ) & 0.21 ( 21 ) & 0.42 ( 19 ) & 0.29 ( 19 ) \\ 
  SD & 52.5\% (20) & 35.8 \% ( 18 ) & 0.13 ( 22 ) & 0.22 ( 20 ) & -0.00087 ( 24 ) & 0.24 ( 21 ) \\ 
  S\_Dbw & 12.1\% (26) & 22.7 \% ( 21 ) & 0.082 ( 23 ) & 0.27 ( 19 ) & 0.36 ( 21 ) & 0.42 ( 18 ) \\ 
  Dunn & 63.2\% (18) & 39.7 \% ( 15 ) & -0.058 ( 25 ) & 0.099 ( 23 ) & -0.15 ( 25 ) & 0.068 ( 24 ) \\ 
  Ratkowsky-Lance & 21.7\% (25) & 16.3 \% ( 25 ) & 0.16 ( 21 ) & 0.094 ( 24 ) & 0.17 ( 22 ) & 0.11 ( 23 ) \\ 
  SSDD & 24.2\% (23) & 14.2 \% ( 26 ) & -0.19 ( 26 ) & 0.012 ( 25 ) & -0.27 ( 26 ) & -0.011 ( 25 ) \\ 
   \hline
\end{tabular}}
 \caption{\label{Table:Experiment1} The percentage of cases for which both the external and internal validity indexes selected the same candidate partition as optimal for each combination of clustering algorithm and dataset (``Top Pick''), in addition to the mean and median Spearman correlations between the rankings of such candidate partitions by the external index and each internal index, reported separately for both Type 1 and Type 2 datasets. In brackets is the relative rank of each index with respect to the corresponding evaluation criterion.}
\end{table*}

The results for the first scenario are summarised in Table \ref{Table:Experiment1} for both Type 1 and Type 2 data. The relative rank of each internal index in terms of their performance according to the evaluation criteria described in Section \ref{method1} is displayed within brackets (from best to worst performing scored from 1 to 26, respectively). Of the 129416 sets of partitions (16177 datasets $\times$ 8 clustering algorithms), 14927 were removed due to failing to produce at least one partition with an ARI greater than 0.6. This represents a removal of approximately 11.5\% of the partitions, with the majority of removals stemming from the Single Linkage and Complete Linkage clustering algorithms.

\begin{figure*}[h!]
 \centering
 \includegraphics[width=0.9\linewidth]{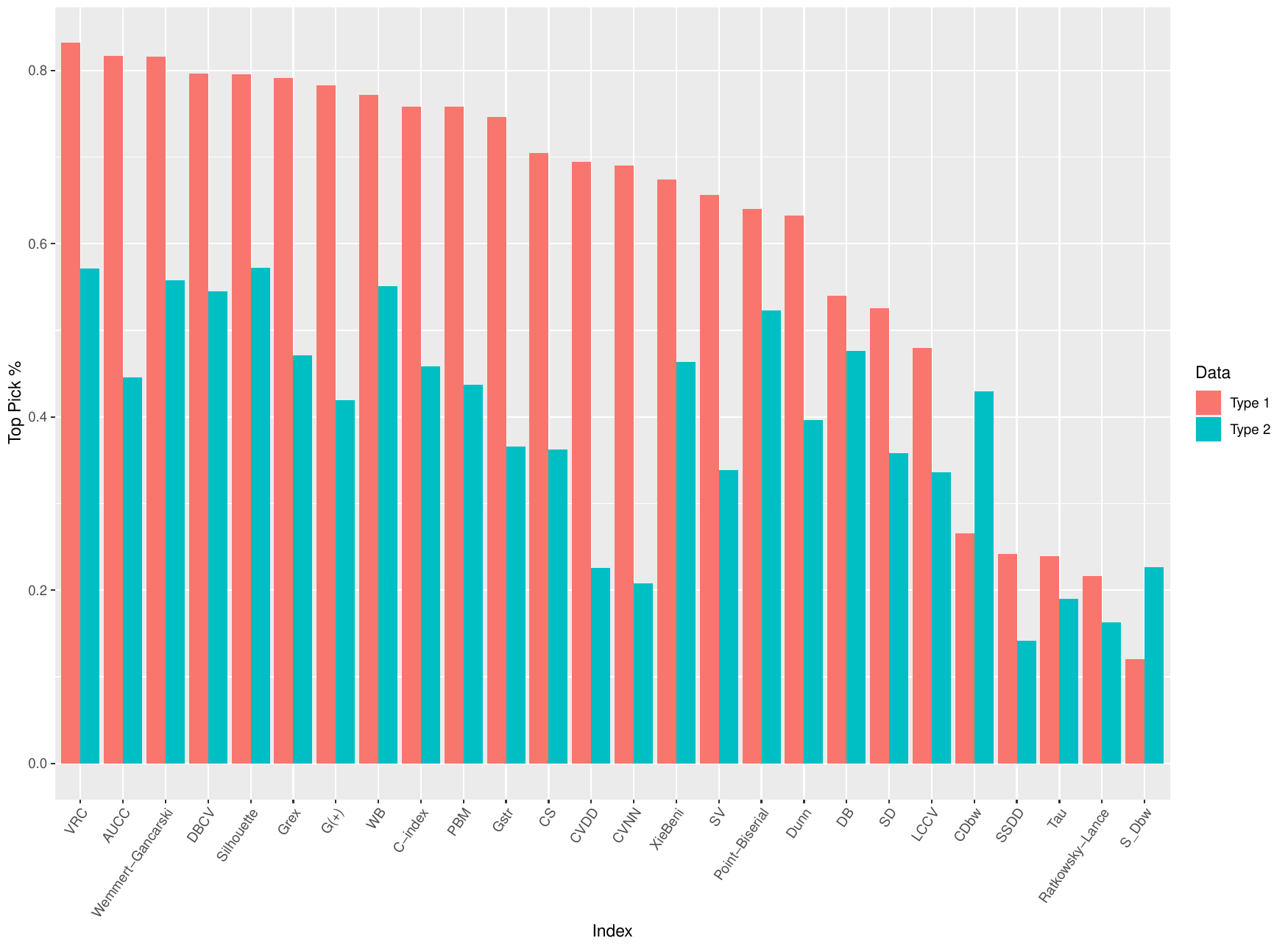}
 \caption{\label{fig:CorrectPercent} Bar chart of the percentage of cases (for each combination of clustering algorithm and dataset) where an internal validity index selected as best the correct candidate partition, i.e., the one also elected as best by the external criterion.}
\end{figure*}

From the percentage of cases where there is agreement in terms of the best partition, we can see in Table~\ref{Table:Experiment1} (\emph{Top Pick} columns) that the ability of many internal indexes to identify such a referential best partition differs largely between the Type 1 and Type 2 datasets. With the exception of CDbw and S\_Dbw, all the other indexes performed notably better overall for the simpler Type 1 datasets compared to the more complex Type 2 data, as also illustrated in Figure \ref{fig:CorrectPercent}. Despite most internal validity indexes performing worse for the Type 2 data, we observe significant disagreement in the ordering of indexes between the two data types, with a number of well performing indexes for the Type 1 data performing disproportionately worse than the other indexes for the Type 2 data. For instance, AUCC and G(+), which ranked well in the Type 1 data, ranked significantly worse for the Type 2 data. Inversely, indexes such as Point-Biserial and DB, which perform comparatively poorly for the Type 1 data, perform among the best for the Type 2 data. The internal validity indexes VRC, Silhouette, Wemmert-Gancarski, WB and DBCV all performed well across both sets of data. For the Type 2 data, it should be noted that despite no individual index consistently identifying the best partitions across most datasets, one or more indexes were able to identify the best partition produced by at least one of the clustering algorithms in all except 13 of the 16177 datasets. There appears to be a large discrepancy in the performance of Point-Biserial between the Type 1 and Type 2 data. The previous study in \cite{Vendramin2010} noticed Point-Biserial does not perform well when few dimensions are present, particularly when a large number of clusters are additionally present. This was also observed within this study for both Type 1 and Type 2 data particularly when the number of dimensions was fewer than six, however as the Type 1 data features a more limited range of dimensions this had a disproportionate affect on its performance for the Type 1 data.

In terms of correlation, it can be seen from Table~ \ref{Table:Experiment1} (last four columns) that the overall reduction in performance from Type 1 to Type 2 data is less prominent than it is for the ``Top Pick'' evaluation, but it is still noticeable, especially in terms of mean correlation. Since only a very few indexes defy this trend, performing notably better for Type 2 than for Type 1 data (e.g. CDbw), no major differences are observed in terms of the overall correlation-based rankings of the indexes across the two data types. These rankings indicate that VRC, WB and Point-Biserial stand out with respect to their overall performance in this aspect. The first two (VRC and WB) were also among the top-performers according to the ``Top Pick'' evaluation, which shows that these indexes performed consistently well across all aspects and types of data considered.

\begin{figure*}[h!]
 \centering
 \includegraphics[width=0.95\linewidth]{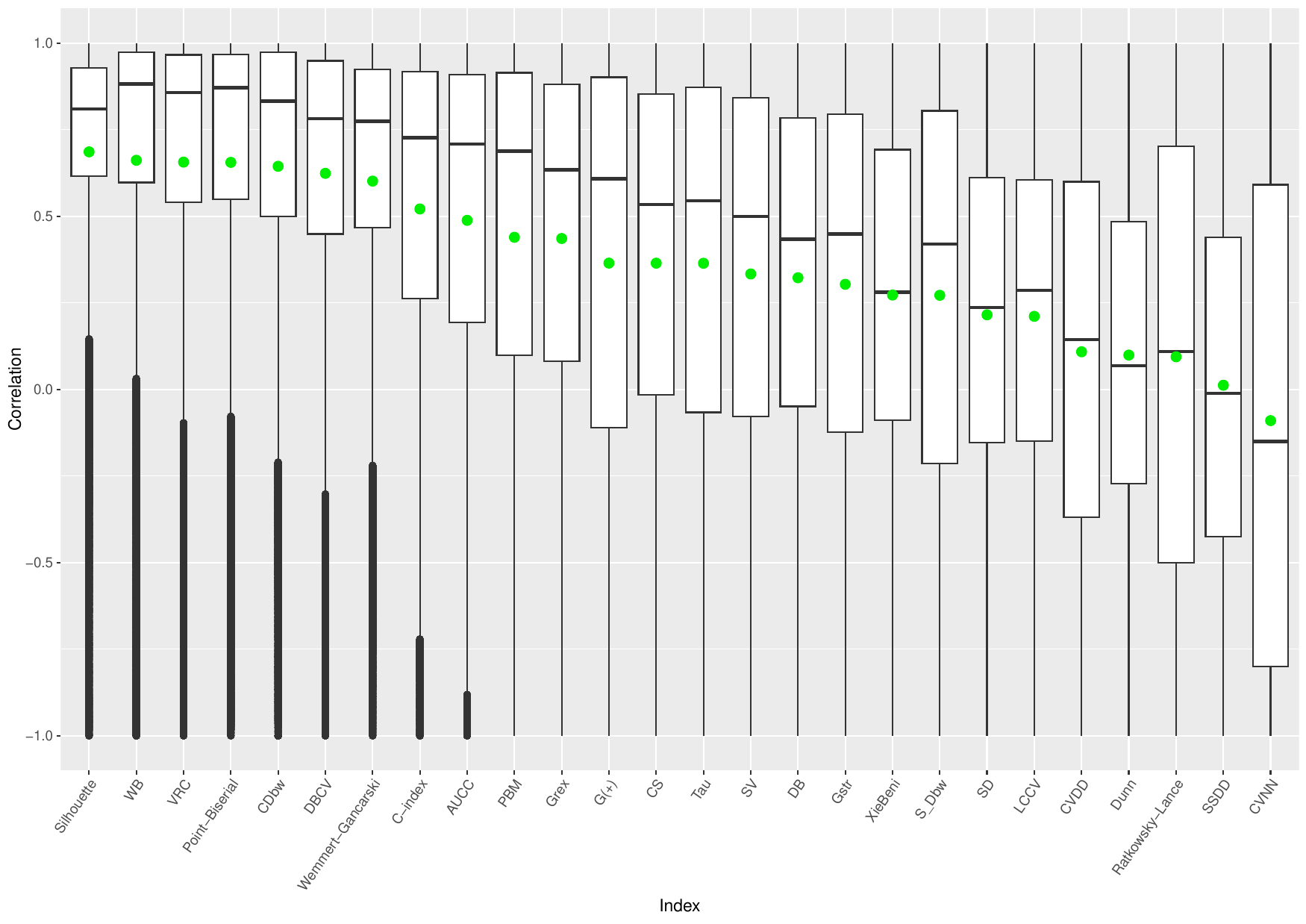}
 \caption{\label{fig:CorType2} Boxplots of the distributions of Spearman correlations between each internal validity index and the external ranking for the Type 2 data, ordered by mean correlation. Mean and median values are displayed with a green dot and line through each segment, respectively, and outlier points displayed as black points.}
\end{figure*}

Recalling that summary statistics such as mean and median correlation may not capture by themselves all relevant aspects behind an index's behaviour, possibly leading to deceptive conclusions, in the following we take a closer look into the correlation results. Due to the distribution of correlations being bounded between 1 and -1, the correlations are negatively skewed. Observing the median correlations in Table \ref{Table:Experiment1}, we can confirm this with generally higher median than mean correlations. For the Type 1 data we see a significant change in rankings for correlation when comparing the mean and median rankings, notably Silhouette and Wemmert-Gancarski drop from fourth and sixth, to fourteenth and thirteenth respectively, while G(+) and AUCC increase from sixteenth and ninth to  sixth and fifth place. However, in comparison the rankings of the indexes between mean and median correlation for the Type 2 datasets are generally unchanged, with the most significant difference being for the Silhouette index which has the highest mean correlation but only the fifth highest median correlation.  In Figure \ref{fig:CorType2}, for the Type 2 data we can see that despite the Silhouette index featuring a lower median correlation compared to the other top performing indexes, its interquartile range is smaller leading to a more compact distribution, leading to a more consistent index in performance compared to other indexes. The indexes WB, VRC, Point-Biserial, CDbw, DBCV and Wemmert-Gancarski can all be observed to also have small interquartile ranges relative to the remaining indexes, although a larger range than that of the Silhouette index which performs the best in this aspect. Despite this, all indexes produce poor performance under some circumstances, noted by outliers with highly negative correlations. This most frequently occurs when all partitions for a dataset are of similar quality to each other resulting in arbitrary rankings, and thus no or occasionally negative correlation with the external index. Although there appears to be many outliers, we should bear in mind that these distributions each contain roughly $\sim{128000}$ points, so infrequent behaviours may seem exaggerated by the boxplots. 

The remainder of the results will focus on the Type 2 data, primarily due to it containing a larger sample size with a greater diversity of clustering problems, in contrast to the limitations of the Type 1 data, such as datasets and ground-truth partitions with limited numbers of clusters and dimensions. As the Type 2 was designed to evenly distribute datasets across wider ranges of multiple properties of interest, the inclusion of Type 1 data along with Type 2 data would have the potential of skewing results, particularly when addressing properties not present in the former.

A pairwise Willcoxon test with Bonferroni correction was performed to determine between which internal validity indexes there is a difference in performance measured by correlation for the Type 2 datasets, as described in Section \ref{stat}. At a 5\% significance level, all internal validity indexes were found to have a statistically significant difference in performance measured by correlation, with the exception of the following four pairings of indexes: (i) CS and G(+); (ii) CS and Tau; (iii) G(+) and Tau: (iv) Grex and PBM.

\begin{figure*}[h!]
 \centering
 \includegraphics[width=0.96\linewidth]{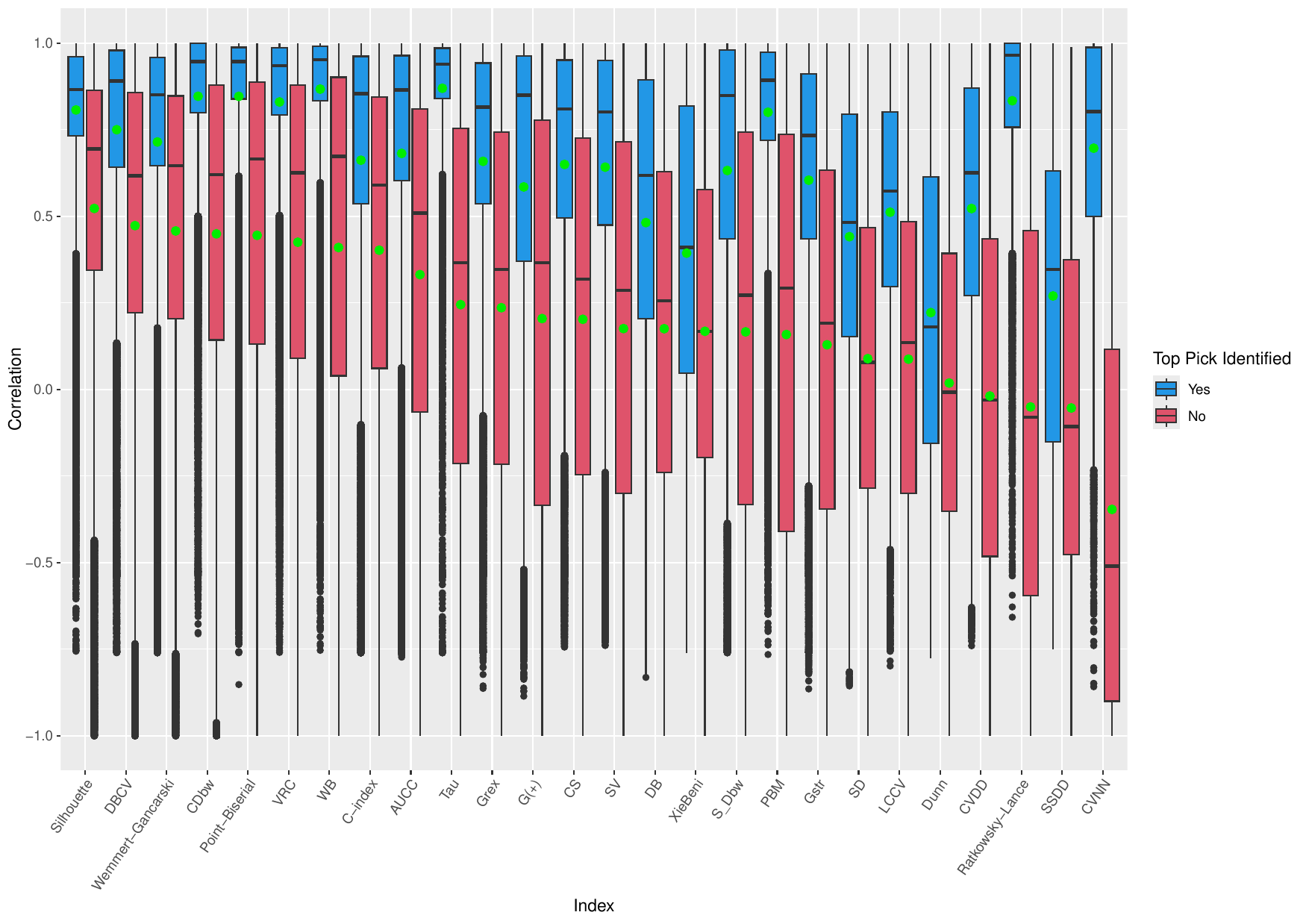}
 \caption{\label{fig:CorType2O} Boxplots of the distributions of Spearman correlations for each index for the Type 2 data, grouped by whether the ``Top Pick'' partition was or was not found by the index (the proportion of datasets within each category for each index is shown in the ``Top Pick, Type 2'' column of Table~\ref{Table:Experiment1}). Mean, median, and outlier values are displayed with a green dot, a line through each segment, and as black dots, respectively.}
\end{figure*}

In Figure \ref{fig:CorType2O}, we present the distribution of correlations using the Type 2 datasets, for each index separately in the two following scenarios: correctly determining the ``Top Pick'' partition, or failing to determine the ``Top Pick'' partition. Notice that the correlation is significantly higher for each index in datasets where the ``Top Pick'' partition is identified by the index, showing some level of agreement or consistency between these two evaluation strategies. However, we also observe many outlying instances where either the ``Top Pick'' partition is correctly identified with a low correlation, or not identified despite a high correlation.
 
For the former case, the primary cause appears to occur when the majority of the candidate partitions are of similar, usually poor quality, where many indexes fail to determine an adequate ordering as there is little distinction between the partitions, with the exception of the ``Top Pick'' one. This most commonly occurs when most of the partitions produced are of low quality containing little practical structure, outside the best one. In this scenario, correlation may unfairly penalise indexes, as it is not reasonable to expect indexes to accurately rank similar partitions or partitions that contain little to no structure. Here one fault of the use of Spearman correlation is unveiled, as when many partitions of similar quality are present, disagreements in rankings may exaggerate the differences compared to other measures of correlation, such as Pearson. Another cause is the presence of non-linear relationships between the internal and external validity indexes, which may reduce the Spearman correlation significantly even when both internal and external indexes identify the same partition as best. These non-linear trends, which will be discussed in more detail below, are generally a sign of poor performance of internal validity indexes, a case in which their performance is accurately reflected by such lower correlation values.

As for the latter case, where the ``Top Pick'' partition is not identified despite a high correlation, it is possible that such a partition violates the assumptions of an internal index in a way that partitions with worse external evaluations may not. An example is if overlapping clusters are present in the ground-truth partition; certain internal indexes may favour partitions where the overlapping clusters are considered a single cluster, and these may not be as close to the ground-truth partition as the candidate indicated as the ``Top Pick'' solution according to the external index. This shows that both assessment measures (``Top Pick'' and correlation) are valuable as both represent different aspects of an internal index's performance, in either their ability to identify the best externally assessed partition among a collection of candidates, or their ability to identify reasonable partitions despite being unable to detect the best one. These cases emphasise the importance of both measures of performance, as each may capture different aspects and likelihood to produce reasonable solutions in real-world clustering problems, in addition to recognising each measure as having drawbacks where performance may be misleading.

In Figure \ref{fig:CorType2O}, some of the top-performing indexes are shown to maintain a higher correlation even when the ``Top Pick'' partition is not correctly identified. The Silhouette index outperforms all indexes in this aspect, with both the highest mean and median correlation depicted for cases where the optimal partition is not determined, on top of identifying the optimal partition more frequently in Table \ref{Table:Experiment1}. DBCV and Wemmert-Gancarski also perform well in this aspect. 

\begin{table}[h]
\centering
\begin{tabular}{lcccc}
  \hline
 Index & All $k$ & $k < k_{O}$ & $k > k_{O}$ & Mean Range \\ 
  \hline
  Silhouette & 0.69 & 0.75 & 0.74 & 0.38 \\ 
  DBCV & 0.62 & 0.78 & 0.53 & 0.39 \\ 
  Wemmert- & & & & \\
  Gancarski & 0.60 & 0.79 & 0.50 & 0.42 \\ 
  Point- & & & & \\
  Biserial & 0.66 & 0.81 & 0.53 & 0.44 \\ 
  VRC & 0.66 & 0.71 & 0.66 & 0.45 \\ 
  CDbw & 0.64 & 0.80 & 0.47 & 0.45 \\ 
  WB & 0.66 & 0.85 & 0.53 & 0.46 \\ 
  Grex & 0.44 & 0.54 & 0.30 & 0.48 \\ 
  LCCV & 0.21 & 0.28 & 0.28 & 0.55 \\ 
  C-index & 0.52 & 0.78 & 0.21 & 0.61 \\ 
  DB & 0.32 & 0.59 & 0.17 & 0.62 \\ 
  PBM & 0.44 & 0.44 & 0.49 & 0.62 \\ 
  Gstr & 0.30 & 0.32 & 0.21 & 0.64 \\ 
  AUCC & 0.49 & 0.75 & 0.16 & 0.66 \\ 
  XieBeni & 0.27 & 0.50 & 0.32 & 0.71 \\ 
  SD & 0.22 & 0.42 & 0.54 & 0.74 \\ 
  CS & 0.36 & 0.72 & 0.04 & 0.74 \\ 
  SSDD & 0.01 & 0.19 & 0.29 & 0.80 \\ 
  CVDD & 0.11 & 0.06 & 0.21 & 0.81 \\ 
  SV & 0.33 & 0.72 & -0.04 & 0.82 \\ 
  Dunn & 0.10 & 0.23 & 0.21 & 0.82 \\ 
  Tau & 0.36 & 0.22 & 0.67 & 0.84 \\ 
  G(+) & 0.36 & 0.75 & -0.07 & 0.85 \\ 
  S\_Dbw & 0.27 & 0.77 & -0.32 & 1.09 \\ 
  Ratkowsky- & & & & \\
  Lance & 0.09 & -0.26 & 0.89 & 1.28 \\ 
  CVNN & -0.09 & -0.42 & 0.87 & 1.37 \\ 
   \hline
   \end{tabular}
\caption{\label{Table:Exp1CorType2Var} Spearman correlations across all partitions (all $k$), across partitions with more clusters than the ``Top Pick'' partition ($k > k_{O}$), across partitions with fewer clusters than the ``Top Pick'' partition ($k < k_{O}$), and the range between these three measures averaged across all datasets and clustering algorithms. All entries are \emph{mean values} computed over all combinations of (Type 2) datasets and clustering algorithms.}
\end{table}

As discussed in Section \ref{subsec:correlation}, the correlation was additionally calculated separately across the subsets of partitions with either fewer ($k < k_{O}$) or more ($k > k_{O}$) clusters than the best externally assessed (``Top Pick'') partition for each clustering algorithm and dataset, and the range across these three correlation values was computed. The mean values of these three correlation measures and their mean range across all combinations of algorithms and datasets are reported in Table \ref{Table:Exp1CorType2Var}. Noticeably, Silhouette, VRC, DBCV, Wemmert-Gancarski, Point-Biserial and CDbw stand out for performing well in terms of the mean overall correlation while also having limited variation across the different correlation measures, indicating more consistent rankings across distinct ranges of clusters comprising under- or over-clustering scenarios. 

Several indexes present correlations close to zero for one region, while high or reasonable correlations for the other region. Most notably, G(+), SV, CS, AUCC and DB perform poorly when $k > k_{O}$, whereas CVDD performs poorly when $k < k_{O}$. This indicates these indexes are unable to distinguish between partitions in these respective regions, significantly reducing their performance. This behaviour is undesirable, as these indexes will likely fail to identify any good partitions produced in regions they perform poorly in. Other indexes such as CVNN, S\_Dbw and Ratkowsky-Lance are positively correlated on one region, while being significantly negatively correlated in the other. This behaviour is consistent with such indexes often acting as monotonic increasing, or decreasing for S\_Dbw, and showing significant bias by favouring partitions with a greater number of clusters regardless of partition quality, which is highly undesirable.

\begin{figure}
 \centering
 \includegraphics[width=\linewidth]{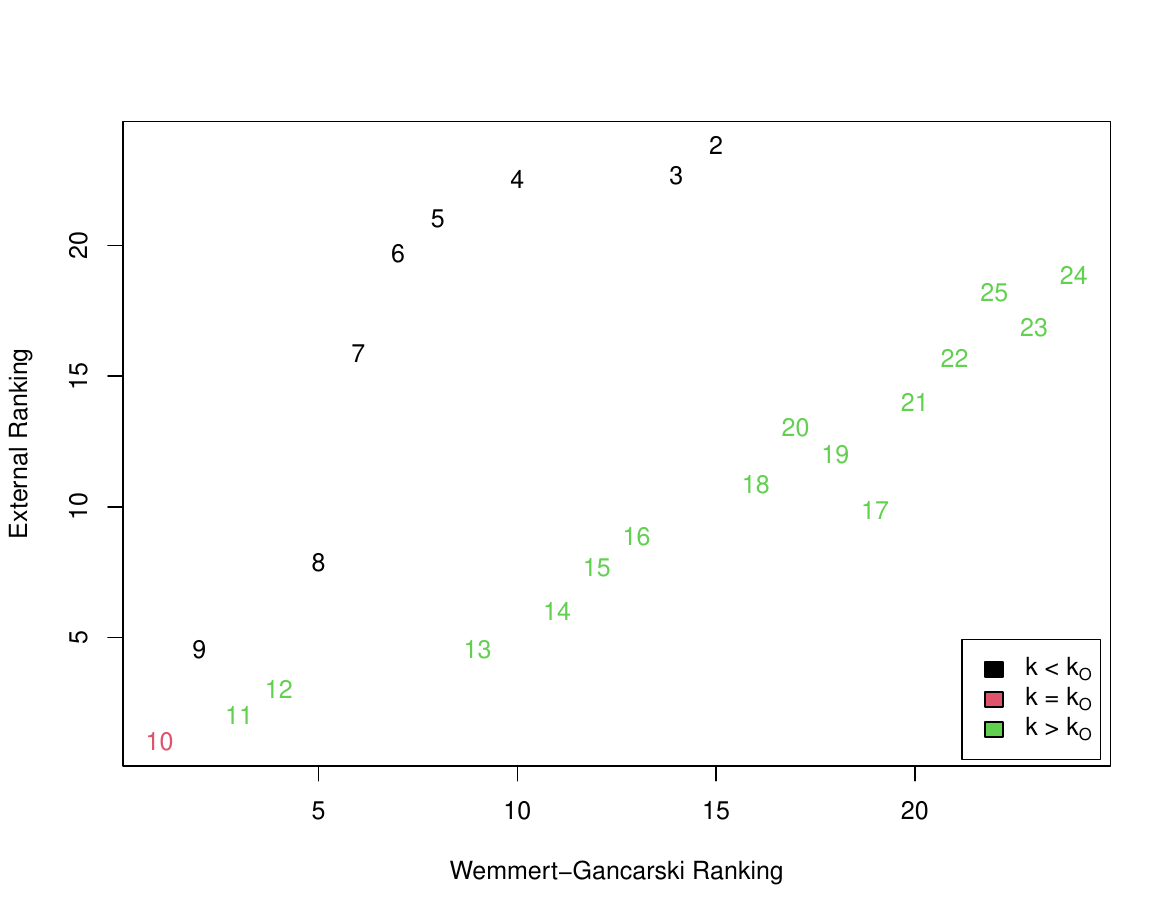} 
 \caption{\label{fig:WGnonL} Scatter plot of the partition \emph{ranking} (the smaller the better) according to the Wemmert-Gancarski index and the aggregated rankings of the six external indexes, showing two separate trends based on the number of clusters. Each point represents a partition produced by Spectral Clustering, its label indicates the number of clusters within that partition, and the color indicates if the number of clusters is less than (black), equal to (red), or greater than (green) the number of clusters in the optimal (best externally assessed) candidate partition. The dataset used contained 10 ground-truth clusters and 120 dimensions.}
\end{figure}

A large discrepancy between such correlations tends to indicate poor performance of an internal index as reflected by an undesirable non-linear relationship between its rankings and those of an external validity index. Three main types of non-linear relationships were seen to occur between the rankings produced by external and internal indexes. The first type of non-linear relationship is when each region is internally well correlated, but the overall relationship has a lower correlation due to disagreement between the rankings of each region relative to each other, as seen in Figure \ref{fig:WGnonL}. This can primarily be seen in indexes with high correlations in both regions, and was observed for Silhouette, VRC, WB, SD, Grex, Wemmert-Gancarski, DB and CS.\footnote{Note that this behaviour can be observed at an individual dataset level, despite the fact that, on average across datasets, some of these aforementioned indexes have exhibited limited correlation discrepancies between regions, as reported in Table \ref{Table:Exp1CorType2Var}.} The primary cause of this is significant differences between internal and external indexes in how over- and underestimation of the number of clusters is handled. This may affect all indexes to some degree, however, it can be less obvious to indexes with poor performance as the impact may be hidden by other variations in the scatter plot. The impact on performance can vary significantly among the indexes where this effect is apparent, and it depends on other factors such as the dataset and clustering algorithm used, resulting in some indexes performing well overall despite this non-linear trend.

\begin{figure}
 \centering
 \includegraphics[width=\linewidth]{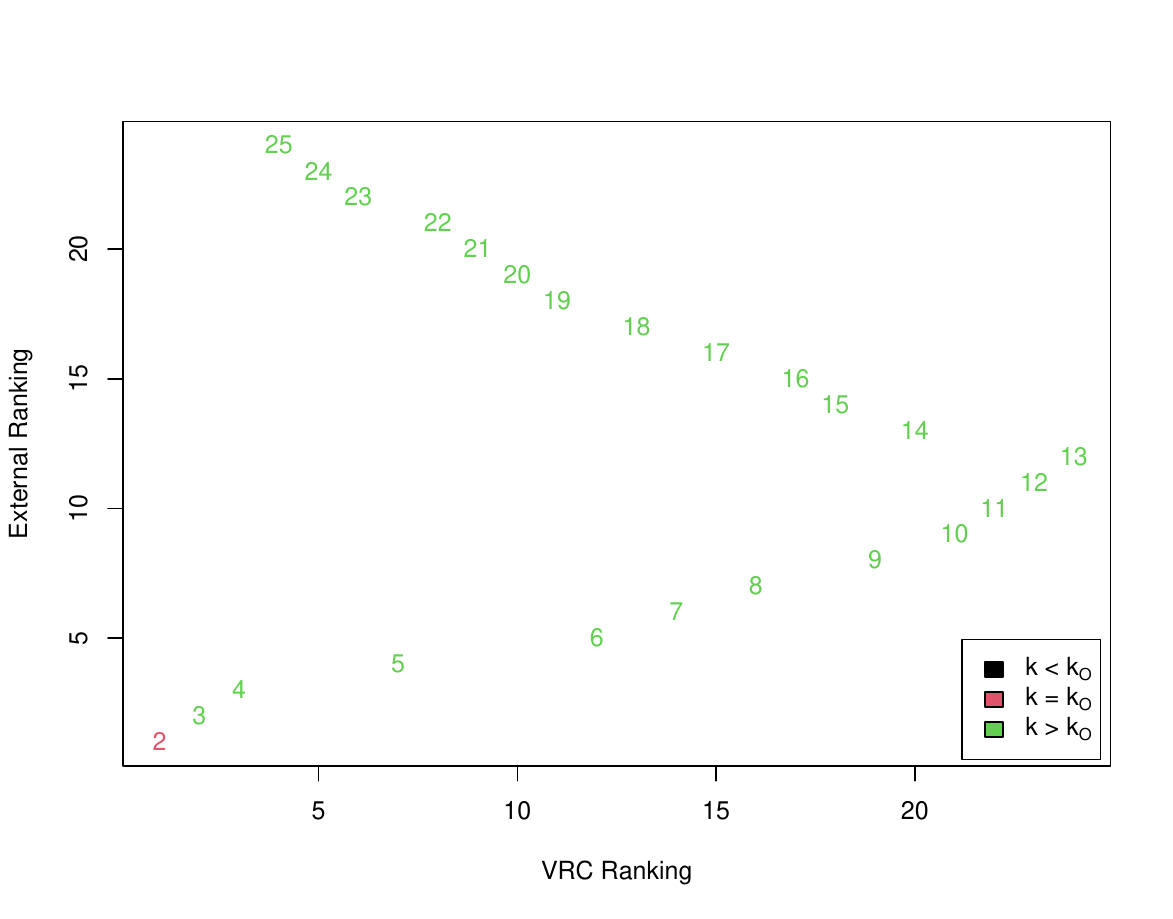} 
 \caption{\label{fig:NonLin2} Scatter plot of the partition \emph{ranking} (the smaller the better) according to the VRC index and the aggregated rankings of the six external indexes, showing a sharp inversion of the monotonic behaviour as the number of clusters in the partition exceeds 13. Each point represents a partition produced by Complete-Linkage, where the numeric label indicates the number of clusters within that partition, and the color indicates if the number of clusters is less than (black --- non-occurring), equal to (red) or greater than (green) the number of clusters in the optimal (best externally assessed) candidate partition. The dataset used contained 2 ground-truth clusters and 200 dimensions.}
\end{figure}

The second type of non-linear relationship occurs when an internal validity index becomes monotonically increasing (or decreasing for minimisation indexes) as the number of clusters in the partition under assessment increases significantly beyond the number of clusters in the optimal partition ($k >> k_{O}$), as seen in Figure \ref{fig:NonLin2} ($k_{O} = 2$), where the internal index can be seen to perform correctly in agreement with the ranks from an external index for partitions with up to $k=13$ clusters, beyond which all successive partitions are ranked better and better by the internal index (but not by the external one) as the index's value itself (omitted in the figure) starts acting monotonically increasing with the number of clusters despite the worse quality of the corresponding partitions (still correctly captured by the external ranking). This has been observed for VRC, Wemmert-Gancarski, Grex, AUCC, DB, PBM, XieBeni, WB, CVDD, SV and CS. 

It is worth noticing that, despite featuring non-linear relationships, some indexes, such as VRC and Wemmert-Gancarski, still exhibit relatively low mean difference between the three correlation measures in Table \ref{Table:Exp1CorType2Var}. This is due to two factors, the first being that these non-linear relationships may be less frequent for these indexes despite being clearly present in some individual datasets. Secondly, the non-linear relationships may be less pronounced for these indexes resulting in a smaller impact on overall average performance. 

The final, potentially critical type of non-linear relationship relates to indexes that only successfully rank partitions in one of the two regions. As it can be seen in Table \ref{Table:Exp1CorType2Var}, these include CVNN, Ratkowsky-Lance, S\_Dbw, G(+), SV and CS. There can be two cases where this behaviour appears. The first case occurs when an index is either monotonic increasing or decreasing across the entire range of partitions, as it has been observed for Ratkowsky-Lance, S\_Dbw, and CVNN, while the second case occurs when an index performs correctly in one region but is uncorrelated or performs poorly in the second region, such as for the indexes G(+), SV and CS.

\begin{figure}[h!]
 \centering
 \includegraphics[width=\linewidth]{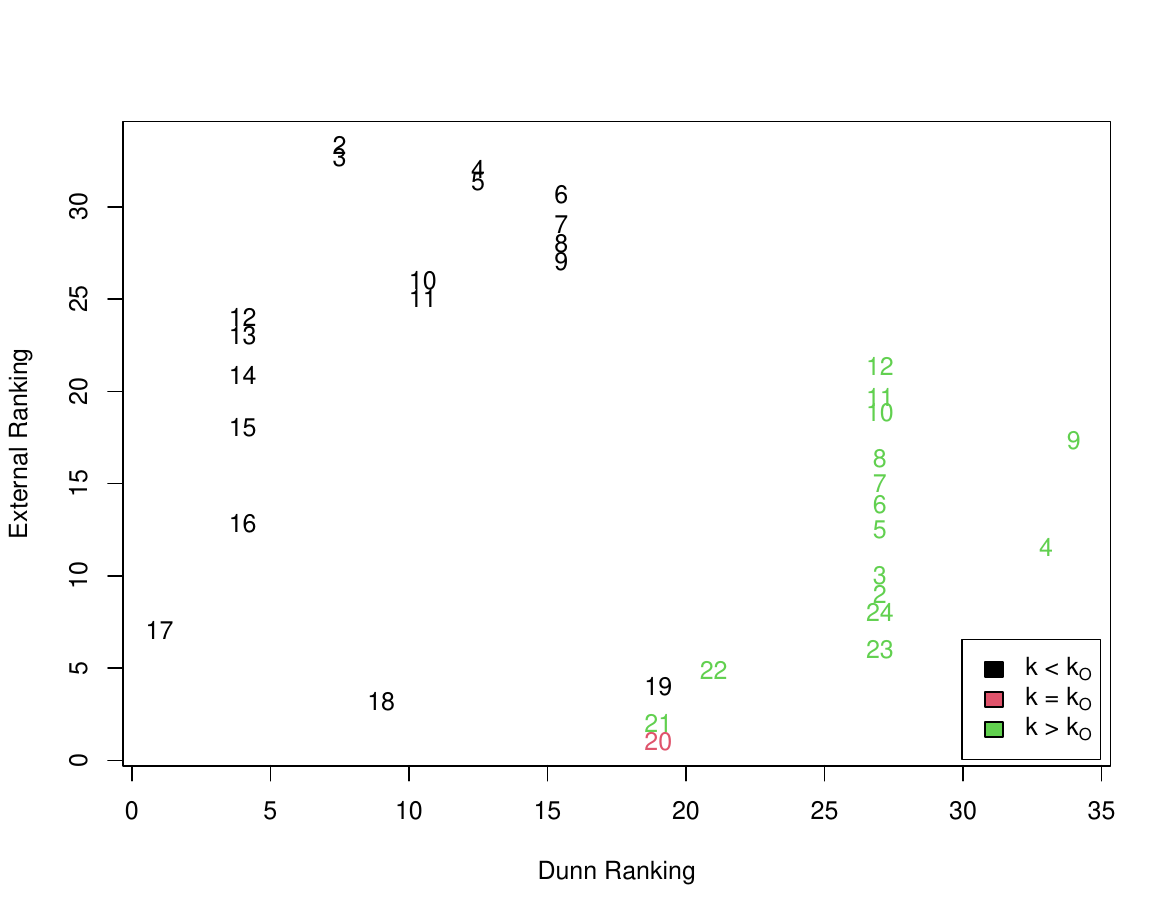} 
 \caption{\label{fig:NonLin3} Scatter plot of the partition \emph{ranking} (the smaller the better) according to the Dunn index and the aggregated rankings of the six external indexes, showing a general lack of relationship between the two indexes. Each point represents a partition produced by Spectral Clustering, where the numeric label indicates the number of clusters within that partition, and the color indicates if the number of clusters is less than (black), equal to (red) or greater than (green) the number of clusters in the optimal (best externally assessed) candidate partition. The dataset used contained 20 ground-truth clusters and 200 dimensions.}
\end{figure}

 As compared to other indexes, Dunn exhibited a unique behaviour in that it was often prone to assessing partitions of varying quality with similar values, thus mostly failing to properly rank partitions. An example of this can be seen in Figure \ref{fig:NonLin3}, where there is no apparent relationship between the external rankings and the Dunn index's rankings for partitions. Despite the lack of correlation, the Dunn index still outperformed certain indexes in selecting the ``Top Pick'' partition, however, it was mostly unable to properly rank other partitions, as it can be seen in Table \ref{Table:Experiment1}.

It is important to note this list of non-linear relationships is not exhaustive, and the frequency and degree to which they occur when assessing partitions across different combinations of datasets and clustering algorithms varies significantly from index to index. The average amount of correlation variation between under- and over-clustered regions, as observed in Table \ref{Table:Exp1CorType2Var}, provides an indication of the overall effect of these relationships on each index, where a higher difference between regions or poor performance in a single region indicates the presence of undesirable non-linear behaviour.

\begin{figure*}[h]
 \centering
 \includegraphics[width=0.95\linewidth]{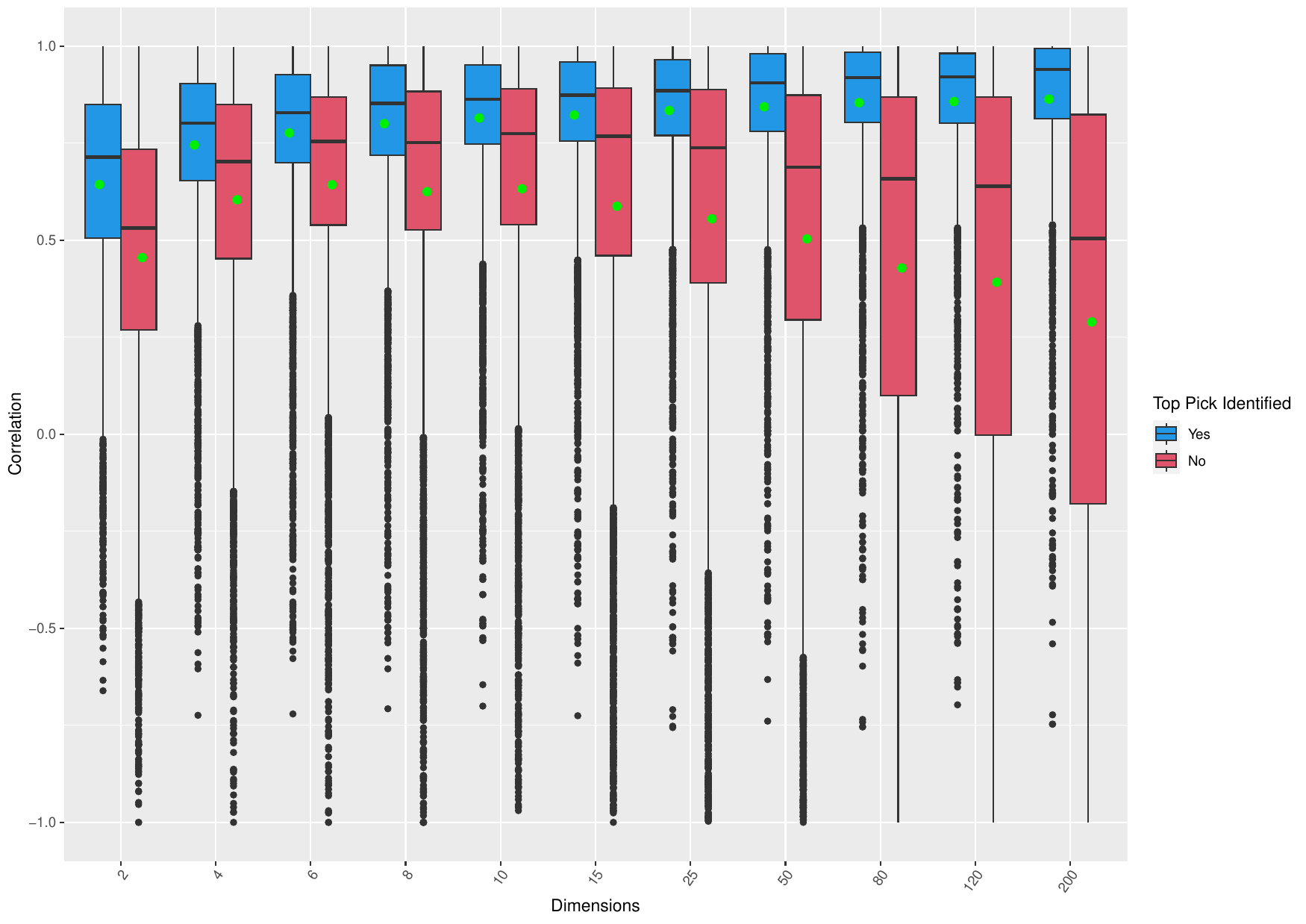}
 \caption{\label{fig:CorDims} Boxplots of the correlations for the Silhouette index across datasets with increasing dimensionality in Evaluation Scenario 1, grouped by whether or not the best externally assessed partition was correctly identified. Mean and median values are displayed with a green dot and a line through each segment, respectively.}
\end{figure*}

The choice of clustering algorithm was seen to significantly affect the performance of most internal validity indexes both in absolute as well as in relative terms, as illustrated in Table \ref{Table:Exp1Alg}, which shows the ranking of each index based on their mean correlations. Silhouette and CDbw stand out for each appearing as the top-ranked index in three out of eight clustering algorithms, with Silhouette being the top-performer for K-Means, Ward and HDBSCAN*, while CDbw performed the best for Single, Average and Complete Linkage. However, their ranks dropped significantly for specific algorithms, namely: for Single and Average Linkage, the Silhouette was outperformed by a number of indexes, despite a reasonably stable performance in absolute terms, while for EM-GMM and Spectral Clustering, CDbw's performance noticeably worsened both in relative as well as in absolute terms. Similar trends were observed for the majority of indexes, with their performance being dependent on the clustering algorithm used to a greater or lesser extent. This suggests that the choice of index should be determined based on the selected clustering algorithm to ensure proper performance. For instance, Tau performed among the best indexes for Spectral Clustering, and reasonably well for EM-GMM, despite its overall poor performance. Both CDbw and DBCV performed significantly better when paired with Single and Average Linkage clustering algorithms, suggesting that the partitions produced by these algorithms may be better suited to density-based validity indexes. In comparison, both of these indexes under-performed for Spectral Clustering, where traditional validity indexes performed better in comparison, indicating the partitions produced by Spectral Clustering may be less compatible with density-based indexes. K-Means, Ward and EM-GMM all produced significantly lower correlations across most validity indexes. Despite the significant impact of the clustering algorithm, several indexes, e.g., Ratkowsky-Lance, CVDD, CVNN, Dunn, and SSDD, consistently performed poorly regardless of the algorithm used.

The {\bf distribution} used to produce clusters had a moderate impact on the rankings of each index. As it can be seen from Table \ref{Table:CorProp} (three rightmost columns), although changes were relatively minor for most indexes, there are notable changes in the rankings of the best performing indexes, such as VRC, which performed the best for uniformly distributed clusters however fifth and seventh best for the Gaussian and logistic-based clusters respectively. For most indexes, the correlations were highest for clusters produced from a uniform distribution, followed by Gaussian clusters, with logistic-based clusters presenting the lowest correlations.

In order to better understand the impact of various data properties of interest on each internal validity index, namely, noise, dimensionality, number of clusters, cluster overlap, (im)balance, and compactness, each of these properties was tested at a  5\% significance level as described in Section \ref{stat}. All statistical tests were performed on the Type 2 datasets due to the larger quantity of data these entailed as well as greater variation in properties including properties not present in the Type 1 data. The performance (mean correlation) of each index is displayed in Table \ref{Table:CorProp} separately for datasets categorized within each property, in order to observe how these affect ranking of the indexes. To facilitate visual assessment of results, in Table \ref{Table:CorProp} the datasets were grouped into two groups for the various properties of interest. Specifically, $k_{high}$ stands for datasets with more than 10 ground-truth clusters, whereas $k_{low}$ comprises data with 10 or fewer ground-truth clusters. Similarly, $D_{low}$ refers to datasets with 25 or fewer dimensions, whereas $D_{high}$ incorporates datasets with more than 25 dimensions. Datasets with any degree of overlap using Equation \ref{eq:overlap} were considered datasets with overlap, while for the property of imbalance, datasets with an imbalance of 0.5 or greater using Equation \ref{eq:imbalance} were considered imbalanced. All datasets containing noise in the ground-truth partition are considered as noise datasets. Finally, datasets produced with a compactness value of 0.1 in MDCGen were considered Compact, while datasets produced with a compactness value of 0.8 were considered Sparse. These groupings are for the purpose of displaying results in Table \ref{Table:CorProp}, and not for the purpose of statistical testing. The results of the statistical tests detailed in Section \ref{stat}, alongside the assessment of the impact of performance seen in Table \ref{Table:CorProp}, are discussed below.

The quantity of {\bf overlap}, as measured by Equation \ref{eq:overlap}, was found to have a statistically significant negative correlation on performance of all internal validity indexes. Additionally, observing the results in Table \ref{Table:CorProp}, the presence of overlapping clusters caused the largest negative impact on the performance of internal validity indexes, with only a few indexes more reasonably sustaining performance to a certain extent. This resulted in a moderate change to the ranking of indexes, despite all indexes being negatively affected, as some indexes were less affected compared to others. Noticeably, WB performed the best for datasets containing overlapping clusters. The Wemmert-Gancarski index saw the largest change in its ranking under the presence of overlapping clusters, performing second best, in contrast to being seventh best overall. 

Increasing the {\bf imbalance} of clusters in the data, as measured by Equation \ref{eq:imbalance}, was found to have a statistically significant positive correlation with 14 of the studied indexes, namely G(+), Wemmert-Gancarski, Gstr, Grex, AUCC, C-index, DB, Dunn, XieBeni, CVDD, DBCV, SV, CS and S\_Dbw. In contrast, a significant negative correlation was found with the others, namely, Point-Biserial, Ratkowsky-Lance, SD, Tau, Silhouette, VRC, PBM, WB, CDbw\footnote{CDbw appears in Table \ref{Table:CorProp} to perform better for Imbalanced datasets due to the table using a binary measure of imbalance, and the correlation for CDbw being close to zero (-0.016).}, CVNN, SSDD and LCCV. Despite this, the impact on performance reported in Table \ref{Table:CorProp} was minimal, resulting in very minor changes in the relative ranks of each index under imbalanced datasets, as contrasted to the ranks across all datasets. 

For most internal validity indexes, performance on datasets with {\bf noise} was found to be worse at a 5\% significance level according to the Kruskal-Wallis test, with the exceptions of LCCV and C-index, which exhibited no significant effect, in addition to Dunn, XieBeni and S\_Dbw, which showed a statistically significant positive effect. Noise had the second largest impact on performance reported in Table \ref{Table:CorProp}. Indexes such as DBCV and Wemmert-Gancarski, despite still being negatively impacted, saw notable improvement in their relative ranks, while other indexes, such as VRC, dropped significantly down the ranking due to their performance being more heavily impacted than other indexes by the presence of noise.

{\bf Dimensionality} was split evenly with the performance of 13 indexes showing a statistically significant positive correlation (Ratkowsky-Lance, SD, Tau, Wemmert-Gancarski, Silhouette, C-Index, VRC, Dunn, XieBeni, WB, DBCV, SV and S\_Dbw) with the number of dimensions, while the performance of the remaining 13 indexes presented a significant negative correlation. The impact of dimensionality resulted in moderate changes to the rankings in Table \ref{Table:CorProp}.

A pattern was observed across several internal indexes, namely, an initial consistent increase in performance (as measured by mean correlation with the external index) with increasing dimensionality, up to a certain number of dimensions beyond which correlation becomes significantly more variable (across the available collections of partitions corresponding to different combinations of dataset and clustering algorithm) in a way that the aggregated result may still keep a positive net trend for some indexes, while reversing and becoming negative for others. This behaviour was more noticeable (to varying degrees) for G(+), Point-Biserial, Tau, Wemmert-Gancarski, Gstr, Grex, AUCC, Silhouette, C-index, PBM, WB, DBCV and CDbw.

\afterpage{
\clearpage

\begin{landscape}
\begin{table}
\centering
\resizebox{0.9\paperheight}{!}{%
\begin{tabular}{rllllllll}
  \hline
 Rank & K-Means & Single & Average & Complete & Ward & EM-GMM & Spectral & HDBSCAN* \\ 
  \hline
1 & Silhouette (0.665) & CDbw (0.877) & CDbw (0.822) & CDbw (0.758) & Silhouette (0.666) & Point-Biserial (0.607) & WB (0.799) & Silhouette (0.882) \\ 
  2 & Point-Biserial (0.652) & DBCV (0.824) & Point-Biserial (0.719) & Point-Biserial (0.684) & Point-Biserial (0.586) & VRC (0.582) & Silhouette (0.775) & Wemmert-Gancarski (0.879) \\ 
  3 & CDbw (0.641) & WB (0.767) & VRC (0.709) & Silhouette (0.634) & CDbw (0.57) & DBCV (0.576) & Tau (0.769) & Point-Biserial (0.847) \\ 
  4 & Wemmert-Gancarski (0.599) & VRC (0.763) & DBCV (0.702) & WB (0.609) & DBCV (0.544) & WB (0.564) & VRC (0.766) & DBCV (0.843) \\ 
  5 & DBCV (0.587) & Point-Biserial (0.737) & WB (0.699) & VRC (0.594) & VRC (0.543) & Silhouette (0.561) & Wemmert-Gancarski (0.696) & WB (0.842) \\ 
  6 & WB (0.557) & C-index (0.67) & Silhouette (0.668) & DBCV (0.584) & Wemmert-Gancarski (0.527) & Tau (0.54) & PBM (0.573) & AUCC (0.84) \\ 
  7 & VRC (0.533) & Silhouette (0.639) & C-index (0.641) & Wemmert-Gancarski (0.544) & WB (0.506) & CDbw (0.497) & C-index (0.565) & S\_Dbw (0.84) \\ 
  8 & Grex (0.436) & AUCC (0.62) & Wemmert-Gancarski (0.61) & Grex (0.508) & PBM (0.324) & PBM (0.42) & Point-Biserial (0.485) & VRC (0.814) \\ 
  9 & PBM (0.395) & Wemmert-Gancarski (0.594) & AUCC (0.591) & C-index (0.486) & C-index (0.301) & C-index (0.414) & Grex (0.475) & CDbw (0.808) \\ 
  10 & AUCC (0.387) & G(+) (0.489) & Grex (0.533) & AUCC (0.464) & Tau (0.283) & AUCC (0.404) & AUCC (0.456) & XieBeni (0.808) \\ 
  11 & C-index (0.385) & DB (0.486) & DB (0.497) & Gstr (0.442) & AUCC (0.263) & Wemmert-Gancarski (0.382) & CDbw (0.435) & C-index (0.805) \\ 
  12 & CS (0.35) & PBM (0.463) & G(+) (0.489) & CS (0.424) & CS (0.25) & G(+) (0.349) & DBCV (0.435) & G(+) (0.793) \\ 
  13 & G(+) (0.327) & Grex (0.449) & PBM (0.447) & G(+) (0.411) & SD (0.234) & Grex (0.187) & Ratkowsky-Lance (0.427) & CS (0.785) \\ 
  14 & Gstr (0.326) & Tau (0.417) & XieBeni (0.433) & SV (0.403) & Grex (0.227) & LCCV (0.155) & Gstr (0.334) & SV (0.779) \\ 
  15 & SV (0.31) & XieBeni (0.395) & CS (0.399) & DB (0.37) & SV (0.211) & CS (0.148) & SD (0.279) & DB (0.761) \\ 
  16 & DB (0.255) & CS (0.331) & Gstr (0.372) & PBM (0.332) & G(+) (0.181) & SV (0.142) & CS (0.274) & Grex (0.721) \\ 
  17 & S\_Dbw (0.191) & S\_Dbw (0.322) & SV (0.366) & S\_Dbw (0.288) & LCCV (0.168) & Ratkowsky-Lance (0.0805) & CVNN (0.258) & Dunn (0.623) \\ 
  18 & SD (0.162) & SV (0.307) & S\_Dbw (0.338) & SD (0.256) & XieBeni (0.168) & S\_Dbw (0.0786) & CVDD (0.24) & PBM (0.586) \\ 
  19 & LCCV (0.161) & LCCV (0.269) & LCCV (0.264) & Tau (0.203) & Ratkowsky-Lance (0.147) & Gstr (0.0583) & DB (0.226) & Gstr (0.553) \\ 
  20 & Tau (0.159) & Gstr (0.253) & CVDD (0.188) & LCCV (0.198) & SSDD (0.146) & Dunn (0.0309) & XieBeni (0.222) & SD (0.547) \\ 
  21 & XieBeni (0.0704) & Ratkowsky-Lance (0.129) & SD (0.187) & XieBeni (0.181) & Gstr (0.114) & SD (0.0274) & SSDD (0.211) & Tau (0.431) \\ 
  22 & CVDD (0.0631) & Dunn (0.124) & Dunn (0.13) & CVDD (0.136) & DB (0.11) & XieBeni (0.0203) & SV (0.205) & SSDD (0.337) \\ 
  23 & Dunn (-0.0306) & CVDD (0.045) & Tau (0.107) & Dunn (0.0451) & CVDD (0.0984) & DB (-0.00621) & LCCV (0.2) & LCCV (0.33) \\ 
  24 & Ratkowsky-Lance (-0.0554) & SD (0.0145) & Ratkowsky-Lance (-0.0212) & Ratkowsky-Lance (-0.115) & S\_Dbw (0.0767) & SSDD (-0.0875) & S\_Dbw (0.144) & CVDD (0.189) \\ 
  25 & SSDD (-0.0584) & CVNN (-0.0567) & CVNN (-0.0509) & SSDD (-0.161) & Dunn (-0.0148) & CVDD (-0.106) & G(+) (0.0365) & Ratkowsky-Lance (0.142) \\ 
  26 & CVNN (-0.213) & SSDD (-0.28) & SSDD (-0.265) & CVNN (-0.175) & CVNN (-0.0279) & CVNN (-0.139) & Dunn (-0.0176) & CVNN (-0.369) \\ 
   \hline
\end{tabular}}
\caption{\label{Table:Exp1Alg} Ranking of each index by mean correlation (within brackets) separately for each clustering algorithm.}
\end{table}
\end{landscape}

\begin{landscape}
\begin{table}
\centering
\resizebox{0.9\paperheight}{!}{%
\begin{tabular}{rllllllllllllll}
  \hline
 Index & All & Overlap & Imbalance & Noise & $D_{low}$ & $D_{high}$ & $k_{low}$ & $k_{high}$ & Compact & Sparse & Uniform & Gaussian & Logistic \\ 
  \hline
  Silhouette & 0.686 (1) & 0.531 (3) & 0.683 (1) & 0.641 (1) & 0.705 (1) & 0.644 (3) & 0.631 (1) & 0.811 (6) & 0.786 (2) & 0.591 (2) & 0.718 (2) & 0.695 (1) & 0.633 (1) \\ 
  WB & 0.662 (2) & 0.593 (1) & 0.663 (2) & 0.565 (4) & 0.66 (5) & 0.665 (1) & 0.575 (4) & 0.859 (1) & 0.707 (6) & 0.628 (1) & 0.712 (4) & 0.654 (2) & 0.603 (2) \\ 
  VRC & 0.657 (3) & 0.504 (4) & 0.648 (3) & 0.521 (7) & 0.68 (3) & 0.603 (5) & 0.592 (3) & 0.803 (8) & 0.757 (4) & 0.565 (3) & 0.738 (1) & 0.651 (5) & 0.552 (7) \\ 
  Point-Biserial & 0.656 (4) & 0.481 (6) & 0.641 (5) & 0.563 (5) & 0.674 (4) & 0.615 (4) & 0.569 (5) & 0.853 (2) & 0.815 (1) & 0.509 (7) & 0.715 (3) & 0.651 (4) & 0.581 (5) \\ 
  CDbw & 0.644 (5) & 0.484 (5) & 0.648 (4) & 0.465 (9) & 0.682 (2) & 0.552 (7) & 0.594 (2) & 0.748 (10) & 0.774 (3) & 0.545 (4) & 0.693 (5) & 0.651 (3) & 0.563 (6) \\ 
  DBCV & 0.624 (6) & 0.462 (7) & 0.63 (6) & 0.609 (2) & 0.608 (7) & 0.66 (2) & 0.531 (6) & 0.835 (4) & 0.739 (5) & 0.511 (6) & 0.65 (6) & 0.631 (6) & 0.581 (4) \\ 
  Wemmert-Gancarski & 0.602 (7) & 0.532 (2) & 0.619 (7) & 0.591 (3) & 0.616 (6) & 0.57 (6) & 0.498 (8) & 0.835 (3) & 0.684 (7) & 0.525 (5) & 0.601 (7) & 0.614 (7) & 0.588 (3) \\ 
  C-index & 0.521 (8) & 0.413 (8) & 0.532 (8) & 0.533 (6) & 0.539 (8) & 0.481 (8) & 0.388 (10) & 0.824 (5) & 0.618 (10) & 0.452 (8) & 0.556 (8) & 0.518 (8) & 0.478 (8) \\ 
  AUCC & 0.488 (9) & 0.321 (9) & 0.502 (9) & 0.484 (8) & 0.53 (9) & 0.395 (10) & 0.35 (11) & 0.802 (9) & 0.642 (8) & 0.361 (9) & 0.544 (10) & 0.488 (9) & 0.412 (9) \\ 
  PBM & 0.439 (10) & 0.25 (16) & 0.436 (11) & 0.266 (18) & 0.518 (10) & 0.264 (19) & 0.397 (9) & 0.537 (16) & 0.62 (9) & 0.288 (13) & 0.545 (9) & 0.424 (11) & 0.314 (15) \\ 
  Grex & 0.436 (11) & 0.297 (12) & 0.455 (10) & 0.292 (17) & 0.494 (11) & 0.306 (14) & 0.337 (12) & 0.661 (14) & 0.595 (11) & 0.329 (11) & 0.479 (11) & 0.441 (10) & 0.371 (10) \\ 
  G(+) & 0.365 (12) & 0.227 (17) & 0.373 (14) & 0.338 (11) & 0.389 (12) & 0.311 (13) & 0.171 (21) & 0.803 (7) & 0.512 (12) & 0.251 (17) & 0.403 (13) & 0.357 (13) & 0.322 (13) \\ 
  CS & 0.365 (13) & 0.305 (10) & 0.404 (12) & 0.343 (10) & 0.376 (13) & 0.34 (11) & 0.219 (18) & 0.697 (12) & 0.452 (14) & 0.296 (12) & 0.411 (12) & 0.35 (14) & 0.318 (14) \\ 
  Tau & 0.364 (14) & 0.293 (13) & 0.373 (15) & 0.31 (15) & 0.329 (17) & 0.443 (9) & 0.504 (7) & 0.0495 (21) & 0.376 (17) & 0.345 (10) & 0.371 (16) & 0.37 (12) & 0.349 (11) \\ 
  SV & 0.334 (15) & 0.283 (14) & 0.375 (13) & 0.325 (14) & 0.336 (16) & 0.328 (12) & 0.18 (20) & 0.685 (13) & 0.416 (15) & 0.268 (15) & 0.381 (15) & 0.315 (16) & 0.291 (17) \\ 
  DB & 0.323 (16) & 0.279 (15) & 0.347 (16) & 0.296 (16) & 0.339 (15) & 0.286 (17) & 0.201 (19) & 0.6 (15) & 0.406 (16) & 0.276 (14) & 0.312 (17) & 0.332 (15) & 0.326 (12) \\ 
  Gstr & 0.304 (17) & 0.133 (19) & 0.314 (17) & 0.149 (22) & 0.35 (14) & 0.199 (20) & 0.222 (17) & 0.488 (17) & 0.49 (13) & 0.206 (19) & 0.385 (14) & 0.284 (19) & 0.215 (19) \\ 
  XieBeni & 0.273 (18) & 0.217 (18) & 0.281 (19) & 0.338 (12) & 0.265 (19) & 0.291 (15) & 0.266 (15) & 0.287 (18) & 0.339 (19) & 0.207 (18) & 0.227 (21) & 0.301 (17) & 0.301 (16) \\ 
  S\_Dbw & 0.272 (19) & 0.301 (11) & 0.294 (18) & 0.33 (13) & 0.265 (18) & 0.288 (16) & 0.0777 (24) & 0.715 (11) & 0.273 (22) & 0.255 (16) & 0.25 (19) & 0.285 (18) & 0.287 (18) \\ 
  SD & 0.216 (20) & 0.122 (20) & 0.213 (20) & 0.197 (20) & 0.19 (21) & 0.272 (18) & 0.322 (14) & -0.0242 (23) & 0.274 (21) & 0.178 (20) & 0.225 (22) & 0.22 (20) & 0.197 (20) \\ 
  LCCV & 0.211 (21) & 0.0532 (22) & 0.202 (21) & 0.208 (19) & 0.237 (20) & 0.144 (22) & 0.226 (16) & 0.18 (19) & 0.367 (18) & 0.0985 (21) & 0.305 (18) & 0.186 (21) & 0.142 (22) \\ 
  CVDD & 0.109 (22) & -0.0507 (24) & 0.118 (22) & -0.0734 (25) & 0.176 (22) & -0.0406 (25) & 0.0971 (23) & 0.137 (20) & 0.287 (20) & -0.0261 (25) & 0.232 (20) & 0.0635 (24) & -0.0142 (24) \\ 
  Dunn & 0.0994 (23) & 0.0842 (21) & 0.114 (23) & 0.194 (21) & 0.0866 (23) & 0.128 (23) & 0.127 (22) & 0.0358 (22) & 0.132 (24) & 0.0813 (22) & 0.056 (24) & 0.113 (22) & 0.143 (21) \\ 
  Ratkowsky-Lance & 0.0945 (24) & 0.0104 (23) & 0.0714 (24) & -0.0338 (23) & 0.0565 (24) & 0.179 (21) & 0.331 (13) & -0.442 (26) & 0.12 (25) & 0.0789 (23) & 0.138 (23) & 0.0839 (23) & 0.0473 (23) \\ 
  SSDD & 0.0123 (25) & -0.0572 (25) & 0.00882 (25) & -0.0392 (24) & 0.0277 (25) & -0.0256 (24) & 0.0454 (26) & -0.107 (24) & 0.0292 (26) & -0.00226 (24) & 0.0456 (25) & 0.000959 (25) & -0.0259 (25) \\ 
  CVNN & -0.0898 (26) & -0.35 (26) & -0.114 (26) & -0.378 (26) & 0.014 (26) & -0.344 (26) & 0.0724 (25) & -0.419 (25) & 0.212 (23) & -0.33 (26) & 0.0367 (26) & -0.14 (26) & -0.226 (26) \\ 
   \hline
\end{tabular}}
\caption{\label{Table:CorProp} Ranking of each index (within brackets) by mean Spearman correlation, separately for the various properties of the Type 2 datasets in the Evaluation Scenario 1, categorized as follows: $k_{high}$ incorporates data with more than 10 ground-truth clusters; $k_{low}$ comprises data with 10 or fewer ground-truth clusters; $D_{low}$ refers to datasets with 25 or fewer dimensions; $D_{high}$ incorporates datasets with more than 25 dimensions; Overlap considers datasets with greater than 0 overlap using Equation \ref{eq:overlap}; Imbalanced datasets feature an imbalance of 0.5 or greater using Equation \ref{eq:imbalance}; Noise considers all datasets containing noise in the ground-truth partition; Compact clusters are defined as being generated with a compactness of 0.1 in MDCGen; Sparse clusters are defined as being generated with a compactness of 0.8 in MDCGen.}
\end{table}
\end{landscape}
}

We conjecture that this behaviour is explained by the fact that the added dimensions are informative, making the clustering problem initially easier, which is reflected in better performance of the indexes, until a point beyond which the benefit of adding further dimensions is outweighed by the detrimental effects of the curse of dimensionality, in particular, the dramatic reduction in distance contrast typically observed in high dimensional spaces. Some datasets are more affected than others in that sense, as it becomes difficult to discriminate e.g. within-cluster from between-cluster distances, causing clustering partitions also to be more difficult to discriminate in terms of quality, with a consequent drop in correlation between the internal and external evaluations.  

This aspect was found to be largely reliant on the clustering algorithm's ability to produce reasonable partitions for the internal validity indexes to rank, as most internal validity indexes were positively affected by high dimensionality when good partitions were produced. This can be clearly seen in Figure \ref{fig:CorDims}, where for datasets in which the Silhouette index could correctly identify the optimal partition, the correlation saw a significant increase with dimensionality; in contrast, when the index could not identify the best partition, often due to either partitions of similar quality being produced or no quality partitions being produced, there was a large decrease in correlation. As a result, all internal validity indexes, regardless of whether performance net increased or decreased, tended to perform less consistently in higher dimensions as compared to in lower dimensions. This also highlights the importance of the clustering algorithm in cluster validity, as it shows that \emph{the performance of an internal validity index is, at least in part, dependent on the performance of the clustering algorithm}.

The {\bf number of clusters} in the ground-truth partition was seen to have a sizeable effect on the performance of the majority of internal validity indexes, even more prominently than dimensionality, with the performance of most indexes showing a statistically significant positive correlation with the number of ground-truth clusters. These indexes were, G(+), Point-Biserial, Wemmert-Gancarski, Gstr, Grex, AUCC, Silhouette, C-index, VRC, DB, PBM, XieBeni, WB, CVDD, DBCV, CDbw, SV, CS and S\_Dbw. In contrast, the performance of the indexes Ratkowsky-Lance, SD, Tau, Dunn, CVNN, SSDD and LCCV, all showed a statistically significant negative correlation with the number of clusters. These results need to be considered with caution though. The reason is that, for datasets with a small number of clusters in the ground-truth, we can often only produce a limited set of candidate clustering partitions with fewer clusters than the optimal one, while there is much more flexibility in producing partitions with more clusters than the optimal one. This imbalance has the potential to bias the results in terms of overall correlation.

To account for the difference in how each index performs when presented with partitions containing either fewer or greater clusters as compared to the best (``Top Pick'') partition, this test was repeated, now considering the correlation for each of these two cases separately. The indexes Ratkowsky-Lance and CVNN show a statistically significant negative correlation with an increasing number of clusters in the ground truth for both cases, while the indexes SD, TAU, PBM, CDbw, and LCCV are only negatively correlated with statistical significance in the first case, i.e., when considering partitions with fewer clusters compared to the best partition. This may indicate for the latter group of indexes that performance tend to decrease as the number of natural clusters in the dataset increases, since they perform poorly when the number of partitions is underestimated, and the number of underestimated partitions increases as the number of ground-truth clusters increases. This matches the results seen in Table~ \ref{Table:Exp1CorType2Var}, except for CDbw. Apart from the aforementioned indexes, all other indexes were positively correlated with an increasing number of clusters for both cases of underestimation and overestimation of the number of clusters.

Several indexes, despite performing better as the number of clusters in the ground-truth partition increased, still saw notable reduction in their relative ranks due to other indexes showing significantly larger improvements in performance. This is notable for Silhouette, VRC, CDbw and PBM.

Testing the performance of each index using a Kruskal-Wallis test on the two levels of {\bf compactness} showed that, as expected, all indexes performed better (with statistical significance) in clustering problems with compact clusters, as compared to more sparse clusters. The different extent to which each index improved, however, caused some prominent changes in ranking among top performing indexes; noticeably, Point-Biserial performed the best for compact clusters, however, dropped to seventh best for sparse clusters. Conversely, WB performed the best for sparse clusters, but only sixth best for compact clusters. Most other indexes saw either minor or no change to their relative ranks when comparing sparse and compact clusters.

\subsection{Evaluation Scenario 2} \label{sec:eval2}

Evaluation Scenario 2 aims to determine the ability of each index to correctly determine the best partition when presented with several partitions containing the same number of clusters, using the methodology described in Section \ref{Method2}. Due to limitations in the Type 1 datasets, it was not possible to produce sufficiently diverse partitions with the same number of clusters for the majority of the datasets. For this reason, Evaluation Scenario 2 features only the Type 2 data collection. In addition, part of the Type 2 collection could not be used for similar reasons (too few unique partitions), resulting in a subset of 12647 datasets (out of 16177) adopted for the Evaluation Scenario 2.

The results are summarised in Tables \ref{Table:Exp2Correct} and \ref{Table:Exp2Cor}, which report the percentage of datasets for which each index correctly identified the ``Top Pick'' partition and the correlation between external and internal indexes, respectively. Table \ref{Table:Exp2Correct} shows results separately for the three following scenarios: partitions containing (i) the correct number of clusters ($k=k^*$); (ii) 30\% fewer clusters than the ground truth ($k<k^*$); and (iii) 30\% more clusters than the ground truth ($k>k^*$). We can see that indexes such as Point-Biserial, Wemmert-Gancarski, Silhouette, DBCV, VRC and WB, which performed well in Evaluation Scenario 1, also performed well for Evaluation Scenario 2, however, with a moderate change to the overall ranking. The percentage of cases where the ``Top Pick'' partition is identified is also similar between Evaluation Scenarios 1 and 2 when the number of clusters match the ground-truth number of cluster ($k=k^*$). However, when the number of clusters was incorrect, we observe that most indexes performed significantly worse, with many indexes that previously performed well failing to identify the ``Top Pick'' partition in this more adverse scenario. Indeed, only a small selection of indexes provided reasonable performance when $k \neq k^ *$, most noticeably, Point-Biserial when $k < k^*$ and Wemmert-Gancarski as well as DBCV when $k > k^*$. In relative terms, most indexes performed better at identifying the ``Top Pick'' partition when $k > k^*$ compared to $k < k^*$, with the exceptions of Point-Biserial, WB, VRC and SSDD.

Overall, Point-Biserial performed the best at identifying the ``Top Pick'' partition, performing best when the number of clusters were either correctly identified or underestimated, with reasonable performance (despite a low relative rank) when the number of clusters were overestimated. The indexes Wemmert-Gancarski, Silhouette, WB, VRC and DBCV also all performed well across multiple categories.

\begin{table}[t]
\centering
\begin{tabular}{lrrr}
  \hline
  Index & $k = k^*$ & $k < k^*$ & $k > k^*$ \\ 
  \hline
    Wemmert- & & & \\ 
    Gancarski & 56.6\%(2) & 26.9\%(6) & 42.3\%(1) \\ 
  Silhouette & 55.5\%(3) & 34\%(4) & 38.7\%(3) \\ 
  Point- & & & \\ 
  Biserial & 59.7\%(1) & 54.4\%(1) & 31.9\%(14) \\ 
  DBCV & 50.9\%(6) & 20.6\%(9) & 41\%(2) \\ 
  VRC & 53.8\%(5) & 35.6\%(3) & 32.3\%(12) \\ 
  WB & 54.9\%(4) & 36.5\%(2) & 29.9\%(16) \\ 
  LCCV & 43.2\%(14) & 24.9\%(8) & 35.2\%(5) \\ 
  PBM & 46.5\%(8) & 26.7\%(7) & 32.2\%(13) \\ 
  XieBeni & 45\%(11) & 17.5\%(11) & 33.9\%(7) \\ 
  AUCC & 45.2\%(10) & 16.4\%(13) & 33.8\%(8) \\ 
  C-index & 44.1\%(13) & 16.3\%(14) & 35.6\%(4) \\ 
  Ratkowsky- & & & \\ 
  Lance & 29.9\%(21) & 32.7\%(5) & 34.3\%(6) \\ 
  G(+) & 44.4\%(12) & 17\%(12) & 33.4\%(9) \\ 
  Grex & 46\%(9) & 15\%(15) & 32.9\%(11) \\ 
  CDbw & 49.1\%(7) & 10.1\%(19) & 25.7\%(19) \\ 
  Dunn & 37.4\%(17) & 10.8\%(18) & 33.2\%(10) \\ 
  Gstr & 38.3\%(16) & 11.3\%(16) & 26.2\%(18) \\ 
  CVNN & 40\%(15) & 11\%(17) & 23.3\%(22) \\ 
  SSDD & 29.2\%(23) & 20.6\%(10) & 8.61\%(25) \\ 
  SD & 29.5\%(22) & 7.26\%(22) & 31.2\%(15) \\ 
  CS & 34.8\%(18) & 5.29\%(23) & 24.7\%(20) \\ 
  DB & 28.6\%(24) & 7.86\%(21) & 27.9\%(17) \\ 
  SV & 33.7\%(19) & 5.06\%(24) & 23.9\%(21) \\ 
  CVDD & 31.7\%(20) & 4.79\%(25) & 16.7\%(23) \\ 
  Tau & 14.2\%(25) & 8.74\%(20) & 14.9\%(24) \\ 
  S\_Dbw & 12.1\%(26) & 1.32\%(26) & 3.13\%(26) \\ 
   \hline
\end{tabular}
\caption{\label{Table:Exp2Correct}
The percentage of cases for which both the external and internal validity indexes selected the same
candidate partition as optimal (“Top Pick”) for each combination of clustering algorithm and dataset in Evaluation Scenario 2, where all candidate partitions have the same number of clusters, $k$. The ground-truth number of clusters in each dataset is denoted by $k^*$. Relative ranks are displayed within brackets.}
\end{table}

\begin{table*}[h]
\centering
\resizebox{0.95\textwidth}{!}{%
\begin{tabular}{llll|lll}
  \hline
 &  & Mean & & & Median &  \\
  Index & $k = k^*$ & $k < k^*$ & $k > k^*$ & $k = k^*$ & $k < k^*$ & $k > k^*$ \\ 
  \hline
Point-Biserial & 0.862 (1) & 0.871 (1) & 0.638 (1) & 0.932 (1) & 0.92 (1) & 0.77 (2) \\ 
  WB & 0.737 (2) & 0.699 (2) & 0.625 (2) & 0.879 (2) & 0.826 (2) & 0.77 (1) \\ 
  VRC & 0.729 (3) & 0.695 (3) & 0.621 (3) & 0.871 (3) & 0.819 (3) & 0.763 (3) \\ 
  Ratkowsky-Lance & 0.675 (4) & 0.669 (4) & 0.605 (4) & 0.787 (6) & 0.797 (4) & 0.735 (7) \\ 
  Wemmert-Gancarski & 0.619 (5) & 0.53 (6) & 0.594 (5) & 0.847 (5) & 0.7 (6) & 0.762 (4) \\ 
  Silhouette & 0.57 (6) & 0.541 (5) & 0.552 (7) & 0.861 (4) & 0.785 (5) & 0.759 (5) \\ 
  DBCV & 0.559 (7) & 0.414 (7) & 0.559 (6) & 0.774 (7) & 0.531 (8) & 0.736 (6) \\ 
  PBM & 0.452 (9) & 0.367 (8) & 0.483 (10) & 0.755 (8) & 0.64 (7) & 0.687 (10) \\ 
  G(+) & 0.427 (11) & 0.339 (11) & 0.512 (8) & 0.585 (12) & 0.418 (12) & 0.711 (8) \\ 
  C-index & 0.394 (13) & 0.315 (13) & 0.486 (9) & 0.637 (10) & 0.451 (10) & 0.699 (9) \\ 
  Grex & 0.429 (10) & 0.338 (12) & 0.38 (13) & 0.587 (11) & 0.391 (13) & 0.449 (13) \\ 
  XieBeni & 0.372 (15) & 0.358 (9) & 0.385 (12) & 0.499 (15) & 0.472 (9) & 0.461 (12) \\ 
  CDbw & 0.498 (8) & 0.241 (16) & 0.323 (14) & 0.668 (9) & 0.305 (16) & 0.44 (14) \\ 
  LCCV & 0.374 (14) & 0.345 (10) & 0.282 (17) & 0.471 (16) & 0.431 (11) & 0.339 (18) \\ 
  AUCC & 0.302 (17) & 0.111 (19) & 0.39 (11) & 0.548 (13) & 0.216 (18) & 0.677 (11) \\ 
  CVNN & 0.414 (12) & 0.285 (14) & 0.062 (24) & 0.529 (14) & 0.345 (14) & 0.0477 (24) \\ 
  Tau & 0.14 (21) & 0.197 (17) & 0.301 (15) & 0.222 (18) & 0.297 (17) & 0.436 (15) \\ 
  DB & 0.239 (18) & 0.171 (18) & 0.294 (16) & 0.222 (19) & 0.169 (19) & 0.315 (19) \\ 
  SSDD & 0.322 (16) & 0.248 (15) & -0.0283 (25) & 0.381 (17) & 0.344 (15) & -0.0509 (25) \\ 
  CS & 0.122 (22) & -0.0426 (24) & 0.28 (18) & 0.133 (21) & -0.0438 (24) & 0.375 (16) \\ 
  SD & 0.167 (20) & 0.105 (20) & 0.235 (22) & 0.119 (22) & 0.102 (20) & 0.246 (22) \\ 
  CVDD & 0.178 (19) & 0.0391 (22) & 0.109 (23) & 0.177 (20) & 0.0372 (21) & 0.0886 (23) \\ 
  SV & 0.0781 (24) & -0.0753 (25) & 0.261 (19) & 0.0588 (23) & -0.094 (25) & 0.353 (17) \\ 
  Gstr & 0.113 (23) & 0.0379 (23) & 0.254 (20) & -0.00138 (24) & -0.028 (23) & 0.261 (21) \\ 
  Dunn & 0.0723 (25) & 0.0795 (21) & 0.235 (21) & -0.00681 (25) & 0.0128 (22) & 0.289 (20) \\ 
  S\_Dbw & -0.339 (26) & -0.346 (26) & -0.28 (26) & -0.641 (26) & -0.583 (26) & -0.53 (26) \\ 
   \hline
\end{tabular}}
\caption{\label{Table:Exp2Cor} Mean and Median Spearman correlation between external and internal validity indexes for Evaluation Scenario 2. Relative ranks are displayed within brackets, separately for each scenario involving a fixed $k$.}
\end{table*}

The mean and median correlation between the external and internal validity indexes are reported in Table \ref{Table:Exp2Cor}, where the rankings can be seen to be roughly similar to those of the percentage of ``Top Pick'' partitions identified for each case of fixed number of clusters, as reported in Table \ref{Table:Exp2Correct}, with the most noticeable change being the Ratkowsky-Lance index now performing among the best. Although the correlations for the cases where $k \neq k^*$ are lower than for $k = k^*$ for the majority of indexes, the relative differences in correlation across the three cases is notably smaller than the differences between the percentages of ``Top Pick'' partitions identified for the same scenarios in Table \ref{Table:Exp2Correct}.
This is frequently caused by the presence of multiple partitions similar to the ``Top Pick'' partition when $k \neq k^*$. In these cases, internal indexes may correctly discriminate good from bad partitions resulting in high overall correlation with the external evaluation, while these can still disagree on which the very best partition is, thus resulting in a low ``Top Pick'' performance.

As discussed in Section \ref{ExtInd}, external validity indexes may sometimes fail to correctly discriminate between better and worse partitions that are similar from a referential ground truth perspective, yet different from a geometric perspective, as these external indexes do not assess the geometry of partitions but instead only their labels. When all candidate partitions have the same number of clusters, this becomes more apparent as there tend to be more similarity between their labels. When considering an external index's ability to determine the ``Top Pick'' partition, they may still perform well when either the ground-truth partition or a similar candidate solution is available, which tends to be a reasonable assumption when $k = k^*$. However, when the ground-truth or similar partition cannot be present, such as when $k \neq k^*$, it is much more likely for external indexes to select as the ``Top Pick'' partition a candidate that may not be the best from a geometric point of view, since geometry is normally not accounted for. This may explain why there is lower agreement in the selection of the ``Top Pick'' partition between internal and external indexes for the cases of $k \neq k^*$, in comparison to the two cases of $k \neq k^*$ in Table \ref{Table:Exp2Correct}. The measure of correlation is less affected by this, resulting in a more consistent measure of an internal index's ability to identify good clustering partitions.

\begin{figure*}[h]
 \centering
\includegraphics[width=0.95\linewidth]{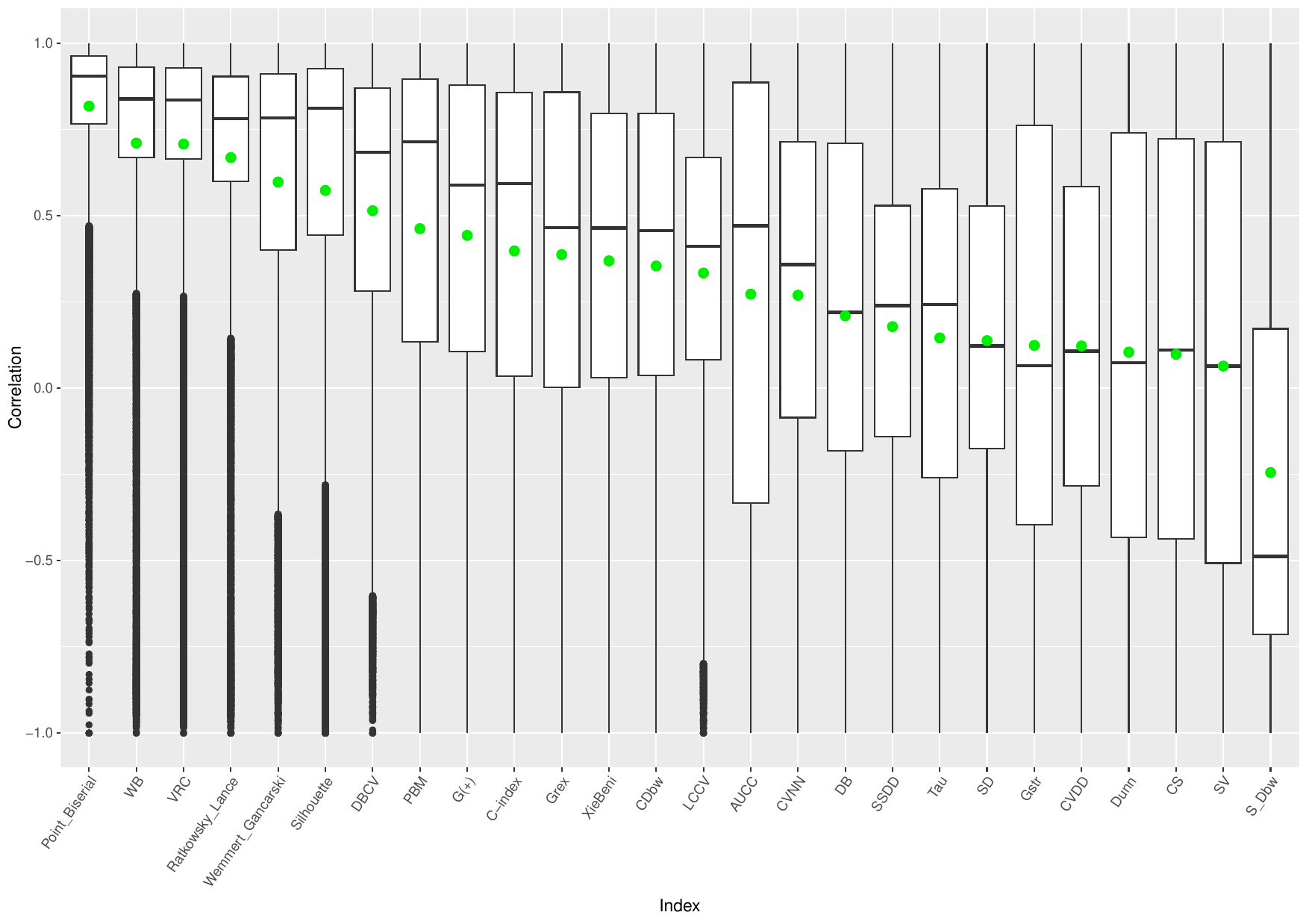}
\caption{\label{fig:RepCorDist} Boxplots of the Spearman correlations between the external and internal validity indexes in Evaluation Scenario 2 aggregated across all three cases ($k = k^*$, $k < k^*$, $k > k^*$), ordered by mean. Mean and median values are displayed with a green dot and a line through each segment, respectively.}
\end{figure*}

There are relatively few differences between the rankings of indexes by mean and median correlation in Table \ref{Table:Exp2Cor}, with once again, median correlations being slightly higher compared to the mean correlation due to the skewed distributions. Figure \ref{fig:RepCorDist} shows the distribution of correlations, where Point-Biserial, WB, VRC and Ratkowsky-Lance are seen to perform better, with both a higher mean correlation and smaller interquartile range compared to all other indexes. Wemmert-Gancarski and Silhouette perform similarly in median correlation to these indexes, however, their distribution of correlations has a larger range and lower mean, resulting in slightly worse performance.

The Ratkowsky-Lance index, which performed poorly across all results for Evaluation Scenario 1, performed significantly better for Evaluation Scenario 2, ranking among the best performing indexes in terms of identifying the best partition for cases with $k < k^*$ and $k > k^*$, in addition to having among the highest correlations across all three scenarios. There was evidence in Evaluation Scenario 1 that the Ratkowsky-Lance index is subject to significant bias towards the number of clusters, which is further supported by the index performing well under circumstances where the number of clusters is fixed.

Unlike Evaluation Scenario 1, there were no obvious signs of non-linear relationships between external and internal validity indexes. This result is expected as these relationships primarily appeared as a result of comparing partitions with differing numbers of clusters, and how the differences between over and under estimation of clusters are handled by both external and internal indexes. Additionally, there are fewer data-points due to limitations in producing diverse clustering solutions with a fixed number of clusters, which gives limited room for such patterns to emerge.

For Evaluation Scenario 2, a pairwise Willcoxon test with Bonferroni correction was performed on the combined results of the three different levels for the number of clusters ($k = k^*$, $k < k^*$, and $k > k^*$). This test, as described in Section \ref{stat}, is intended to determine if the differences in performance as measured by correlation is significant between indexes. At a 5\% significance level, all internal validity indexes were found to have a statistically significant difference in performance measured by correlation, with the exception of the following four pairs of indexes: (i) C-index and Grex; (ii) CS and CVDD; (iii) Gstr and SD; (iv) SD and Tau.

A second statistical testing for Evaluation Scenario 2, also performed on the aggregated results of the three levels for the number of clusters, has been carried out to analyse the impact of each data property on performance. This follows the same methods previously performed in Evaluation Scenario 1, and also described in Section \ref{stat}. Notice that dimensionality, number of clusters, and imbalance have been grouped for the purpose of displaying mean values of correlation across distinct groups for these properties in Table \ref{Table:RepProp}. These groupings, however, were not used for the purpose of statistical testing. 

The performance of all indexes were found to have a statistically significant negative correlation with the number of dimensions, with the exception of Point-Biserial and SSDD, which showed no statistically significant correlation. The same behaviour with increasing dimensionality, where the performance of several validity indexes would increase with dimensionality up to a certain number of dimsensions before becoming significantly more variable, previously noted in Evaluation Scenario 1 was also observed in Evaluation Scenario 2 for the same indexes. 

Similar to Evaluation Scenario 1, the performance of most indexes presented a statistically significant positive correlation with an increase in ground-truth clusters, with the exceptions of Tau, SSDD and LCCV, for which performance was negatively correlated.

Similar to Evaluation Scenario 1, roughly half the indexes had a statistically significant positive correlation with the imbalance of clusters measured by Equation \ref{eq:imbalance}, while the indexes G(+), Point-Biserial, SD, Grex, AUCC, Dunn, CVDD, CVNN, SSDD and S\_Dbw were negatively correlated. The indexes Gstr, DB and XieBeni showed no statistically significant correlation with the imbalance of clusters. 

All indexes were found to have a statistically significant negative correlation with the quantity of overlap measured by Equation \ref{eq:overlap}, except for SSDD, which had a statistically significant positive correlation.
Testing Noise using a Kruskal-Wallis test, it was found noise had a statistically significant negative impact on the performance of all indexes with the exception of SSDD, which presented a statistically significant positive impact. Similarly, testing the effect of cluster compactness, all indexes performed better with compact clusters at the 5\% significance level, except for SSDD, which performed better with sparse clusters.

In Table \ref{Table:RepProp} we can see the correlation results for Evaluation Scenario 2 separated by specific properties within each dataset. A group of top-3 indexes systematically outperformed all other indexes, namely, Point-Biserial, WB and VRC in this order, except for the case of compact clusters, where Silhouette and Wemmert-Gancarski performed second and third best, respectively. Although the rankings \emph{outside} the best performing indexes (say, top-10) did change significantly with each property, the correlation performance of many of these indexes is low enough that the changes in their ranks makes no practical difference. The presence of overlapping clusters, high-dimensionality, sparsity, and noise had the largest impacts on performance across the majority of indexes, with most other properties having less noticeable effect. Regarding dimensionality, in contrast to Evaluation Scenario 1, most indexes performed significantly worse for high dimensional datasets compared to low dimensional ones. This is likely due to fewer geometric differences between partitions with the same number of clusters in high dimensions, due to the curse of dimensionality.

\afterpage{
\clearpage%
\begin{landscape}
\begin{table}
\centering
\resizebox{0.85\paperheight}{!}{%
\begin{tabular}{rllllllllllllll}
  \hline
 Index & All & Overlap & Imbalance & Noise & $D_{low}$ & $D_{high}$ & $k_{low}$ & $k_{high}$ & Compact & Sparse & Uniform & Gaussian & Logistic \\ 
   \hline
Point-Biserial & 0.785 (1) & 0.751 (1) & 0.777 (1) & 0.74 (1) & 0.777 (1) & 0.798 (1) & 0.769 (1) & 0.833 (1) & 0.851 (1) & 0.729 (1) & 0.793 (1) & 0.781 (1) & 0.779 (1) \\ 
  WB & 0.686 (2) & 0.509 (2) & 0.699 (2) & 0.658 (2) & 0.755 (2) & 0.56 (2) & 0.663 (2) & 0.757 (2) & 0.804 (4) & 0.6 (2) & 0.779 (2) & 0.718 (2) & 0.53 (2) \\ 
  VRC & 0.68 (3) & 0.506 (3) & 0.694 (3) & 0.643 (3) & 0.75 (3) & 0.553 (3) & 0.656 (3) & 0.754 (3) & 0.796 (5) & 0.596 (3) & 0.773 (3) & 0.711 (3) & 0.526 (3) \\ 
  Ratkowsky-Lance & 0.648 (4) & 0.487 (4) & 0.663 (4) & 0.566 (4) & 0.708 (6) & 0.54 (4) & 0.625 (4) & 0.72 (6) & 0.746 (8) & 0.574 (4) & 0.735 (5) & 0.676 (4) & 0.506 (4) \\ 
  Wemmert-Gancarski & 0.583 (5) & 0.305 (6) & 0.601 (5) & 0.515 (5) & 0.716 (4) & 0.343 (6) & 0.532 (5) & 0.741 (4) & 0.809 (3) & 0.42 (5) & 0.728 (6) & 0.583 (5) & 0.4 (5) \\ 
  Silhouette & 0.555 (6) & 0.19 (8) & 0.57 (6) & 0.49 (6) & 0.712 (5) & 0.271 (8) & 0.498 (6) & 0.731 (5) & 0.84 (2) & 0.354 (6) & 0.746 (4) & 0.558 (6) & 0.31 (6) \\ 
  DBCV & 0.515 (7) & 0.153 (10) & 0.527 (7) & 0.423 (7) & 0.572 (9) & 0.412 (5) & 0.461 (7) & 0.683 (7) & 0.767 (6) & 0.3 (7) & 0.703 (7) & 0.507 (7) & 0.289 (7) \\ 
  PBM & 0.437 (8) & 0.0667 (14) & 0.458 (8) & 0.38 (8) & 0.611 (7) & 0.125 (15) & 0.376 (8) & 0.629 (9) & 0.763 (7) & 0.209 (13) & 0.638 (8) & 0.452 (8) & 0.168 (13) \\ 
  G(+) & 0.431 (9) & 0.132 (11) & 0.437 (9) & 0.241 (12) & 0.566 (10) & 0.187 (10) & 0.36 (10) & 0.651 (8) & 0.693 (9) & 0.281 (8) & 0.603 (9) & 0.425 (9) & 0.222 (11) \\ 
  C-index & 0.403 (10) & 0.0605 (15) & 0.413 (10) & 0.214 (13) & 0.573 (8) & 0.0961 (16) & 0.357 (11) & 0.546 (10) & 0.692 (10) & 0.236 (12) & 0.596 (10) & 0.408 (10) & 0.154 (14) \\ 
  Grex & 0.384 (11) & 0.173 (9) & 0.387 (11) & 0.0744 (17) & 0.524 (11) & 0.131 (14) & 0.332 (12) & 0.544 (11) & 0.6 (13) & 0.254 (10) & 0.498 (13) & 0.379 (11) & 0.246 (10) \\ 
  XieBeni & 0.372 (12) & 0.0384 (16) & 0.38 (12) & 0.187 (14) & 0.502 (12) & 0.138 (12) & 0.33 (13) & 0.504 (12) & 0.677 (11) & 0.195 (14) & 0.556 (11) & 0.369 (12) & 0.147 (15) \\ 
  CDbw & 0.357 (13) & 0.111 (12) & 0.363 (13) & 0.311 (9) & 0.463 (14) & 0.161 (11) & 0.33 (14) & 0.443 (14) & 0.571 (14) & 0.182 (15) & 0.504 (12) & 0.324 (14) & 0.21 (12) \\ 
  LCCV & 0.332 (14) & 0.259 (7) & 0.344 (14) & 0.251 (11) & 0.375 (16) & 0.252 (9) & 0.373 (9) & 0.206 (23) & 0.437 (20) & 0.253 (11) & 0.376 (18) & 0.351 (13) & 0.273 (9) \\ 
  AUCC & 0.276 (15) & -0.134 (20) & 0.267 (15) & 0.18 (15) & 0.486 (13) & -0.101 (21) & 0.213 (18) & 0.471 (13) & 0.635 (12) & 0.0667 (19) & 0.494 (14) & 0.278 (15) & 0.00267 (20) \\ 
  CVNN & 0.249 (16) & 0.094 (13) & 0.257 (16) & 0.0856 (16) & 0.308 (19) & 0.137 (13) & 0.232 (16) & 0.3 (17) & 0.412 (21) & 0.135 (16) & 0.375 (19) & 0.206 (17) & 0.135 (16) \\ 
  DB & 0.238 (17) & -0.0533 (18) & 0.247 (17) & -0.0508 (21) & 0.391 (15) & -0.0384 (19) & 0.203 (19) & 0.347 (15) & 0.484 (17) & 0.0866 (18) & 0.431 (16) & 0.217 (16) & 0.0203 (18) \\ 
  Tau & 0.215 (18) & -0.0394 (17) & 0.24 (18) & -0.00783 (19) & 0.327 (17) & 0.0129 (17) & 0.248 (15) & 0.112 (25) & 0.395 (23) & 0.0985 (17) & 0.354 (21) & 0.205 (18) & 0.0515 (17) \\ 
  SSDD & 0.173 (19) & 0.321 (5) & 0.164 (20) & 0.296 (10) & 0.101 (25) & 0.3 (7) & 0.221 (17) & 0.0149 (26) & 0.0319 (25) & 0.265 (9) & 0.091 (25) & 0.167 (19) & 0.275 (8) \\ 
  SD & 0.172 (20) & -0.0828 (19) & 0.172 (19) & -0.0451 (20) & 0.278 (22) & -0.0188 (18) & 0.139 (20) & 0.278 (19) & 0.409 (22) & 0.0519 (20) & 0.314 (24) & 0.164 (20) & 0.00486 (19) \\ 
  Gstr & 0.141 (21) & -0.189 (22) & 0.146 (21) & -0.223 (25) & 0.281 (21) & -0.111 (22) & 0.089 (21) & 0.303 (16) & 0.453 (19) & -0.017 (21) & 0.336 (22) & 0.122 (21) & -0.0809 (21) \\ 
  Dunn & 0.133 (22) & -0.25 (23) & 0.139 (23) & -0.209 (24) & 0.313 (18) & -0.192 (25) & 0.0864 (22) & 0.277 (20) & 0.474 (18) & -0.0615 (22) & 0.365 (20) & 0.109 (22) & -0.131 (23) \\ 
  CS & 0.129 (23) & -0.326 (24) & 0.144 (22) & -0.0751 (22) & 0.287 (20) & -0.157 (23) & 0.08 (23) & 0.281 (18) & 0.504 (15) & -0.115 (24) & 0.432 (15) & 0.0836 (23) & -0.199 (24) \\ 
  CVDD & 0.111 (24) & -0.186 (21) & 0.109 (24) & 0.0259 (18) & 0.221 (24) & -0.0875 (20) & 0.0601 (24) & 0.271 (21) & 0.383 (24) & -0.0632 (23) & 0.315 (23) & 0.0639 (24) & -0.0889 (22) \\ 
  SV & 0.0971 (25) & -0.37 (25) & 0.108 (25) & -0.126 (23) & 0.255 (23) & -0.188 (24) & 0.0467 (25) & 0.254 (22) & 0.487 (16) & -0.156 (25) & 0.411 (17) & 0.048 (25) & -0.239 (25) \\ 
  S\_Dbw & -0.32 (26) & -0.549 (26) & -0.312 (26) & -0.554 (26) & -0.228 (26) & -0.489 (26) & -0.464 (26) & 0.136 (24) & -0.123 (26) & -0.456 (26) & -0.16 (26) & -0.346 (26) & -0.495 (26) \\ 
   \hline
\end{tabular}}
\caption{\label{Table:RepProp} Ranking of each index (within brackets) by mean Spearman correlation, separately for the various properties of the Type 2 datasets in the Evaluation Scenario 2, categorized as follows: $k_{high}$ incorporates data with more than 10 ground-truth clusters; $k_{low}$ comprises data with 10 or fewer ground-truth clusters; $D_{low}$ refers to datasets with 25 or fewer dimensions; $D_{high}$ incorporates datasets with more than 25 dimensions; Overlap considers datasets with greater than 0 overlap using Equation \ref{eq:overlap}; Imbalanced datasets feature an imbalance of 0.5 or greater using Equation \ref{eq:imbalance}; Noise considers all datasets containing noise in the ground-truth partition; Compact clusters are defined as being generated with a compactness of 0.1 in MDCGen; Sparse clusters are defined as being generated with a compactness of 0.8 in MDCGen.}
\end{table}
\end{landscape}
}

\subsection{Evaluation Scenario 3} \label{sec:eval3}

Recall that in Evaluation Scenario 3 candidate partitions are systematically produced independently of any clustering algorithm, and their quality --- in terms of the degree of departure from the ground truth --- follows a natural order that allows for their ranking to be produced independently of any external index. For the reasons previously justified in Section \ref{AlgInd}, the datasets used in conjunction with this experimental methodology are a subset of the Type 2 datasets limited to Gaussian clusters with no background noise.     

The results are displayed in Tables \ref{Table:Exp3Correct} and \ref{Table:Exp3Cor}. Table \ref{Table:Exp3Correct} contains the percentage of cases where each index correctly identifies the ``Top Pick'' partition, which is now the best candidate partition according to the \emph{corresponding referential ranking} (instead of an external index). The results are shown across four cases, namely, both Procedures 1 and 2, each with either a varied number of clusters or a fixed number of clusters. Recall that in Procedures 1 and 2 candidates \emph{other than the ground truth} contain either fewer ($k < k^*$) or more ($k > k^*$) clusters than the ground truth, respectively (see Section \ref{AlgInd}). The ground truth partition is only present in the ``Varied $k$'' case, where (by construction) it necessarily coincides with the ``Top-Pick'' partition. In addition to the ``Top Pick'' results in Table \ref{Table:Exp3Correct}, Table \ref{Table:Exp3Cor} displays the correlation between the known referential rankings and those produced by internal validity indexes, across the same aforementioned four cases.

\begin{table*}[ht]
\centering
\resizebox{0.79\textwidth}{!}{
\begin{tabular}{lrrrr}
  \hline
  & Procedure 1 &  Procedure 1 & Procedure 2 & Procedure 2 \\ 
 Index & Varied $k$ & Fixed $k$ & Varied $k$ & Fixed $k$ \\
  \hline
WB & 91.3\%(2) & 55.9\%(17.5) & 97.3\%(4) & 29.5\%(2.5) \\ 
  VRC & 89.7\%(4) & 55.9\%(17.5) & 98.4\%(3) & 29.5\%(2.5) \\ 
  Wemmert-Gancarski & 77.8\%(8.5) & 58.9\%(7) & 96.6\%(5) & 19.2\%(9) \\ 
  CS & 68.8\%(17) & 60.1\%(4) & 91.8\%(11) & 26.9\%(5) \\ 
  DB & 73.9\%(12) & 56.2\%(16) & 94\%(10) & 30.3\%(1) \\ 
  PBM & 91.4\%(1) & 55.2\%(20) & 86.1\%(17) & 26.3\%(6) \\ 
  C-index & 75.6\%(10) & 62.8\%(1) & 78\%(19) & 16.5\%(14) \\ 
  Grex & 83.6\%(5) & 55\%(21) & 94.2\%(9) & 16.7\%(13) \\ 
  DBCV & 71\%(14.5) & 61.6\%(3) & 91.7\%(12) & 13.9\%(19.5) \\ 
  Ratkowsky-Lance & 2.9\%(26) & 55.7\%(19) & 98.6\%(2) & 28.3\%(4) \\ 
  SV & 63\%(18) & 58.2\%(11) & 88.7\%(16) & 24.5\%(7) \\ 
  AUCC & 77.8\%(8.5) & 59.5\%(5) & 67.9\%(22) & 15.7\%(17) \\ 
  Gstr & 72.6\%(13) & 56.5\%(15) & 89.8\%(15) & 19\%(10) \\ 
  Dunn & 55.3\%(21) & 58.9\%(7) & 91.5\%(13) & 17.4\%(12) \\ 
  Silhouette & 75\%(11) & 57.2\%(13.5) & 91\%(14) & 16.4\%(15) \\ 
  XieBeni & 69.5\%(16) & 49.3\%(23) & 94.8\%(7) & 23.3\%(8) \\ 
  CDbw & 79.8\%(7) & 62.3\%(2) & 71.5\%(21) & 6.86\%(24.5) \\ 
  G(+) & 82.3\%(6) & 58.7\%(9) & 59.9\%(24) & 14.1\%(18) \\ 
  CVDD & 60.2\%(19) & 58.3\%(10) & 96.5\%(6) & 9.07\%(22) \\ 
  S\_Dbw & 90\%(3) & 57.9\%(12) & 72\%(20) & 9\%(23) \\ 
  CVNN & 57.2\%(20) & 58.9\%(7) & 94.7\%(8) & 6.86\%(24.5) \\ 
  SD & 19.4\%(24) & 39.7\%(24) & 99\%(1) & 18.5\%(11) \\ 
  Tau & 16.8\%(25) & 57.2\%(13.5) & 81.9\%(18) & 15.9\%(16) \\ 
  Point-Biserial & 71\%(14.5) & 50.2\%(22) & 67.3\%(23) & 11.5\%(21) \\ 
  SSDD & 38.9\%(23) & 19.2\%(26) & 33.2\%(26) & 13.9\%(19.5) \\ 
  LCCV & 50.7\%(22) & 20.9\%(25) & 34.6\%(25) & 6.79\%(26) \\ 
   \hline
\end{tabular}}
\caption{\label{Table:Exp3Correct} Percentage of cases in which an internal index selected the ``Top-Pick'' partition among a collection of candidates systematically produced according to the methodological procedures in Evaluation Scenario 3 (Section \ref{AlgInd}), where ``Top-Pick'' is the best candidate partition according to a referential ranking that is independent of any external index or clustering algorithm. In Procedures 1 and 2, candidates \emph{other than the ground truth} contain either fewer ($k < k^*$) or more ($k > k^*$) clusters than the ground truth, respectively. The ground truth partition is only present in the ``Varied $k$'' case, where it corresponds to the ``Top-Pick'' partition.}
\end{table*}

\begin{table*}[ht]
\centering
\begin{tabular}{lrrrr}
  \hline
   & Procedure 1 & Procedure 1 & Procedure 2 & Procedure 2 \\ 
  Index & Varied $k$ & Fixed $k$ & Varied $k$ & Fixed $k$ \\
  \hline
VRC & 0.94(5) & 0.61(9.5) & 0.98(2) & 0.36(2.5) \\ 
  WB & 0.96(3) & 0.61(9.5) & 0.97(4) & 0.36(2.5) \\ 
  PBM & 0.96(2) & 0.6(11) & 0.88(13) & 0.31(8) \\ 
  CS & 0.86(11) & 0.62(5) & 0.9(12) & 0.32(6) \\ 
  DB & 0.89(9) & 0.57(16) & 0.93(11) & 0.39(1) \\ 
  Wemmert-Gancarski & 0.9(8) & 0.55(19) & 0.98(3) & 0.29(9) \\ 
  SV & 0.81(15) & 0.63(4) & 0.87(14) & 0.31(7) \\ 
  Ratkowsky-Lance & -0.81(26) & 0.58(14) & 0.99(1) & 0.35(4) \\ 
  Grex & 0.94(6) & 0.58(15) & 0.95(6) & 0.096(19) \\ 
  C-index & 0.87(10) & 0.61(7) & 0.71(17) & 0.21(12) \\ 
  CVDD & 0.78(17) & 0.68(2) & 0.96(5) & -0.052(23) \\ 
  CDbw & 0.94(4) & 0.69(1) & 0.68(18) & -0.075(24) \\ 
  Silhouette & 0.86(12) & 0.55(18) & 0.95(7) & 0.21(11) \\ 
  Gstr & 0.79(16) & 0.6(12) & 0.95(8) & 0.16(15) \\ 
  AUCC & 0.86(13) & 0.61(8) & 0.6(20) & 0.2(13) \\ 
  CVNN & 0.62(19) & 0.66(3) & 0.95(9) & -0.12(26) \\ 
  G(+) & 0.92(7) & 0.6(13) & 0.29(24) & 0.15(16) \\ 
  Tau & -0.32(25) & 0.62(6) & 0.84(15) & 0.2(14) \\ 
  DBCV & 0.85(14) & 0.54(20) & 0.94(10) & 0.11(18) \\ 
  S\_Dbw & 0.98(1) & 0.57(17) & 0.57(21) & -0.097(25) \\ 
  XieBeni & 0.69(18) & 0.45(23) & 0.4(23) & 0.32(5) \\ 
  SD & -0.25(24) & 0.34(24) & 0.72(16) & 0.26(10) \\ 
  Dunn & 0.5(21) & 0.53(22) & 0.65(19) & 0.12(17) \\ 
  Point-Biserial & 0.57(20) & 0.54(21) & 0.53(22) & 0.085(20) \\ 
  LCCV & 0.3(23) & 0.1(25) & 0.27(25) & 0.0035(21) \\ 
  SSDD & 0.36(22) & 0.014(26) & 0.22(26) & -0.044(22) \\ 
   \hline
\end{tabular}
\caption{\label{Table:Exp3Cor} Mean Spearman correlation between the known referential ranks systematically induced by the partition generation procedures in Evaluation Scenario 3 (Section \ref{AlgInd}) and the partition ranks according to the internal validity indexes. The referential ranks of these partitions are \emph{independent of any external index or clustering algorithm}. In Procedures 1 and 2, candidates other than the ground truth (only present in the ``Varied $k$'' case) contain either fewer ($k < k^*$) or more ($k > k^*$) clusters than the ground truth, respectively.}
\end{table*}

Observing the ``Top Pick'' results in Table \ref{Table:Exp3Correct}, firstly for the two cases Procedure 1 with varied $k$ (column one) and Procedure 2 with varied $k$ (column three), we see that most indexes perform noticeably better in Evaluation Scenario 3 (being able to correctly identify the ``Top Pick'' partition with a high frequency) than in Evaluation Scenario 1 (see Table \ref{Table:Experiment1}, ``Top Pick Type 2'' results in column two). In relative terms, indexes such as WB, VRC and Wemmert-Gancarski, which were top performers in Evaluation Scenario 1, still appear among the best here in Evaluation Scenario 3, particularly when considering their aggregated ranks across Procedures 1 and 2 (for varied $k$). In contrast, the Silhouette index performed relatively poorly in this regard, dropping from previously the best in terms of ``Top Pick'' percentage in Evaluation Scenario 1 to now ranking eleventh and fourteenth for Procedures 1 and 2, respectively. While the respective results based on correlation in Table \ref{Table:Exp3Cor} (for varied $k$, in columns one and three) show some moderate changes in rankings, the overall conclusions remain similar, with WB, VRC and Wemmert-Gancarski appearing again as the best indexes in terms of their aggregated ranks, while the Silhouette index ranks twelfth and seventh for Procedure 1 and 2, respectively. In both measures of performance, ``Top Pick'' and correlation, most indexes show similar relative performance between Procedure 1 and 2, however some indexes such as CDbw, PBM, G(+), Ratkowsky-Lance and SD excel in one procedure, while performing significantly worse in the other. 

One possible cause of the discrepancy in the Silhouette ranks between Evaluation Scenarios 1 and 3 could be a result of external validity indexes favouring the Silhouette due to similar behaviours in ranking partitions with over- and under-estimated numbers of clusters. Recall in Evaluation Scenario 1 and Table \ref{Table:Exp1CorType2Var} that the Silhouette was noted to display less evidence of and impact from non-linear relationships with the external validity indexes. In contrast, WB, VRC and Wemmert-Gancarski were all seen to have notable non-linear relationships with external validity indexes and less consistency between the cases of under- and over-estimation of the number of clusters. The absence of the external indexes and such non-linear relationships may explain at least in part why these indexes outperform the Silhouette here. 

Another possible reason could be
that the absolute performance of the Silhouette in Evaluation Scenario 1 is more consistent across clustering algorithms than other validity indexes, being more comparable across most algorithms despite changes in relative rankings. In contrast, the performance of indexes such as WB, VRC and Wemmert-Gancarski varies significantly between algorithms. This indicates Silhouette is more robust to the different types of partitions produced, an aspect that is not captured here in Evaluation Scenario 3. 
The fact that it displayed better relative performance than other indexes for the more diverse Type 2 data collection in Evaluation Scenario 1 may also support the conjecture that its robustness to more diverse types of clustering solutions cannot be fully captured here in Evaluation Scenario 3, where synthetically generated, custom-tailored partitions are adopted.  

As noted above, some indexes were seen to perform well in one procedure but poorly in the other. Two primary causes can be determined for the case of a varied number of clusters. The first is due to the ``Top Pick’’ partition always being the one that, for Procedure 1, contains the highest number of clusters with decreasing partition quality as the number of clusters decreases, whereas for Procedure 2, it is the one containing fewest clusters, with decreasing partition quality as the number of clusters increases. This would result in any preference towards fewer or greater cluster granularity increasing performance in one procedure, while reducing performance in the other. Looking at their absolute performance, this is obvious for indexes such as Ratkowsky-Lance, SD and Tau, and this aspect likely contributes to other indexes performing better in one procedure over the other as well. 

Another cause may be due to the differences between how Procedures 1 and 2 produce partitions. Some indexes may have limited sensitivity to the manner by which partitions are either joined or split. In this case, it would indicate an index is either not sensitive enough to changes in compactness for Procedure 1, or separation for Procedure 2. Either way, this is an undesirable behaviour captured here in Evaluation Scenario 3. 

Looking at the results for the cases where the number of clusters were fixed, let's first consider the ``Top Pick'' percentage evaluation in Table \ref{Table:Exp3Correct} (columns two and four), where we see the majority of indexes performing better with Procedure 1 ($k < k^*$) than with Procedure 2 ($k > k^*$). This may be in part due to Procedure 1, by joining different clusters to produce solutions in contrast to splitting clusters along different axes, potentially being able to more closely mimic real-world clustering solutions when the number of clusters is fixed as compared to Procedure 2. Similar to Evaluation Scenarios 1 and 2, indexes appear to perform worse when presented with solutions containing the same number of clusters in comparison to a varied number of clusters, likely due to a limited diversity among partitions which all contain the same number of clusters. In terms of rankings, the CS index performs the best overall with respect to ``Top Pick'' performance and fixed number of clusters, but it should be noted that in absolute terms several other indexes such as WB, VRC, DB, Ratkowsky-Lance, SV and PBM also exhibit highly competitive performances in this regard.

In terms of correlation for the case of a fixed number of clusters, the results in Table \ref{Table:Exp3Cor}(columns two and four) are somewhat similar to the ``Top Pick'' evaluation. The CS index still performs the best overall, however, VRC, WB, SV and PBM now appear much closer in comparison, all sharing comparable performances. 

In contrast to Evaluation Scenario 2, where Point-Biserial exhibited top performance, in Evaluation Scenario 3 it performed poorly according to both measures (``Top Pick'' and correlation) and under both procedures when the number of clusters is fixed. One aspect of this may be due to the limited dimensionality of the datasets produced by Procedure 1, as this index is frequently seen to perform poorly in lower dimensional problems, as noted in Evaluation Scenario 1. However, this cannot explain the results for Procedure 2. Another potential cause for this discrepancy is due to the behaviour of the external validity indexes used in Evaluation Scenario 2. Measuring the correlation between CDistance and Point-Biserial for Procedure 1 with a fixed number of clusters, we find a high Spearman correlation of 0.96. Given the weak performance of Point-Biserial in this case here in Scenario 3 (where correlation is computed against the referential ranking, rather than against CDistance's ranking), this suggests evidence that there may be an external index related bias towards Point-Biserial in Evaluation Scenario 2.

Similar to the case of a varied number of clusters, we again see several indexes that perform well in one procedure but rather poorly according to the other procedure, such as CDbw, G(+), C-index, XieBeni, DBCV and AUCC. Unlike for the case of a varied number of clusters, this cannot be a bias relating to the number of clusters present in the clustering solutions as all solutions contained the same number of clusters. This means the primary cause of this discrepancy is each index's ability to differentiate between partitions with varying compactness for Procedure 1, and varying separation for Procedure 2. 

Comparing the performance of indexes for each procedure in the cases of varied or fixed number of clusters, we can gain insight into their behaviours, either relating to the number of clusters or to a specific procedure. For the Ratkowsky-Lance index, when the number of clusters varied the index performed poorly for Procedure 1 by both measures of performance in Tables \ref{Table:Exp3Correct} and \ref{Table:Exp3Cor} (column one), but among the best for Procedure 2 (column three), and it also exhibited good performance with both procedures when the number of clusters were fixed (columns two and four). It is clear this is primarily a result of bias towards the number of clusters in a solution. In comparison, for SD and XieBeni using ``Top Pick'' performance in Table \ref{Table:Exp3Correct}, both performed poorly in both cases of Procedure 1, i.e., varied and fixed number of clusters (columns one and two). However, they performed well in both cases of Procedure 2 (columns three and four), which indicates these indexes may not perform correctly at assessing a partition's compactness. The opposite was seen for CDbw, G(+), C-index and AUCC, indicating these indexes may struggle in accurately ranking partitions based on varying levels of separation.

The properties of datasets were seen to have little impact on the ranking of best performing indexes in Evaluation Scenario 3. 
This was consistent across both Procedures 1 and 2, for both cases of a varying number of clusters and a fixed number of clusters. As the various properties showed little impact on high-performing indexes for these tests, rarely resulting in any change in rankings between these indexes, no testing of properties was carried out for Evaluation Scenario 3. The mean correlation for each index in datasets with each property paired with each of the four cases (Procedures 1 and 2 with varied and fixed number of clusters) can be found in the Appendix. These results suggest that the changes in performance previously observed in other evaluation scenarios, in relation to the various properties within the datasets, may result to a large extent from how the partitions produced by clustering algorithms are affected by these various properties, rather than simply from a direct impact on the behaviour of internal validity indexes.

\begin{table*}[h!]
\centering
\begin{tabular}{rcccc}
  \hline
 & Procedure 1 & Procedure 1 & Procedure 2 & Procedure 2 \\ 
 Index & Varied $k$ & Fixed $k$ & Varied $k$ & Fixed $k$ \\
  \hline
NMI4 & 1.00 & -0.07 & 1.00 & 0.19 \\ 
  Powers & 1.00 & -0.06 & 1.00 & 0.19 \\ 
  ARI & 1.00 & -0.07 & 1.00 & 0.18 \\ 
  Jaccard & 1.00 & -0.07 & 1.00 & 0.18 \\ 
  SK3 & 1.00 & -0.07 & 1.00 & 0.18 \\ 
  NID & 1.00 & -0.07 & 1.00 & 0.19 \\ 
  CDistance & 1.00 & 0.45 & 1.00 & -0.03 \\ 
   \hline
\end{tabular}
\caption{\label{Table:Exp3CorE} Mean Spearman correlation between the known referential ranks (produced by Procedures 1 and 2 in Evaluation Scenario 3) and those according to the external indexes used in previous evaluation scenarios.}
\end{table*}

Table \ref{Table:Exp3CorE} displays the correlations between the external validity indexes used in previous evaluation scenarios and the referential rankings produced by Procedures 1 and 2 here in Evaluation Scenario 3. We can see that for the two cases of varied $k$ (columns one and three), all external indexes performed as expected, in that there is a maximal correlation of 1. However, for the case of fixed $k$, only CDistance is positively correlated for Procedure 1 (column two), while no index prominently correlates with the known rankings for Procedure 2 (column four). This behaviour is expected as all external indexes, except for CDistance, only utilise partition labels, as discussed in Section \ref{ExtInd}. For the scenarios with a fixed number of clusters, there is often not a significant difference in quality in terms of partition labels, instead the difference between partitions is largely geometry based. This demonstrates a fundamental flaw in most external validity indexes in terms of their ability (or lack thereof) to determine the quality of partitions with respect to geometric differences. This also means that the results of Evaluation Scenario 2 should be interpreted with caution as they are potentially partially impacted by the use of a subset of external indexes that can fail to properly rank partitions for a fixed number of clusters.

\section{Discussion} \label{discussion}

Datasets and their properties were seen to have a significant impact on the ranking and decision of which internal validity index should be used. In Evaluation Scenario 1, it was seen that when comparing the Type 1 and Type 2 datasets, there were significant disagreements in the performance of each index. This was further observed with respect to the various properties of the Type 2 datasets, where the absolute and relative performances of indexes changed based on how the datasets differ according to those properties. This would in principle suggest that analysts should take such properties into account when selecting internal validity indexes for practical purposes. However, with the exception of dimensionality, the other properties investigated within this paper are not directly observable in unlabelled real-world datasets. This calls for further research on proxy observable measures for those properties and if/how these could be potentially used to guide a tailored choice of more suitable indexes for a given dataset at hand. This also stresses the importance of relying on data that is representative of the class of problems of interest when performing benchmark studies of existing validity indexes or newly proposed ones. The datasets within this paper have encompassed a wide range of properties in comparison to previous papers, however, it can still be further extended to focus on aspects such as including non-globular clustering problems.

It is therefore arguable that there is no such a thing as the best internal validity index(es) overall, unless a specific class of clustering problems of interest is specified. \emph{An important finding in this paper is that, when it comes to performance of internal validity indexes, such a class of clustering problems does not depend only on the properties of the datasets but also on the choice of clustering algorithm}. Indeed, the algorithm used to produce the candidate partitions has been found to have a higher impact on an index's performance than the datasets themselves. Since the algorithm is not only known but it is also typically a design choice, a paired choice of an index (or indexes) that is more suitable for the algorithm in hand is strongly recommended based on our results. Some specific recommendations in this regard will be summarized in Section \ref{conclusion}.

Several non-linear relationships appeared when assessing the correlation between external and internal rankings of partitions for Evaluation Scenario 1. One interesting feature about these non-linear relationships is they frequently appeared as two distinct regions when plotted. The dividing point between these regions was often found to be between partitions with fewer and greater clusters compared to the optimal partition. The primary cause of these distinct regions is due to there being a difference in how partitions with too many clusters are evaluated compared to how partitions with too few clusters are evaluated. Most validity indexes are based on measures of separation and compactness, basically differing in their strategies to determine the best combination of each in their aim for partitions with compact and separated clusters. Partitions with too few clusters are expected to exhibit poor compactness since at least one cluster in the partition must contain points from multiple ground-truth clusters, while partitions with too many clusters are expected to feature poor separation due to at least one cluster in the ground truth being partitioned into two or more different clusters. This results in many of the internal validity indexes functioning noticeably differently between these regions. When the ground-truth partition is not found by a clustering algorithm and, accordingly, it is not present among the candidates, it has been observed that the dividing point most often becomes the best partition according to the external index (``Top Pick''), even when its number of clusters differ from the number of clusters in the ground-truth partition, as this partition will generally display the best compromise between separation and compactness among the collection of candidates. This holds particularly true for hierarchical clustering algorithms, as partitions with more clusters than the ``Top Pick'' one can only be generated by splitting clusters that exist within the ``Top Pick'' partition, thus reducing its separation. Similarly, partitions with fewer clusters than the ``Top Pick'' partition can only be produced by joining its clusters, thus reducing its compactness.

The different behaviours of both external and internal indexes dependent on the number of clusters in candidate partitions was one of the primary causes of non-linear relationships. The use of an aggregated external ranking helped mitigate such differences from the external index's perspective. Differing scales between internal and external indexes was another contributing factor to non-linear relationships and inconsistent performance based on correlation, which is entirely removed through the use of Spearman correlation. These changes, however, did come with a negative side: when several partitions are scored similarly by the external index and/or by the internal indexes, then the total penalty due to several slight differences in ranks across the internal and external evaluations can be exaggerated by the rank-based correlation, resulting in poor performance, when it may not be reasonable to expect an internal validity index to be able to distinguish between the partitions in question as they are all similar in quality. However, this aspect is outweighed by the benefits, as long as it is not overlooked when assessing results produced following such a rank-correlation-based evaluation methodology.

As most non-linear relationships are a result of comparing clustering solutions with differing numbers of cluster, either due to differences in how internal and external indexes evaluate partitions or due to certain behaviours of internal indexes relating to the number of clusters, Evaluation Scenarios 2 and 3 did not present any such non-linear relationships. For Evaluation Scenario 3, this can also be due to the partitions produced in this setting being constructed directly from the known solution to the clustering problem and, as such, being closer to it in that sense.

Among the density-based validation indexes, DBCV and CDbw stood out, having performed comparably to (or among) top ranked traditional non density-based indexes in different scenarios. However, the remainder density-based indexes generally performed worse than most traditional indexes in terms of both identifying the best candidate partition as well as in terms of their correlations with external validity indexes. Their performance tended to improve only in settings that arguably contained easier clustering problems. This was the case e.g. with CVNN and CVDD for Type 1 data in Evaluation Scenario 1 and for three of the four tests in Evaluation Scenario 3. On the other hand, it should be noted that the clustering problems faced by these indexes involved only globular clusters, while density-based indexes may have an advantage in problems featuring non-globular clusters.

The results for Evaluation Scenario 3 were largely consistent with the results of Evaluation Scenario 1, however, there were some discrepancies in the performance of indexes for the case where the number of clusters were varied, the most notable of which was for the Silhouette index. The cause of this discrepancy was discussed within Section~\ref{sec:eval3}, with the main conclusion that the crafted, systematically generated candidate partitions of a limited dataset sub-collection (from which they could have been produced) in Evaluation Scenario 3 cannot fully portray the much higher diversity of clustering solutions considered in Scenario 1. As such, they have not allowed the corresponding evaluation methodology to fully capture the greater robustness of certain internal index (most noticeably the Silhouette) to more diverse types of clustering problems, involving candidate solutions produced by various real-world clustering algorithms from datasets with varied properties and levels of complexity.
Despite this, Evaluation Scenario 3 has played a critical role in effectively disambiguating the impact of clustering algorithms and external validity indexes on the results.

Overall, Evaluation Scenario 3 has shown limited external-index-related bias impacting the results of Evaluation Scenario 1, which involves problems with a varying number of clusters, with generally consistent results between the two scenarios. This strongly endorses the conclusions for Evaluation Scenario 1, where Silhouette, Wemmert-Gancarski, WB, VRC, Point-Biserial and DBCV stood out in performance.

When comparing the results of Evaluation Scenarios 2 and 3 for a fixed number of clusters, we saw significant discrepancies in cross-scenario performance of many indexes, both in absolute as well as in relative terms, and in how performance compared between cases where there are fewer or more clusters than the optimal partition in each of these scenarios. Two factors contributing to these discrepancies were: (i) differences in how partitions are formed by the 8 clustering algorithms in Scenario 2 as opposed to the artificial methods adopted in Scenario 3; and (ii) the reliability (or lack thereof) of external validity indexes adopted in Scenario 2 in properly assessing the quality of partitions when the number of clusters is fixed. 

Regarding the former, as previously discussed, several aspects of more complex clustering problems such as those in Evaluation Scenarios 2 and 1 cannot be fully captured by the synthetic, controlled approach adopted in Evaluation Scenario 3.
The simpler, artificially generated clustering solutions in Scenario 3 were adopted mainly due to difficulties in maintaining a justifiable referential ranking for candidate partitions containing more complex geometries without the use of an external index, but also to exclude the clustering algorithm factor from the analyses. More diverse candidate solutions produced by various real-world clustering algorithms from datasets with varied properties and levels of complexity will impact each validity index differently, and again, like in Scenario 1, it partially explains differences in performance between Scenarios 2 and 3. 

Regarding the questionable reliability of external indexes when the number of clusters was fixed, these indexes were shown to have little to no correlation with the known referential rankings adopted in Scenario 3, with the exception of CDistance in a particular experimental setting (Procedure 1). This behaviour may artificially reduce the level of agreement (correlation) with the internal indexes, thus unfairly penalising evaluation of their actual performance once the external indexes used as reference may not be accurately ranking the partitions. This result indicates that experiments involving partitions somehow similar to those produced by Procedures 1 and 2 in Evaluation Scenario 3 could have been not accurately assessed by the methodology adopted in Evaluation Scenario 2.

Given the above considerations, it appears that neither evaluation scenario (2 or 3) suffices alone for assessing the performance of internal indexes for a fixed number of clusters. However, indexes that performed well in both scenarios appear to be reliable and fit for purpose in this setting, namely, Wemmert-Gancarski, WB, VRC, DBCV and PBM.

While this study advances the understanding of clustering validity indices, it also presents certain limitations that should be acknowledged. Although novel behaviours of internal validity indexes and their relationships to different aspects of clustering were identified, this study does not aim to explain the underlying reasons why specific indexes exhibit these patterns. Addressing these explanatory questions lies beyond the scope of the present work and represents an important direction for future research.

It should also be noted that certain data generation properties are not fully independent, which may confound the interpretation of their individual effects. For example, overlap cannot be enforced directly but is influenced by the compactness and separation of the generated clusters; reducing compactness or separation tends to increase overlap. Similarly, dimensionality may interact with problem difficulty, as adding additional informative dimensions may make clustering problems easier.

Finally, in Evaluation Scenario 1, approximately 11.5\% of datasets were excluded because the clustering algorithms failed to produce a partition of sufficient quality as measured by an external validity index. This exclusion may introduce some bias, particularly against indexes that might perform well on these specific datasets. However, evaluating index performance on datasets where no discernible clustering structure was found was not considered methodologically appropriate, as such cases do not allow for a meaningful assessment of clustering validity under the framework used within this study. 

\subsection{Comparison To Previous Study} \label{comp}

This section compares the results of the original benchmark study in \cite{Vendramin2010} with our current results from Evaluation Scenario 1 (Section \ref{method1}). Recall that the methodology in Evaluation Scenario 1 was specifically built upon and designed to improve on the methodology originally adopted in \cite{Vendramin2010}. Our results also expand on the results in \cite{Vendramin2010} in that they additionally utilise the new Type 2 dataset collection, which is significantly more representative of a more diverse class of clustering problems than the original Type 1 collection. For this reason, our comparisons focus on the results from the Type 2 datasets. For a comparison of the Type 1 and Type 2 dataset experiments, we refer the reader to Section \ref{method1} (Evaluation Scenario 1). Notice that only Evaluation Scenario 1 is considered in the comparisons in this section because \cite{Vendramin2010} didn't include experiments following methodologies comparable to the ones adopted in our Evaluation Scenarios 2 or 3.

In the previous study \cite{Vendramin2010}, VRC performed among the best in determining the optimal number of clusters, however, performed poorly in most correlation based tests. With the updated methodology in our study, we instead saw consistency between the performances of VRC both in terms of selecting the optimal partition (``Top-Pick'') as well as in terms of correlation with external indexes, performing among the best by all criteria. This was likely due to the use of the Spearman correlation, which is insensitive to a frequently observed non-linear yet monotonic (exponential) trend of VRC's $\times$ external values, a behaviour that significantly reduced the Pearson correlation measure in the previous study. Additionally, VRC was also noticed to exhibit more protruding non-linear relationships with some particular external indexes, however, these were lessened in our current results by the use of an aggregate external ranking. Originally in \cite{Vendramin2010}, VRC performed best when the number of clusters present in the ground-truth partition was low, which was an artefact primarily due to the aforementioned exponential behaviour. As mentioned above, this effect has been neutralised in our study, revealing VRC's performance as being nearly identical between cases with few and many clusters. Similar to the previous study, VRC performed better in datasets with higher dimensionality. 

In both studies the Dunn index performed poorly in tests involving correlation, despite otherwise reasonable performance in determining the optimal number of clusters in \cite{Vendramin2010} and, to some extent, in determining the optimal partition (``Top-Pick'') here in our study as well. The inclusion of more difficult datasets of Type 2 caused Dunn’s performance in terms of correlation to drop even further in our results. This reinforces the original findings from \cite{Vendramin2010}, where it is apparent that among the clustering problems studied, there doesn't seem to be a scenario where the choice of this index should be considered over others. 

In the original study \cite{Vendramin2010}, the DB index presented a higher correlation in datasets with fewer clusters, however, relative to other indexes it overall performed better when more clusters were present in the ground-truth partition. In our study, DB has instead a higher correlation with datasets containing a larger number of clusters in the ground-truth partition, and its relative performance to other indexes is barely affected by this property. Irrespective of these differences, in both studies DB exhibited intermediate to low performance in most evaluation scenarios, so the original overall recommendation regarding this index remains unchanged: among the clustering problems studied, there doesn't seem to be a scenario where the choice of this index should be considered over others. 

The Silhouette index, which performed well across all tests in the previous study in \cite{Vendramin2010}, also performed similarly well in our study, with no noticeable changes in behaviour or performance. This index is confirmed as among the best performers overall in both studies.

C-index performed significantly better overall in our study, while still showing similar changes in performance based on changes in dimensionality and number of clusters as in \cite{Vendramin2010}.

In both studies, the Ratkowsky-Lance index (called $C/\sqrt(k)$ in \cite{Vendramin2010}) only performed well in terms of correlation and when few ground-truth clusters were present in the dataset. This is due to the index often acting as monotonic decreasing, which explains why it performed well in the test for few clusters in the previous study, where datasets with only 2, 4 or 6 clusters were involved. In this setting, there isn't much room for candidate partitions to contain fewer clusters than the ground truth, so most candidates are over-clustered, and the more over-clustered they are the lower tends to be their quality as assessed by an external index. As a result, an index with monotonic decreasing behaviour with respect to the number of clusters may be found to be highly correlated with the external index in this particular setting. This is, however, just an artefact that has now been undisclosed in our study, particularly from the results in Table \ref{Table:Exp1CorType2Var}, where Ratkowsky-Lance was negatively correlated when candidate partitions contained fewer clusters compared to the optimal partition, yet highly positively correlated where candidates contained a greater number of clusters.

PBM, similar to VRC, was seen to exhibit non-linear  correlation with external indexes. Once again, the use of a rank-based correlation measure in our study prevented this behaviour from unfairly penalising its correlation-based assessment, resulting in more consistent performance. In contrast, in the previous study \cite{Vendramin2010}, PBM performed significantly better in terms of correlation for datasets with few clusters due to this effect being less significant when there isn't much room to produce candidate partitions with fewer clusters than the ground truth. Outside this, the index showed similar correlation performances between studies. In terms of the other aspects, however, there are further discrepancies. In \cite{Vendramin2010}, PBM exhibited excellent performance in determining the optimal number of clusters, whereas it has not managed to maintain similar performance in determining the optimal partition (``Top-Pick'') in the more diverse collection of clustering problems considered here in our study.

In \cite{Vendramin2010}, Point-Biserial was indicated as one of the overall best performers, alongside the Silhouette, VRC, and PBM. In particular, it appeared in that study as the top performer in terms of correlation across almost all scenarios, except for the case involving datasets with a combination of very low dimensionality and large number of clusters, where performance significantly dropped. The reasons were not explained in \cite{Vendramin2010}. Based on our results, we noted Point-Biserial being frequently negatively correlated with the external validity indexes when the number of dimensions were 2 or 4. Overall, Point-Biserial still exhibited good overall correlation performance in our study (e.g. see Figures \ref{fig:CorType2}, \ref{fig:CorType2O}, and Table \ref{Table:Exp1CorType2Var}), but not as high as previously observed in \cite{Vendramin2010}, which again can be explained by the more diverse (and potentially more difficult) collection of clustering problems considered here, such as in terms of datasets and clustering algorithms adopted.

It is important to stress that our study included many additional indexes not previously included in \cite{Vendramin2010}, some of which have now appeared among the best performers overall, such as, e.g., Wemmert-Gancarski, DBCV, and WB. It should also be noticed that we cannot compare our findings in terms of how the behaviour of indexes is affected by factors such as noise, cluster overlap, cluster compactness, and cluster distribution, which were not considered in \cite{Vendramin2010}. One of the most important such factors found in our study was the clustering algorithm used to produce the candidate partitions, which showed to significantly impact the performance of indexes in different ways, however, it was not previously explored in \cite{Vendramin2010}.   

\section{Conclusion} \label{conclusion}

In this paper we have carried out a large scale benchmark study of internal clustering validity indexes across three separate test scenarios. We have improved on previous studies through addressing the flaws of using correlation as a measure of performance in addition to the use of traditional methods of measuring index performance. The use of rank-based correlation and an aggregated ranking from multiple external validity indexes, associated with a novel strategy of separating analyses involving over-clustered and under-clustered solutions, aided in removing various differences in scales, artifacts, and misleading effects due to non-linear relationships, thus resulting in a more reliable assessment of performance. We have included datasets containing various levels of 7 different properties, and assessed the performance of 26 (unique) internal validity indexes when assessing candidate clustering solutions produced by 8 different clustering algorithms following different clustering paradigms.

\begin{table*}[h!]
\centering
\resizebox{\textwidth}{!}{%
\begin{tabular}{lcc}
  Index & Clustering Algorithm & Properties \\
  \hline
  Silhouette & \makecell{K-Means, Complete-Linkage, Ward-Linkage, \\ Spectral and HDBCAN*} &  \makecell{Overlap, Imbalance, Noise, $D_{low}$, $D_{high}$, $k_{low}$, Compact, \\ Sparse, Uniform, Gaussian and Logistic} \\
  \hline
  WB & Single-Linkage and Spectral & \makecell{ Overlap, Imbalance, $D_{high}$, $k_{high}$, \\ Sparse, Gaussian, Logistic} \\ 
  \hline
  VRC & Average-Linkage, EM-GMM and Spectral & \makecell{Imbalance, $D_{low}$, $k_{low}$, \\ Sparse and Uniform} \\
  \hline
  Point Biserial &  \makecell{K-Means, Average-Linkage, Complete-Linkage, \\ Ward-Linkage, EM-GMM and HDBSCAN*} & $k_{high}$, Compact and Uniform \\
  \hline
  DBCV & Single-Linkage and EM-GMM & Noise and $D_{high}$ \\
  \hline
  Wemmert-Gancarski & HDBSCAN* & Overlap, Noise, $k_{high}$ and Logistic \\
  \hline
  CDbw & \makecell{ K-Means, Single-Linkage, Average-Linkage,\\ Complete-Linkage and Ward-Linkage} & $D_{low}$, $k_{low}$, Compact and Gaussian \\
\end{tabular}}
\caption{\label{Table:Recommended} Recommended use cases among overall top performing internal validity indexes for both choice of clustering algorithm and properties within datasets. Although CDbw is \emph{not} an overall top performer, it is also included in the selected index collection here due to its top performance for several clustering algorithms when considered independently. An index within this collection is recommended if it falls in the top-3 best performers for each specific case.  Disclaimer: Recommendations should not be  generalised to datasets of a different nature from those included in this study without further assessment of their suitability in different scenarios.}
\end{table*}

Several notable behaviours of external validity indexes were seen throughout this study which have shown to significantly impact the results, the primary two being: (i) Significant differences in how internal and external indexes each handle overestimation or underestimation of the number clusters; and (ii) limitations related to lack of consideration for the geometry of solutions by most external validity indexes. Both of these issues have the potential to significantly impact the results of any assessment of internal clustering validity indexes, for instance, by contributing to the existence of strong non-linear relationships between internal and external indexes, which have been analysed in depth. Three separate evaluation scenarios were performed in an attempt to reduce the individual impact of each of these on the conclusions drawn.

For Evaluation Scenario 1, the indexes Silhouette, WB, VRC, Wemmert-Gancarski, DBCV, and Point-Biserial performed the best overall, appearing among the top performers in most of the tests within that scenario. Yet, the dataset properties, as well as the clustering algorithms used to produce candidate partitions, impacted these indexes, with cases where their performance dropped in absolute and/or in relative terms. Examples include VRC in the presence of noise, Point-Biserial in the presence of overlap, Wemmert-Gancarski when assessing EM-GMM partitions, and DBCV when assessing Spectral Clustering partitions.

The choice of clustering algorithm had an even more noticeable impact on performance than the dataset properties, to the extent that, for three of the eight algorithms (single-, complete-, and average-linkage), the algorithm-specific best index was not one of the aforementioned overall top performers, but rather CDbw, which otherwise only displayed an overall intermediate performance in Evaluation Scenario 1. In fact, CDbw under-performed in the presence of noise, high dimensionality, larger numbers of clusters, and partitions produced by algorithms such as Spectral Clustering and EM-GMM. This emphasises one of the main findings in our study, namely: an index should not be recommended over others without considering the nature of the clustering problem (data and algorithm) at hand.

A summary of which internal validity indexes are recommended for the case of varied number of clusters based on either the clustering algorithm selected, or properties within the dataset, is found within Table~\ref{Table:Recommended}. This table contains the recommended use cases for each of the best overall performing indexes, in addition to CDbw (due to performing as the best, or among the best, for several clustering algorithms independently). An index within this selected group is recommended for each use case if it performs within the top three for each clustering algorithm or property. The recommendations in this paper, in particular Table~\ref{Table:Recommended}, should only be considered within the context of the dataset types included in our experiments. Also, they should only be interpreted in relation to the other indexes studied herein. For instance, as noted in Section~\ref{IntInd}, standard practice for EM-GMM is to use model-based quality measures such as BIC. As we have not compared the performance of algorithm-specific measures against the general-purpose internal indexes within this paper, our recommendations should not be interpreted as claiming superiority over these specialised  measures. 

For Evaluation Scenario 2, which comprises cases where the number of clusters is fixed, Point-Biserial was the top performer overall, exhibiting the best performance in most tests, with WB, Wemmert-Gancarski, VRC and Silhouette again also performing among the best in this scenario. As expected, the properties of the datasets once more impacted these indexes' performances in different ways, for instance, causing noticeable drops in performance for the Silhouette in the presence of overlap or high dimensionality, but overall the impact was less important for Evaluation Scenario 2 tests as they did not significantly change the group of best indexes in most tests.

At a high level, the results of Evaluation Scenario 3 supported the main results of Evaluation Scenarios 1 and complemented those of Scenario 2, reinforcing top performance of some indexes or helping us gain further insights on areas that may have contributed to poor performance of others.

Two density-based validity indexes were seen to perform well, namely, DBCV and CDbw, which outperformed all other density-based validity indexes. Both showed comparable performance to the best indexes in many scenarios. Overall, DBCV exhibited more consistent and robust performance across different tests and evaluation scenarios in comparison to CDbw. 

In summary, several indexes performed well overall across all three evaluation scenarios, namely, DBCV, Point-Biserial, Silhouette, VRC, WB, and Wemmert-Gancarski, with no specific order (other than alphabetical) suggested here, since our results clearly show that the choice of a particular index cannot be made without considering the nature of the clustering problem of interest. Given that certain properties that affect this choice are not directly observable, it is good practice to rely on a collection of more reliable indexes rather than on a single one. We believe that our study can serve as a guide for practitioners in this context, while it also sheds light on aspects that still deserve further investigation by the research community.

\printbibliography

\newpage

\end{document}


\section{Appendix}

\begin{table}[h!]
\centering
\resizebox{\textwidth}{!}{%
\begin{tabular}{rllllllll}
  \hline
 Index & All & Overlap & Imbalance & $k_{low}$ & $k_{high}$ & Compact & Space \\ 
  \hline
S\_Dbw & 0.975 (1) & 0.956 (1) & 0.976 (1) & 0.977 (1) & 0.967 (2) & 0.999 (1) & 0.97 (1) \\ 
  PBM & 0.958 (2) & 0.939 (2) & 0.945 (2) & 0.956 (3) & 0.968 (1) & 0.985 (4) & 0.933 (3) \\ 
  WB & 0.958 (3) & 0.937 (3) & 0.944 (3) & 0.958 (2) & 0.957 (3) & 0.984 (5) & 0.932 (4) \\ 
  CDbw & 0.943 (4) & 0.902 (7) & 0.941 (4) & 0.951 (4) & 0.905 (5) & 0.995 (2) & 0.919 (6) \\ 
  VRC & 0.939 (5) & 0.907 (5) & 0.924 (6) & 0.94 (6) & 0.931 (4) & 0.981 (9) & 0.893 (7) \\ 
  Grex & 0.938 (6) & 0.903 (6) & 0.926 (5) & 0.949 (5) & 0.887 (6) & 0.981 (8) & 0.927 (5) \\ 
  G(+) & 0.917 (7) & 0.915 (4) & 0.902 (7) & 0.932 (7) & 0.847 (7) & 0.904 (16) & 0.94 (2) \\ 
  Wemmert-Gancarski & 0.903 (8) & 0.838 (10) & 0.892 (8) & 0.918 (8) & 0.836 (9) & 0.983 (6) & 0.854 (10) \\ 
  DB & 0.889 (9) & 0.814 (11) & 0.879 (9) & 0.899 (9) & 0.841 (8) & 0.979 (10) & 0.823 (11) \\ 
  C-index & 0.872 (10) & 0.855 (8) & 0.86 (11) & 0.886 (11) & 0.812 (10) & 0.882 (17) & 0.879 (8) \\ 
  CS & 0.862 (11) & 0.776 (12) & 0.861 (10) & 0.887 (10) & 0.745 (15) & 0.963 (12) & 0.769 (12) \\ 
  Silhouette & 0.86 (12) & 0.772 (13) & 0.833 (14) & 0.876 (12) & 0.786 (11) & 0.97 (11) & 0.76 (13) \\ 
  AUCC & 0.857 (13) & 0.838 (9) & 0.851 (12) & 0.873 (13) & 0.781 (12) & 0.872 (19) & 0.859 (9) \\ 
  DBCV & 0.853 (14) & 0.75 (14) & 0.844 (13) & 0.872 (14) & 0.766 (13) & 0.982 (7) & 0.738 (14) \\ 
  SV & 0.811 (15) & 0.693 (16) & 0.813 (15) & 0.836 (15) & 0.692 (18) & 0.951 (13) & 0.669 (15) \\ 
  Gstr & 0.793 (16) & 0.698 (15) & 0.766 (16) & 0.807 (16) & 0.727 (16) & 0.912 (15) & 0.662 (16) \\ 
  CVDD & 0.779 (17) & 0.613 (17) & 0.756 (17) & 0.783 (17) & 0.76 (14) & 0.993 (3) & 0.477 (19) \\ 
  XieBeni & 0.692 (18) & 0.547 (19) & 0.656 (18) & 0.691 (18) & 0.699 (17) & 0.879 (18) & 0.513 (18) \\ 
  CVNN & 0.622 (19) & 0.367 (20) & 0.557 (19) & 0.629 (19) & 0.59 (19) & 0.948 (14) & 0.135 (22) \\ 
  Point-Biserial & 0.572 (20) & 0.549 (18) & 0.531 (20) & 0.584 (20) & 0.513 (20) & 0.536 (21) & 0.64 (17) \\ 
  Dunn & 0.5 (21) & 0.297 (22) & 0.477 (21) & 0.514 (21) & 0.436 (22) & 0.751 (20) & 0.373 (21) \\ 
  SSDD & 0.362 (22) & 0.36 (21) & 0.431 (22) & 0.382 (22) & 0.268 (23) & 0.326 (23) & 0.376 (20) \\ 
  LCCV & 0.303 (23) & 0.165 (23) & 0.279 (23) & 0.269 (23) & 0.462 (21) & 0.495 (22) & 0.0533 (23) \\ 
  SD & -0.251 (24) & -0.188 (24) & -0.282 (24) & -0.239 (24) & -0.305 (25) & -0.402 (24) & -0.134 (24) \\ 
  Tau & -0.323 (25) & -0.269 (25) & -0.357 (25) & -0.33 (25) & -0.289 (24) & -0.453 (25) & -0.25 (25) \\ 
  Ratkowsky-Lance & -0.811 (26) & -0.791 (26) & -0.863 (26) & -0.792 (26) & -0.9 (26) & -0.871 (26) & -0.848 (26) \\ 
   \hline
\end{tabular}}
\caption{\label{Table:Exp3CorP1} Mean Spearman correlation for Evaluation Scenario 3 Procedure 1 with varied clusters separated by each property. The categories are comprised as follows: $k_{high}$ incorporates data with more than 10 ground-truth clusters, $k_{low}$ comprises data with 10 or fewer ground-truth clusters, overlap considers datasets with greater than 0 overlap using Equation \ref{eq:overlap}, Imbalanced datasets feature an imbalance of 0.5 or greater using Equation \ref{eq:imbalance}, Noise considers all datasets containing noise in the ground-truth partition, Compact clusters are defined as being generated with a compactness of 0.1 in MDCGen, Sparse clusters are defined as being generated with a compactness of 0.8 in MDCGen.}
\end{table}

\begin{table*}[h!]
\centering
\resizebox{\textwidth}{!}{%
\begin{tabular}{rllllllll}
  \hline
  Index & All & Overlap & Imbalance & $k_{low}$ & $k_{high}$ & Compact & Sparse \\ 
  \hline
CDbw & 0.685 (1) & 0.664 (1) & 0.676 (1) & 0.682 (1) & 0.706 (1) & 0.747 (1) & 0.619 (2) \\ 
  CVDD & 0.682 (2) & 0.657 (2) & 0.667 (2) & 0.679 (2) & 0.696 (3) & 0.739 (2) & 0.628 (1) \\ 
  CVNN & 0.662 (3) & 0.657 (3) & 0.63 (3) & 0.656 (3) & 0.7 (2) & 0.7 (5) & 0.606 (3) \\ 
  SV & 0.626 (4) & 0.584 (6) & 0.627 (4) & 0.618 (4) & 0.67 (8) & 0.711 (3) & 0.529 (17) \\ 
  CS & 0.624 (5) & 0.587 (5) & 0.619 (5) & 0.617 (5) & 0.668 (9) & 0.705 (4) & 0.54 (15) \\ 
  Tau & 0.623 (6) & 0.595 (4) & 0.59 (6) & 0.61 (6) & 0.693 (4) & 0.695 (6) & 0.592 (4) \\ 
  C-index & 0.613 (7) & 0.584 (7) & 0.576 (8) & 0.605 (7) & 0.657 (11) & 0.689 (11) & 0.578 (5) \\ 
  AUCC & 0.612 (8) & 0.581 (8) & 0.574 (9) & 0.603 (8) & 0.668 (10) & 0.69 (10) & 0.577 (6) \\ 
  VRC & 0.608 (9.5) & 0.577 (9.5) & 0.572 (10.5) & 0.596 (9.5) & 0.681 (6.5) & 0.692 (8.5) & 0.573 (8.5) \\ 
  WB & 0.608 (9.5) & 0.577 (9.5) & 0.572 (10.5) & 0.596 (9.5) & 0.681 (6.5) & 0.692 (8.5) & 0.573 (8.5) \\ 
  PBM & 0.604 (11) & 0.576 (11) & 0.57 (12) & 0.591 (12) & 0.681 (5) & 0.686 (12) & 0.575 (7) \\ 
  Gstr & 0.6 (12) & 0.56 (13) & 0.583 (7) & 0.592 (11) & 0.647 (13) & 0.694 (7) & 0.567 (11) \\ 
  G(+) & 0.598 (13) & 0.565 (12) & 0.552 (13) & 0.59 (13) & 0.641 (14) & 0.678 (13) & 0.559 (12) \\ 
  Ratkowsky-Lance & 0.577 (14) & 0.541 (19) & 0.542 (16) & 0.564 (16) & 0.649 (12) & 0.665 (15) & 0.529 (16) \\ 
  Grex & 0.576 (15) & 0.553 (14) & 0.538 (17) & 0.566 (14) & 0.63 (15) & 0.639 (17) & 0.547 (14) \\ 
  DB & 0.575 (16) & 0.544 (17) & 0.546 (15) & 0.566 (15) & 0.626 (16) & 0.641 (16) & 0.55 (13) \\ 
  S\_Dbw & 0.571 (17) & 0.551 (15) & 0.549 (14) & 0.564 (17) & 0.611 (17) & 0.622 (19) & 0.57 (10) \\ 
  Silhouette & 0.555 (18) & 0.534 (20) & 0.5 (20) & 0.552 (18) & 0.573 (21) & 0.614 (20) & 0.508 (20) \\ 
  Wemmert-Gancarski & 0.553 (19) & 0.541 (18) & 0.504 (19) & 0.547 (19) & 0.588 (19) & 0.603 (21) & 0.511 (19) \\ 
  DBCV & 0.542 (20) & 0.55 (16) & 0.496 (21) & 0.535 (21) & 0.58 (20) & 0.566 (22) & 0.516 (18) \\ 
  Point-Biserial & 0.54 (21) & 0.507 (21) & 0.467 (22) & 0.537 (20) & 0.562 (22) & 0.625 (18) & 0.479 (21) \\ 
  Dunn & 0.529 (22) & 0.47 (22) & 0.53 (18) & 0.518 (22) & 0.589 (18) & 0.674 (14) & 0.474 (22) \\ 
  XieBeni & 0.446 (23) & 0.415 (23) & 0.415 (23) & 0.431 (23) & 0.536 (23) & 0.507 (23) & 0.426 (23) \\ 
  SD & 0.34 (24) & 0.33 (24) & 0.312 (24) & 0.333 (24) & 0.385 (24) & 0.354 (24) & 0.342 (24) \\ 
  LCCV & 0.101 (25) & 0.16 (25) & 0.0706 (25) & 0.102 (25) & 0.0975 (25) & 0.0202 (26) & 0.132 (25) \\ 
  SSDD & 0.0135 (26) & -0.0101 (26) & 0.0153 (26) & 0.0223 (26) & -0.0371 (26) & 0.0477 (25) & 0.0125 (26) \\ 
   \hline
\end{tabular}}
\caption{\label{Table:Exp3CorP2} Mean Spearman correlation for Evaluation Scenario 3 Procedure 1 with fixed clusters separated by each property. The categories are comprised as follows: $k_{high}$ incorporates data with more than 10 ground-truth clusters, $k_{low}$ comprises data with 10 or fewer ground-truth clusters, overlap considers datasets with greater than 0 overlap using Equation \ref{eq:overlap}, Imbalanced datasets feature an imbalance of 0.5 or greater using Equation \ref{eq:imbalance}, Noise considers all datasets containing noise in the ground-truth partition, Compact clusters are defined as being generated with a compactness of 0.1 in MDCGen, Sparse clusters are defined as being generated with a compactness of 0.8 in MDCGen.}
\end{table*}

\begin{table*}[h!]
\centering
\resizebox{\textwidth}{!}{%
\begin{tabular}{rlllllllll}
  \hline
  Index & All & Overlap & Imbalance & $D_{low}$ & $D_{high}$ & $k_{low}$  &  Compact & Sparse \\ 
  \hline
Ratkowsky-Lance & 0.988 (1) & 0.977 (1) & 0.987 (1) & 0.981 (1) & 0.999 (1) & 0.986 (1) & 0.998 (1) & 1 (2) & 0.972 (2) \\ 
  VRC & 0.983 (2) & 0.969 (2) & 0.981 (2) & 0.973 (2) & 0.998 (2) & 0.982 (2) & 0.989 (3) & 0.996 (11) & 0.972 (1) \\ 
  Wemmert-Gancarski & 0.979 (3) & 0.959 (3) & 0.975 (3) & 0.971 (3) & 0.992 (3) & 0.978 (3) & 0.991 (2) & 1 (2) & 0.95 (3) \\ 
  WB & 0.966 (4) & 0.941 (6) & 0.963 (4) & 0.951 (5) & 0.99 (4) & 0.964 (4) & 0.98 (4) & 0.992 (12) & 0.945 (6) \\ 
  CVDD & 0.965 (5) & 0.944 (4) & 0.955 (5) & 0.955 (4) & 0.98 (5) & 0.964 (5) & 0.971 (5) & 0.991 (14) & 0.947 (4) \\ 
  Grex & 0.949 (6) & 0.906 (8) & 0.944 (7) & 0.933 (10) & 0.976 (6) & 0.95 (7) & 0.945 (8) & 0.999 (8) & 0.917 (7) \\ 
  Silhouette & 0.949 (7) & 0.905 (9) & 0.951 (6) & 0.937 (8) & 0.969 (8) & 0.948 (9) & 0.955 (7) & 1 (2) & 0.897 (9) \\ 
  Gstr & 0.948 (8) & 0.92 (7) & 0.943 (8) & 0.94 (7) & 0.959 (9) & 0.948 (8) & 0.942 (10) & 0.998 (9) & 0.913 (8) \\ 
  CVNN & 0.947 (9) & 0.943 (5) & 0.922 (11) & 0.95 (6) & 0.942 (12) & 0.95 (6) & 0.925 (11) & 0.947 (18) & 0.947 (5) \\ 
  DBCV & 0.941 (10) & 0.888 (10) & 0.942 (9) & 0.934 (9) & 0.951 (10) & 0.937 (10) & 0.969 (6) & 1 (4) & 0.868 (10) \\ 
  DB & 0.933 (11) & 0.877 (11) & 0.925 (10) & 0.909 (11) & 0.972 (7) & 0.931 (11) & 0.942 (9) & 0.997 (10) & 0.858 (11) \\ 
  CS & 0.903 (12) & 0.816 (12) & 0.894 (12) & 0.878 (12) & 0.943 (11) & 0.901 (12) & 0.914 (12) & 1 (5) & 0.768 (14) \\ 
  PBM & 0.879 (13) & 0.808 (13) & 0.867 (13) & 0.846 (13) & 0.933 (13) & 0.881 (13) & 0.866 (15) & 0.992 (13) & 0.796 (12) \\ 
  SV & 0.87 (14) & 0.761 (14) & 0.86 (14) & 0.843 (14) & 0.915 (14) & 0.869 (14) & 0.877 (13) & 0.999 (6) & 0.706 (15) \\ 
  Tau & 0.836 (15) & 0.701 (16) & 0.84 (15) & 0.812 (15) & 0.875 (15) & 0.831 (15) & 0.872 (14) & 0.999 (7) & 0.666 (16) \\ 
  SD & 0.721 (16) & 0.747 (15) & 0.717 (16) & 0.703 (16) & 0.751 (17) & 0.721 (16) & 0.718 (18) & 0.652 (22) & 0.794 (13) \\ 
  C-index & 0.71 (17) & 0.487 (19) & 0.714 (17) & 0.682 (17) & 0.757 (16) & 0.702 (17) & 0.773 (17) & 0.973 (17) & 0.468 (19) \\ 
  CDbw & 0.684 (18) & 0.444 (20) & 0.663 (19) & 0.657 (18) & 0.728 (18) & 0.669 (18) & 0.793 (16) & 0.943 (19) & 0.401 (21) \\ 
  Dunn & 0.654 (19) & 0.62 (17) & 0.663 (18) & 0.622 (19) & 0.705 (19) & 0.654 (19) & 0.654 (21) & 0.674 (21) & 0.661 (17) \\ 
  AUCC & 0.603 (20) & 0.387 (21) & 0.618 (20) & 0.615 (20) & 0.583 (20) & 0.593 (20) & 0.672 (19) & 0.988 (15) & 0.294 (23) \\ 
  S\_Dbw & 0.566 (21) & 0.565 (18) & 0.579 (21) & 0.575 (21) & 0.551 (21) & 0.565 (21) & 0.571 (22) & 0.574 (23) & 0.571 (18) \\ 
  Point-Biserial & 0.529 (22) & 0.234 (25) & 0.527 (22) & 0.534 (22) & 0.52 (22) & 0.51 (22) & 0.659 (20) & 0.976 (16) & 0.133 (25) \\ 
  XieBeni & 0.398 (23) & 0.37 (22) & 0.389 (23) & 0.357 (23) & 0.466 (23) & 0.406 (23) & 0.343 (24) & 0.417 (24) & 0.428 (20) \\ 
  G(+) & 0.293 (24) & -0.0933 (26) & 0.32 (25) & 0.294 (24) & 0.292 (25) & 0.272 (24) & 0.443 (23) & 0.936 (20) & -0.176 (26) \\ 
  LCCV & 0.269 (25) & 0.324 (23) & 0.279 (26) & 0.254 (25) & 0.294 (24) & 0.267 (25) & 0.28 (26) & 0.301 (25) & 0.288 (24) \\ 
  SSDD & 0.217 (26) & 0.241 (24) & 0.323 (24) & 0.202 (26) & 0.243 (26) & 0.205 (26) & 0.305 (25) & 0.138 (26) & 0.3 (22) \\ 
   \hline
\end{tabular}}
\caption{\label{Table:Exp3CorP3} Mean Spearman correlation for Evaluation Scenario 3 Procedure 2 with varied clusters separated by each property. The categories are comprised as follows: $k_{high}$ incorporates data with more than 10 ground-truth clusters, $k_{low}$ comprises data with 10 or fewer ground-truth clusters, $D_{low}$ refers to datasets with 25 or fewer dimensions, $D_{high}$ incorporates datasets with more than 25 dimensions, overlap considers datasets with greater than 0 overlap using Equation \ref{eq:overlap}, Imbalanced datasets feature an imbalance of 0.5 or greater using Equation \ref{eq:imbalance}, Noise considers all datasets containing noise in the ground-truth partition, Compact clusters are defined as being generated with a compactness of 0.1 in MDCGen, Sparse clusters are defined as being generated with a compactness of 0.8 in MDCGen.}
\end{table*}

\begin{table*}[h!]
\centering
\resizebox{\textwidth}{!}{%
\begin{tabular}{rlllllllll}
  \hline
 Index & All & Overlap & Imbalance & $D_{low}$ & $D_{high}$ & $k_{low}$ & $k_{high}$ & Compact & Sparse \\ 
  \hline
DB & 0.386 (1) & 0.357 (1) & 0.376 (1) & 0.376 (1) & 0.402 (1) & 0.372 (1) & 0.483 (1) & 0.388 (1) & 0.386 (1) \\ 
  VRC & 0.361 (2.5) & 0.342 (2.5) & 0.35 (2.5) & 0.356 (2.5) & 0.369 (2.5) & 0.354 (2.5) & 0.411 (2.5) & 0.368 (2.5) & 0.367 (2.5) \\ 
  WB & 0.361 (2.5) & 0.342 (2.5) & 0.35 (2.5) & 0.356 (2.5) & 0.369 (2.5) & 0.354 (2.5) & 0.411 (2.5) & 0.368 (2.5) & 0.367 (2.5) \\ 
  Ratkowsky-Lance & 0.349 (4) & 0.333 (4) & 0.33 (4) & 0.339 (4) & 0.365 (4) & 0.343 (4) & 0.394 (4) & 0.357 (4) & 0.355 (5) \\ 
  XieBeni & 0.325 (5) & 0.315 (7) & 0.317 (5) & 0.3 (6) & 0.364 (5) & 0.319 (5) & 0.368 (6) & 0.328 (6) & 0.317 (7) \\ 
  CS & 0.32 (6) & 0.327 (5) & 0.314 (6) & 0.311 (5) & 0.335 (7) & 0.312 (6) & 0.375 (5) & 0.265 (9) & 0.356 (4) \\ 
  SV & 0.312 (7) & 0.319 (6) & 0.308 (7) & 0.296 (8) & 0.338 (6) & 0.306 (7) & 0.356 (7) & 0.262 (10) & 0.35 (6) \\ 
  PBM & 0.309 (8) & 0.275 (8) & 0.303 (8) & 0.298 (7) & 0.328 (8) & 0.303 (8) & 0.354 (8) & 0.352 (5) & 0.287 (8) \\ 
  Wemmert-Gancarski & 0.29 (9) & 0.263 (9) & 0.286 (9) & 0.269 (9) & 0.323 (9) & 0.286 (9) & 0.319 (10) & 0.298 (8) & 0.285 (9) \\ 
  SD & 0.265 (10) & 0.24 (10) & 0.253 (10) & 0.23 (10) & 0.321 (10) & 0.257 (10) & 0.319 (9) & 0.312 (7) & 0.213 (11) \\ 
  Silhouette & 0.212 (11) & 0.175 (12) & 0.209 (11) & 0.187 (12) & 0.252 (11) & 0.21 (12) & 0.222 (12) & 0.232 (12) & 0.201 (14) \\ 
  C-index & 0.21 (12) & 0.182 (11) & 0.2 (12) & 0.187 (11) & 0.245 (12) & 0.21 (11) & 0.204 (13) & 0.22 (13) & 0.212 (12) \\ 
  AUCC & 0.197 (13) & 0.17 (13) & 0.185 (14) & 0.167 (13) & 0.244 (13) & 0.196 (13) & 0.2 (14) & 0.206 (14) & 0.214 (10) \\ 
  Tau & 0.196 (14) & 0.163 (14) & 0.186 (13) & 0.167 (14) & 0.241 (14) & 0.195 (14) & 0.198 (15) & 0.199 (16) & 0.211 (13) \\ 
  Gstr & 0.156 (15) & 0.089 (18) & 0.161 (15) & 0.139 (15) & 0.183 (16) & 0.146 (16) & 0.226 (11) & 0.245 (11) & 0.111 (18) \\ 
  G(+) & 0.154 (16) & 0.106 (17) & 0.147 (16) & 0.111 (17) & 0.223 (15) & 0.154 (15) & 0.152 (17) & 0.205 (15) & 0.137 (15) \\ 
  Dunn & 0.116 (17) & 0.109 (15) & 0.113 (17) & 0.119 (16) & 0.111 (19) & 0.119 (17) & 0.0955 (18) & 0.118 (19) & 0.113 (17) \\ 
  DBCV & 0.108 (18) & 0.0815 (19) & 0.108 (18) & 0.0901 (19) & 0.136 (18) & 0.111 (18) & 0.0874 (19) & 0.126 (18) & 0.0982 (19) \\ 
  Grex & 0.0957 (19) & 0.107 (16) & 0.0927 (19) & 0.0974 (18) & 0.0931 (20) & 0.0854 (19) & 0.168 (16) & 0.0613 (20) & 0.133 (16) \\ 
  Point-Biserial & 0.085 (20) & 0.0292 (20) & 0.0783 (20) & 0.045 (20) & 0.149 (17) & 0.0848 (20) & 0.0865 (20) & 0.184 (17) & 0.0125 (20) \\ 
  LCCV & 0.00348 (21) & 0.000133 (21) & 0.0134 (21) & -8.39e-05 (21) & 0.00915 (21) & 0.00544 (21) & -0.0101 (21) & 0.00903 (21) & -0.000157 (21) \\ 
  SSDD & -0.0442 (22) & -0.0323 (23) & -0.0545 (23) & -0.0338 (22) & -0.0608 (22) & -0.0431 (22) & -0.0518 (23) & -0.0405 (22) & -0.0459 (23) \\ 
  CVDD & -0.0521 (23) & -0.0127 (22) & -0.0357 (22) & -0.0399 (23) & -0.0716 (23) & -0.0513 (23) & -0.0578 (24) & -0.0875 (24) & -0.0282 (22) \\ 
  CDbw & -0.0746 (24) & -0.0619 (24) & -0.0695 (24) & -0.0624 (24) & -0.0939 (24) & -0.0809 (24) & -0.0309 (22) & -0.0774 (23) & -0.0773 (24) \\ 
  S\_Dbw & -0.0966 (25) & -0.0978 (26) & -0.0803 (25) & -0.0854 (25) & -0.114 (25) & -0.0926 (25) & -0.124 (26) & -0.107 (25) & -0.119 (25) \\ 
  CVNN & -0.116 (26) & -0.0918 (25) & -0.114 (26) & -0.0955 (26) & -0.148 (26) & -0.121 (26) & -0.0832 (25) & -0.12 (26) & -0.131 (26) \\ 
  \hline
\end{tabular}}
\caption{\label{Table:Exp3CorP4} Mean Spearman correlation for Evaluation Scenario 3 Procedure 2 with fixed clusters separated by each property. The categories are comprised as follows: $k_{high}$ incorporates data with more than 10 ground-truth clusters, $k_{low}$ comprises data with 10 or fewer ground-truth clusters, $D_{low}$ refers to datasets with 25 or fewer dimensions, $D_{high}$ incorporates datasets with more than 25 dimensions, overlap considers datasets with greater than 0 overlap using Equation \ref{eq:overlap}, Imbalanced datasets feature an imbalance of 0.5 or greater using Equation \ref{eq:imbalance}, Noise considers all datasets containing noise in the ground-truth partition, Compact clusters are defined as being generated with a compactness of 0.1 in MDCGen, Sparse clusters are defined as being generated with a compactness of 0.8 in MDCGen.}
\end{table*}